\pdfoutput=1
\documentclass[11pt,letterpaper,twoside]{article}
\usepackage[utf8]{inputenc}
\usepackage[letterpaper,margin=1.375in]{geometry}
\usepackage[document]{ragged2e}
\usepackage[singlespacing]{setspace}
\usepackage{lineno}
\usepackage{xcolor}

\usepackage{fancyhdr}

\usepackage{sectsty}
\allsectionsfont{\bfseries}
\sectionfont{\centering\large}
\subsectionfont{\normalsize}

\usepackage[natbibapa,nodoi]{apacite}
\makeatletter
\NAT@longnamesfalse
\makeatother

\usepackage{amsmath,amssymb}
\numberwithin{equation}{section}
\usepackage[hidelinks]{hyperref}
\usepackage{enumitem}

\usepackage{graphicx}
\usepackage[format=plain,labelformat=empty,font=small]{caption}

\newcommand{\panelcase}{\uppercase}
\newcommand{\panel}[1]{\panelcase{(#1)}}

\newcommand{\fig}[2]{Fig.~\ref{fig:#1}\panelcase{#2}}

\newcommand{\sfig}[2]{Figure~\ref{fig:#1}\panelcase{#2}}

\newcommand{\mainfigure}[1]{
\begin{figure}[ht!]
	\centering
	\includegraphics{#1}
	\caption{}\label{fig:#1}
	\vspace{-.33in}
\end{figure}
}

\newcommand{\maincaption}[3]{{\small\fontfamily{cmss}\selectfont
{\sfig{#1}{}:}\; {#2} {}#3
}}

\newcommand{\suppfig}[2]{Supplementary Fig.~\ref{suppfig:#1}\panelcase{#2}}

\newcommand{\ssuppfig}[2]{Supplementary Figure~\ref{suppfig:#1}\panelcase{#2}}

\newcommand{\suppfigure}[1]{
\begin{figure}[ht!]
	\centering
	\includegraphics{#1}
	\caption{}\label{suppfig:#1}
	\vspace{-.33in}
\end{figure}
}

\newcommand{\suppcaption}[3]{{\small\fontfamily{cmss}\selectfont
{\ssuppfig{#1}{}:}\; {#2} {}#3
}}

\usepackage{booktabs}

\newcommand{\tbl}[1]{Table~\ref{tab:#1}}

\newcommand{\maintablecaption}[3]{{\small\fontfamily{cmss}\selectfont
{\tbl{#1}:}\; {#2} {}#3
\vspace{-.33in}
}}

\newcommand{\supptbl}[1]{Supplementary Table~\ref{tab:#1}}

\newcommand{\supptablecaption}[3]{{\small\fontfamily{cmss}\selectfont
{\supptbl{#1}:}\; {#2} {}#3
\vspace{-.33in}
}}

\newcommand{\apptbl}[1]{Table A.\ref{tab:#1}}

\newcommand{\apptablecaption}[3]{{\small\fontfamily{cmss}\selectfont
{\apptbl{#1}:}\; {#2} {}#3
\vspace{-.33in}
}}

\title{\textbf{\Large{Fast Multi-Group Gaussian Process Factor Models}}}

\usepackage{authblk}

\author[1*]{Evren Gokcen}
\author[2]{Anna I. Jasper}
\author[2,3,4$\dagger$]{\\Adam Kohn}
\author[5$\dagger$]{Christian K. Machens}
\author[1,6$\dagger$]{Byron M. Yu}

\affil[1]{Dept. of Electrical and Computer Engineering, Carnegie Mellon University}
\affil[2]{Dominick Purpura Dept. of Neuroscience, Albert Einstein College of Medicine}
\affil[3]{Dept. of Ophthalmology and Visual Sciences, Albert Einstein College of Medicine}
\affil[4]{Dept. of Systems and Computational Biology, Albert Einstein College of Medicine}
\affil[5]{Champalimaud Neuroscience Programme, Champalimaud Foundation}
\affil[6]{Dept. of Biomedical Engineering, Carnegie Mellon University}
\affil[*]{Corresponding author.}
\affil[$\dagger$]{These authors contributed equally to this work.}

\date{}

\begin{document}

\maketitle
\thispagestyle{empty}

\section*{Abstract}

Gaussian processes are now commonly used in dimensionality reduction approaches tailored to neuroscience, especially to describe changes in high-dimensional neural activity over time.
As recording capabilities expand to include neuronal populations across multiple brain areas, cortical layers, and cell types, interest in extending Gaussian process factor models to characterize multi-population interactions has grown.
However, the cubic runtime scaling of current methods with the length of experimental trials and the number of recorded populations (groups) precludes their application to large-scale multi-population recordings.
Here, we improve this scaling from cubic to linear in both trial length and group number.
We present two approximate approaches to fitting multi-group Gaussian process factor models based on (1) inducing variables and (2) the frequency domain.
Empirically, both methods achieved orders of magnitude speed-up with minimal impact on statistical performance, in simulation and on neural recordings of hundreds of neurons across three brain areas.
The frequency domain approach, in particular, consistently provided the greatest runtime benefits with the fewest trade-offs in statistical performance.
We further characterize the estimation biases introduced by the frequency domain approach and demonstrate effective strategies to mitigate them.
This work enables a powerful class of analysis techniques to keep pace with the growing scale of multi-population recordings, opening new avenues for exploring brain function.

\clearpage

\section{Introduction}
\label{sec:introduction}

With the proliferation of high dimensional neuronal population recordings, dimensionality reduction has become a widely used class of multivariate statistical techniques \citep{cunningham_dimensionality_2014}.
Increasingly, these recordings capture not only many neurons but many distinct neuronal populations, spanning brain areas, cortical layers, and cell types \citep{ahrens_brain-wide_2012, yang_vivo_2017, steinmetz_neuropixels_2021}.
Interest has grown, therefore, in dimensionality reduction methods capable of characterizing the interactions among these recorded populations \citep{semedo_statistical_2020,kang_approaches_2020,keeley_modeling_2020,machado_multiregion_2022}.

Recent development efforts have focused, in particular, on methods that characterize multi-population, or multi-group, interactions over time.
These approaches incorporate latent variables, or factors, that follow a time series model, based typically on a dynamical system (state space model; \citealp{semedo_extracting_2014, glaser_recurrent_2020, bong_latent_2020, karniol-tambour_modeling_2024}) or a Gaussian process (GP; \citealp{keeley_identifying_2020, gokcen_disentangling_2022, balzani_probabilistic_2023, li_multi-region_2024, gondur_multi-modal_2024}).
GP-based approaches can be especially useful for exploratory analyses of neural recordings, where an appropriate parametric dynamical model is unknown \textit{a priori}.

The delayed latents across groups (DLAG) framework \citep{gokcen_disentangling_2022} is a representative member of this class of multi-group GP factor models.
DLAG leverages latent variables with time delays to disentangle the bidirectional, concurrent interactions between two neuronal populations.
DLAG's recent extension to more than two populations, mDLAG \citep{gokcen_uncovering_2023}, determines (1) the subset of populations described by each latent dimension, (2) the direction of signal flow among those populations, and (3) how those signals evolve over time within and across experimental trials.
While multi-group GP factor models like mDLAG could advance the study of concurrent signaling throughout the brain, they are ultimately limited by computational scalability.
The multi-output GP kernels \citep{alvarez_kernels_2012} at their core lead to runtimes that scale cubicly in the number of time points per experimental trial and in the number of analyzed neuronal populations, or groups.

Addressing this computational challenge is thus critical for these statistical methods to remain a practical tool for analyzing recordings that already span dozens of brain areas \citep[e.g.,][]{steinmetz_distributed_2019}, and continue to grow in scale.
Here, we develop two approximate methods to accelerate the fitting of multi-group GP factor models (specifically, mDLAG) based on (1) inducing variables and (2) the frequency domain.
To our knowledge, these methods are the first for this class of models to achieve linear scaling in both trial length and group number.
This work thus enables a broad class of dimensionality reduction techniques to keep pace with the rapidly growing scale of multi-population neural recordings.

We begin with a synthesis of the diverse array of existing approaches to accelerating GP inference and estimation (Section \ref{sec:related-work}).
Then we provide a mathematical overview of the core methods considered and developed in this work: the baseline mDLAG formulation (Section \ref{sec:mdlag-time}) and the two accelerated methods, mDLAG with inducing variables (Section \ref{sec:mdlag-induce}) and mDLAG in the frequency domain (Section \ref{sec:mdlag-freq}).
We demonstrate empirically, in simulation and on neural recordings of hundreds of neurons across three brain areas, that both accelerated methods achieve orders of magnitude speed-up over baseline with minimal impact on statistical performance (Sections \ref{sec:demo}--\ref{sec:npx}).
The frequency domain approach, in particular, consistently provided the greatest runtime benefits with the fewest trade-offs in statistical performance.
We further characterize the estimation biases introduced by the frequency domain approach, and demonstrate straightforward but effective strategies to mitigate those biases (Section \ref{sec:bias}).

\section{Related Work: Fast Methods for Gaussian Processes}
\label{sec:related-work}

The computational challenges associated with GP posterior inference and \\(hyper)parameter estimation are particularly pronounced when GPs are used in latent variable models.
Model fitting is iterative, and GP inference steps therefore require hundreds to thousands of re-evaluations.
In turn, the potential runtime benefits are that much more pronounced for methods that accelerate those computationally intensive steps.

Methods for fast GP inference have an extensive history, spanning several decades and disciplines.
Much of that work has origins in GP regression \citep[Chapter 8]{rasmussen_gaussian_2006}, without any appeal to the class of multi-group latent variable models described above.
In Sections \ref{sec:related-work-matrix}--\ref{sec:related-work-frequency}, below, we review a selection of approaches relevant to the scope of the present work: multi-output GPs for time series (i.e., GPs with one input dimension), as an ingredient in latent variable models.

\subsection{Numerical Methods that Exploit Matrix Structure}
\label{sec:related-work-matrix}

In specific applications of single-group latent variable models (and single-output GPs), the cubic scaling in the number of time points, $T$, can be alleviated by exploiting matrix structure.
Stationary GP kernels, for instance (a standard modeling choice), give rise to a GP covariance matrix and its derivatives (the source of computational bottlenecks in GP inference and parameter estimation) with Toeplitz structure.
Certain computations involving Toeplitz matrices (matrix inversion, matrix products) can be implemented in $\mathcal{O}(T^2)$ operations \citep{golub_matrix_2013} and even $\mathcal{O}(T)$ storage \citep{zhang_time-series_2005}.
The fast Fourier transform (FFT) can also be leveraged to reduce Toeplitz matrix products to $\mathcal{O}(T \log T)$ operations \citep{silverman_kernel_1982}.
Many GP-based methods thus seek out advantageous matrix structure (Toeplitz, Kronecker, etc.), through careful model structural choices or a series of \textit{ad hoc} approximations during model fitting \citep{cunningham_fast_2008, aoi_scalable_2017, foreman-mackey_fast_2017, jensen_scalable_2021}.
These bespoke approaches are not as broadly applicable as the approaches that follow, particularly in the multi-group context.

\subsection{State Space Representations}
\label{sec:related-work-statespace}

The first more general class of approaches to GP acceleration exploits the deep mathematical relationship between GPs and state space models (i.e., linear dynamical systems).
Any well-behaved stationary GP time series can be approximated by an appropriate linear-Gaussian state space model with a sufficiently expanded state dimension, $p$ \citep{loper_general_2021}.
Many paths can be taken to parameterize a GP as a state space model.
Approaches that rely on spectral factorization are particularly well known \citep{sayed_survey_2001, hartikainen_kalman_2010}, and have also been extended to the multi-output case \citep{li_multi-region_2024}.
Alternative parameterizations can be obtained via the Latent Exponentially Generated (LEG) family of state space models \citep{loper_general_2021}, or regression between the state space model parameters and the GP covariance function (kernel) itself \citep{li_markovian_2024}.
\citet{li_multi-region_2024} and \citet{li_markovian_2024} have applied this state space representation directly to DLAG.
In a slightly different approach, \citet{dowling_linear_2023} arrive at an arbitrarily expressive state space representation by leveraging the Hida-Mat\'ern class of GP covariance functions.

Regardless of the path one takes to a state space formulation of a GP, the goal is the same: inference of the GP posterior can be carried out via dynamic programming, most commonly the Kalman filtering \citep{kalman_new_1961} and Rauch–Tung–Striebel smoothing \citep{rauch_maximum_1965} algorithms or information filtering \citep{dowling_linear_2023}.
The complexity of posterior inference is then reduced from $\mathcal{O}(T^3)$ operations and $\mathcal{O}(T^2)$ storage to $\mathcal{O}(p^3 T)$ operations and $\mathcal{O}(p^2 T)$ storage: linear in the number of time points per trial, $T$, at the added cost of cubic scaling in the number of expanded states in the state space approximation, $p$.
In practice however, it appears that common GP covariance functions are well approximated by only a handful of terms, $p$ \citep{hartikainen_kalman_2010, solin_explicit_2014, loper_general_2021, li_markovian_2024}, and hence the added cost is effectively a modest constant factor.
Additional approximations can ameliorate the cubic scaling in the number of state dimensions to quadratic, in both the single-output GP \citep{solin_infinite-horizon_2018} and multi-output GP cases \citep{lim_multi-output_2021}.
As an ingredient in multi-group GP factor models, however, the state space approach still leads to cubic scaling in the number of groups $M$---i.e., $\mathcal{O}(p^3 M^3 T)$ operations \citep{li_markovian_2024}---where we would like to achieve linear scaling, if possible.

\subsection{Sparsity via Inducing Variables}
\label{sec:related-work-inducing}

The second general class of approaches relies on sparse approximations, specifically via inducing variables \citep{quinonero-candela_unifying_2005}.
Inducing variable methods leverage a straightforward insight: for sufficiently smooth signals, perhaps only some smaller number of samples, $T_{\text{ind}} < T$, are needed to carry out accurate inference.
In a useful mathematical abstraction, these $T_{\text{ind}}$ inducing variables need not belong to the set of $T$ original samples, but can instead be defined at arbitrary inducing points or ``pseudo-inputs'' \citep{snelson_sparse_2005} in the same domain as that of the original samples (time, for time series data).

These approaches pass the computational burden to the smaller set of inducing points: $\mathcal{O}(T T_{\text{ind}}^2)$ operations and $\mathcal{O}(T T_{\text{ind}})$ storage---linear in the original number of samples (see also \citealp{hensman_gaussian_2013} for a formulation that achieves $\mathcal{O}(T_{\text{ind}}^3)$ operations).
\citet{titsias_variational_2009} developed a variational formulation of inducing variables that facilitated their use in extended versions of the single-group dimensionality reduction method Gaussian process factor analysis (GPFA; \citealp{yu_gaussian-process_2009, duncker_temporal_2018}).
\citet{alvarez_efficient_2010} extended this approach to multi-output GP regression, suggesting that, as an ingredient in multi-group GP factor models, inducing variables could produce $\mathcal{O}(M T T_{\text{ind}}^2)$ scaling---linear in both the number of time points $T$ and the number of groups $M$.
To date, the efficacy of assimilating inducing variables into multi-group GP factor models has not been explored.
Here, we develop an mDLAG model with inducing variables (mDLAG-inducing) that achieves the desired linear scaling properties.

\subsection{Frequency Domain Representations}
\label{sec:related-work-frequency}

The final class of approaches invokes the frequency domain (Fourier basis) representation of stationary GPs (these ideas can be combined with inducing variable approaches with some success, see for example \citealp{hensman_variational_2018}).
In the large $T$ limit, any well-behaved stationary GP can be represented as a spectral process with independent frequency components (\citealp{kolmogorov_stationary_1941}; \citealp[Section 4.11]{priestley_spectral_1981}).
In practical terms, this theoretical result suggests that, if we work with a frequency domain representation of observed time series, we can perform approximate GP inference and parameter estimation with a diagonal GP covariance matrix, rather than the dense GP covariance matrix in the time domain (see Section~\ref{sec:mdlag-freq} for details).
Computations involving this diagonal matrix could thus scale linearly in the number of time points, $T$, or at least $\mathcal{O}(T \log T)$ if a FFT of the data need be calculated first.
In spectral estimation, this approach dates back at least to \citet{whittle_hypothesis_1951}.
The Whittle likelihood (a ``quasi-likelihood'') is a biased approximation to Gaussian likelihoods of time domain signals, computed efficiently from the periodogram of the data \citep{rao_reconciling_2021, sykulski_debiased_2019}.

More recently, frequency domain GP representations have been leveraged for runtime benefits in several latent variable modeling applications.
Applications include Bayesian smoothing for spatial modeling \citep{paciorek_bayesian_2007}, modeling of natural sounds \citep[Section 3.2]{turner_statistical_2010}, and certain single-group GP factor models for neuroscience.
For example, \citet{keeley_identifying_2020} proposed a frequency domain quasi-likelihood for a GPFA model with Poisson observation noise, enabling $\mathcal{O}(T \log T)$ scaling.
Similar ideas have been incorporated into multi-modal GP variational autoencoders \citep{gondur_multi-modal_2024}.
\citet{dowling_linear_2023} explicitly incorporated the Whittle likelihood into GPFA-related models as a subroutine of fitting, to accelerate GP hyperpameter updates (like GP inference, na\"ively $\mathcal{O}(T^3)$ operations) to $\mathcal{O}(T \log T)$ operations.

These studies were limited to GP factor models that employed single-output GPs.
The cross-spectral factor analysis method of \citet{gallagher_cross-spectral_2017}, on the other hand, employed multi-output GPs to describe interactions across univariate recording channels (analogously, single neurons).
Drawing directly from \citet{ulrich_gp_2015}, \citet{gallagher_cross-spectral_2017} employed a frequency domain quasi-likelihood that produced linear scaling in $T$, but still maintained cubic scaling in the number of recording channels (analogously, $\mathcal{O}(M^3 T)$ operations for $M$ neuronal populations, or groups).
Here, we develop an approximate formulation of mDLAG (mDLAG-frequency) fully in the frequency domain, from the generative model to posterior inference and fitting.
By performing posterior inference and fitting completely in the frequency domain, the approach effectively achieves linear scaling in both the number of time points per trial and the number of groups, $\mathcal{O}(MT)$.

\section{Model and Algorithmic Overviews}
\label{sec:models}

\subsection{Delayed Latents Across Multiple Groups (mDLAG-time)}
\label{sec:mdlag-time}

We start with an overview of the mDLAG model and fitting procedure from \citet{gokcen_uncovering_2023}, which will serve as the baseline for the other two approaches developed here (Sections \ref{sec:mdlag-induce} and \ref{sec:mdlag-freq}).
Sections \ref{sec:mdlag-induce} and \ref{sec:mdlag-freq} then constitute the core contributions of this work.
Since the baseline mDLAG approach relies fully on the time domain (without invoking inducing variables), we will refer to it as mDLAG-time, to disambiguate it from those that follow.
Note that we will explicitly define all variables and parameters as they appear, but for reference, we include an explanation of notation and a glossary of common variables and parameters in Appendix~\ref{app:notation}, \apptbl{notation-shape} -- \apptbl{notation-freq}.

In brief, analyzing multi-population neural recordings presents two key challenges: (1) identifying network-level interactions and (2) disentangling concurrent signal flow.
These interactions are difficult to unravel from raw neural activity but can be pinpointed along certain latent dimensions.
mDLAG leverages that insight, and addresses each of the two challenges with a dedicated model component.
For the first, mDLAG employs automatic relevance determination (ARD) to promote group-wise sparsity for each estimated latent dimension and identify the groups involved in each interaction (Section \ref{sec:mdlag-time_obs}).
For the second, mDLAG recognizes that communication between neuronal populations is not instantaneous, and estimates for each latent variable a set of time delays that describes the signal flow across the involved groups (Section \ref{sec:mdlag-time_state}).

\subsubsection{Observation Model and Automatic Relevance Determination}
\label{sec:mdlag-time_obs}

For group $m$ (comprising $q_m$ units) at time $t$ on trial $n$, we define a linear relationship between observed activity, $\mathbf{y}^m_{n,t} \in \mathbb{R}^{q_m}$, and latent variables (latents), $\mathbf{x}^m_{n,t} \in \mathbb{R}^p$ (\fig{method_intro}{a}):
\begin{align}
    \mathbf{y}^m_{n,t} = C^m \mathbf{x}^m_{n,t} + \mathbf{d}^m + \boldsymbol{\varepsilon}^m \label{eq:mdlag_obs1} \\
    \boldsymbol{\varepsilon}^m \sim \mathcal{N}(\mathbf{0}, (\Phi^m)^{-1}) \label{eq:mdlag_obs2}
\end{align}
where the loading matrix $C^m \in \mathbb{R}^{q_m \times p}$ and mean parameter $\mathbf{d}^m \in \mathbb{R}^{q_m}$ are model parameters.
The vector $\boldsymbol{\varepsilon}^m$ is a zero-mean Gaussian random variable with noise precision matrix $\Phi^m \in \mathbb{S}^{{q_m \times q_m}}$ ($\mathbb{S}^{{q_m \times q_m}}$ is the set of $q_m \times q_m$ symmetric matrices).
We constrain the precision matrix $\Phi^m = \text{diag}(\phi^m_1,\ldots,\phi^m_{q_m})$ to be diagonal to capture variance that is independent to each unit.
This constraint encourages the latents to explain as much of the shared variance among units as possible.
As we will describe, at time point $t$, latents $\mathbf{x}^m_{n,t}$, $m = 1,\ldots,M$ are coupled across groups, and thus each group has the same number of latents, $p$.
Because we seek a low-dimensional description of observed activity, the number of latents is less than the total number of units, i.e., $p < q$, where $q = \sum_m q_m$.

mDLAG seeks to identify multiple network-level interactions across the observed groups.
To do so requires estimating the number of latents across all groups, and which subset of groups each latent involves.
This estimation problem needs to scale tractably with the number of groups.
mDLAG therefore takes a Bayesian approach to the problem, where $\mathbf{d}^m$, $\Phi^m$, and $C^m$ are taken to be probabilistic parameters with prior distributions.

\clearpage

\mainfigure{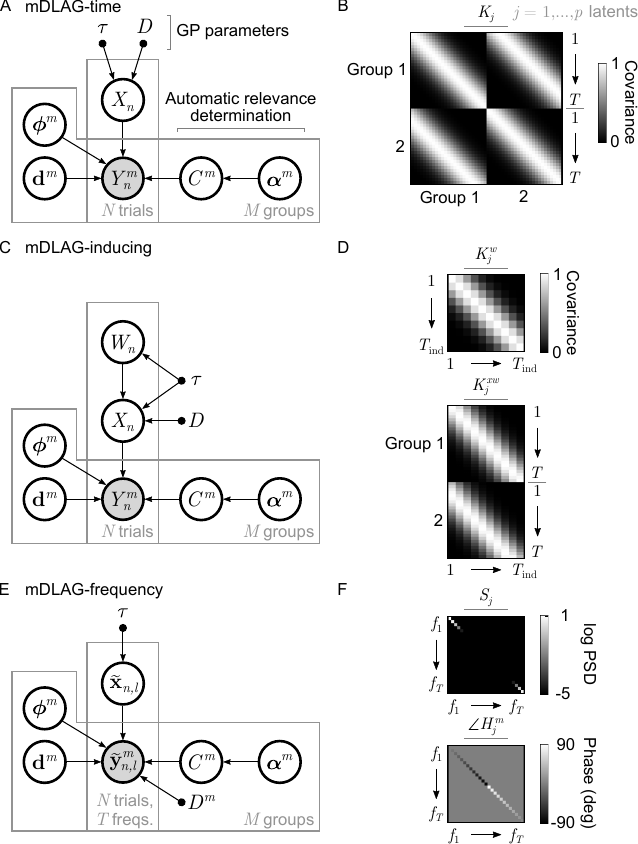}
\maincaption{method_intro}{Summary of generative models and covariance structures.}{\panel{a} Delayed latents across multiple groups (mDLAG) in the time domain (mDLAG-time; \citealp{gokcen_uncovering_2023}).
Filled circles represent observed variables.
Unfilled circles represent probabilistic latent variables and parameters.
Black dots represent deterministic parameters.
Arrows indicate conditional dependence relationships between variables.
Boxes indicate repetition of the enclosed variables or parameters over a particular index ($n = 1,\ldots,N$ trials or $m = 1,\ldots,M$ groups), where those repetitions are mutually independent.
\panel{b} An example GP covariance matrix ($K_j$, for a latent $j = 1,\ldots,p$), corresponding to the mDLAG-time model.
The matrix $K_j$ was generated with squared exponential GP timescale $\tau = 100$ ms and time delay $D = 80$ ms, for trial length $T = 25$ time points (with 20 ms sampling period) and $M = 2$ groups.
\panel{c} mDLAG with inducing variables (mDLAG-inducing).
Same conventions as in panel \panel{a}.
\panel{d} Top: An example inducing variable covariance matrix ($K^w_j$). Bottom: Corresponding covariance matrix between latent $j$ and its inducing variable ($K^{xw}_j$).
Both $K^w_j$ and $K^{xw}_j$ were generated with the same GP parameters as $K_j$ in panel \panel{a}, with $T_{\text{ind}} = 10$ inducing points.
\panel{e} mDLAG in the frequency domain (mDLAG-frequency).
Same conventions as in panel \panel{a}.
Note the box indicating independence across not only trials but also $l = 1,\ldots,T$ frequency components (freqs.). 
\panel{f} Top: An example (diagonal) GP power spectral density (PSD) matrix ($S_j$, displayed on a log scale). Bottom: An example (diagonal) phase-shift matrix from latent $j$ to an observed group $m$ ($H^m_j$, displayed in degrees, $\angle H^m_j$).
The PSD matrix $S_j$ was generated with GP timescale $\tau = 100$ ms, for trial length $T = 25$ time points.
The phase-shift matrix $H^m_j$ was generated with time delay $D^m_j = 10$ ms.
Frequency $f_1 = 0$ is the zero frequency, frequencies $f_2$ through $f_{13}$ are increasing positive frequencies, and frequencies $f_{14}$ through $f_{25}$ ($f_{T}$) are negative frequencies increasing toward zero (see Section~\ref{sec:mdlag-freq_state}).}

\vspace{22pt}
The parameter $\mathbf{d}^m$ describes the mean activity of each unit over time and trials.
We set a Gaussian prior over $\mathbf{d}^m$:
\begin{equation}
    P(\mathbf{d}^m) = \mathcal{N}(\mathbf{d}^m \ | \ \mathbf{0}, \beta^{-1} I_{q_m}) \label{eq:dprior}
\end{equation}
where $\beta \in \mathbb{R}_{>0}$ is a hyperparameter, and $I_{q_m}$ is the $q_m \times q_m$ identity matrix.
We set the conjugate Gamma prior over each $\phi^m_r$, for each unit $r = 1,\ldots,q_m$:
\begin{equation}
    P(\phi^m_r) = \Gamma(\phi^m_r \ | \ a_{\phi}, b_{\phi}) \label{eq:phiprior}
\end{equation}
where $a_{\phi}, b_{\phi}\in \mathbb{R}_{>0}$ are hyperparameters.

The loading matrix $C^m$ linearly combines latents and maps them to observed neural activity.
In particular, the $j$\textsuperscript{th} column of $C^m$, $\mathbf{c}_j^m \in \mathbb{R}^{q_m}$, maps the $j$\textsuperscript{th} latent $x^m_{n,j,t}$ to group $m$.
To determine which subset of groups is described by each latent, mDLAG employs ARD \citep{mackay_bayesian_1994, neal_bayesian_1995}.
Specifically, we define the following prior over the columns of each $C^m$ \citep{klami_group_2015}:
\begin{align}
    &P(\mathbf{c}_j^m \ | \ \alpha_j^m) = \mathcal{N}(\mathbf{c}_j^m \ | \ \mathbf{0}, (\alpha_j^m)^{-1} I_{q_m}) \label{eq:Cprior}\\
    &P(\alpha_j^m) = \Gamma(\alpha_j^m \ | \ a_{\alpha}, b_{\alpha}) \label{eq:alphaprior}
\end{align}
where $\alpha_j^m \in \mathbb{R}_{>0}$ is the ARD parameter for latent $j$ and group $m$, and $a_{\alpha}, b_{\alpha}\in \mathbb{R}_{>0}$ are hyperparameters.
As $\alpha_j^m$ becomes large, the magnitude of $\mathbf{c}_j^m$ becomes concentrated around $0$, and hence the $j$\textsuperscript{th} latent $x^m_{n,j,t}$ will have a vanishing influence on group $m$.
The ARD prior encourages group-wise sparsity for each latent during model fitting, where the loading matrix coefficients will be pushed toward zero for latents that explain an insignificant amount of shared variance within a group, and remain non-zero otherwise.
Intuitively, dimensions that appear in two or more groups indicate the presence of correlated activity across those groups.

\subsubsection{Gaussian Process State Model}
\label{sec:mdlag-time_state}

For each latent, mDLAG seeks to characterize the direction of signal flow among the involved groups (determined by ARD) and how those signals evolve over time within and across trials.
We therefore employ GPs \citep{rasmussen_gaussian_2006}, and define a GP over all time points $t = 1,\ldots,T$ for each latent $j = 1,\ldots,p$ as follows (\fig{method_intro}{b}):
\begin{equation} \label{eq:xprior_time}
    \begin{bmatrix}
    \mathbf{x}^1_{n,j,:} \\
    \vdots \\
    \mathbf{x}^M_{n,j,:}
    \end{bmatrix} \sim \mathcal{N}\left(\mathbf{0}, \begin{bmatrix} K_{1,1,j} & \cdots & K_{1,M,j} \\
    \vdots & \ddots & \vdots \\
    K_{M,1,j} & \cdots & K_{M,M,j}\end{bmatrix}\right)
\end{equation}
Under equation $\ref{eq:xprior_time}$, latents are independent and identically distributed across trials.
The diagonal blocks $K_{1,1,j} = \cdots = K_{M,M,j} \in \mathbb{S}^{T \times T}$ describe the autocovariance of each latent, and each $T$-by-$T$ off-diagonal block describes the cross-covariance that couples two groups.

To define these matrices, we introduce additional notation.
Specifically, we indicate groups with two subscripts, $m_1 = 1,\ldots,M$ and $m_2 = 1,\ldots,M$.
Then, we define $K_{m_1,m_2,j} \in \mathbb{R}^{T \times T}$ to be either the auto- or cross-covariance matrix between latent $\mathbf{x}^{m_1}_{n,j,:} \in \mathbb{R}^{T}$ in group $m_1$ and latent $\mathbf{x}^{m_2}_{n,j,:} \in \mathbb{R}^T$ in group $m_2$ on trial $n$.
mDLAG is immediately compatible with any stationary covariance function.
Here, we explore in depth the commonly used squared exponential function for GP covariances.
Therefore, element $(t_1, t_2)$ of each $K_{m_1,m_2,j}$ can be computed as follows \citep{lakshmanan_extracting_2015,gokcen_disentangling_2022}:
\begin{align}
    k_{m_1,m_2,j}(t_1,t_2) &= \left(1 - \sigma_j^2\right) \exp \Biggl(-\frac{\left(\Delta t\right)^2}{2 \tau_j^2}\Biggr) + \sigma_j^2 \cdot \delta_{\Delta t} \label{eq:k} \\
    \Delta t &= \left(t_2 - D^{m_2}_j\right) - \left(t_1 - D^{m_1}_j\right) \label{eq:delta_k}
\end{align}
where the characteristic timescale, $\tau_j \in \mathbb{R}_{>0}$ is a deterministic model parameter to be estimated from observed activity.
The GP noise variance, $\sigma_j^2 \in (0,1)$, is conventionally fixed to a small value ($10^{-3}$).
$\delta_{\Delta t}$ is the kronecker delta, which is $1$ for $\Delta t = 0$ and $0$ otherwise.
The GP is normalized so that $k_{m_1,m_2,j}(t_1,t_2) = 1$ if $\Delta t = 0$, thereby removing model redundancy in the scaling of the latents and loading matrices $C^m$.

Two parameters, the time delay to group $m_1$, $D^{m_1}_{j} \in \mathbb{R}$, and the time delay to group $m_2$, $D^{m_2}_{j} \in \mathbb{R}$, are key to describing signal flow across groups.
First notice that, when computing the auto-covariance for group $m$ (i.e., when $m_1 = m_2 = m$; \fig{method_intro}{b}, diagonal blocks of $K_j$), the time delay parameters $D^{m_1}_{j}$ and $D^{m_2}_{j}$ are equal, and so $\Delta t$ (equation \ref{eq:delta_k}) reduces simply to the time difference $(t_2 - t_1)$.
Time delays are therefore only relevant when computing the cross-covariance between distinct groups $m_1$ and $m_2$.
The time delay to group $m_1$, $D^{m_1}_{j}$, and the time delay to group $m_2$, $D^{m_2}_{j}$, by themselves have no physically meaningful interpretation.
Their difference $D^{m_2}_{j} - D^{m_1}_{j}$, however, represents a well-defined, continuous-valued time delay from group $m_1$ to group $m_2$ (\fig{method_intro}{b}, off-diagonal blocks of $K_j$).
The sign of the relative time delay indicates the directionality of the lead-lag relationship between groups captured by latent $j$ (positive: group $m_1$ leads group $m_2$; negative: group $m_2$ leads group $m_1$), which we interpret as a description of signal flow.
Note that time delays need not be integer multiples of the sampling period or spike count bin width of recorded neural activity.
Without loss of generality, we designate group $m = 1$ as the reference area, and fix the delays for group $1$ at $0$ (i.e., $D^{1}_{j} = 0$ for all latents $j = 1,\ldots,p$).

\subsubsection{Posterior Inference, Fitting, and Computational Scaling}
\label{sec:mdlag-time_fit}

Let $Y$ and $X$ be collections of all observed and latent variables, respectively, across all time points and trials.
Similarly, let $\mathbf{d}$, $\boldsymbol{\phi}$, $C$, $\mathcal{A}$, $\tau$, and $D$ be collections of the mean parameters, noise precisions, loading matrices, ARD parameters, GP timescales, and time delays, respectively.
From the observed activity, we seek to estimate posterior distributions over the probabilistic model components $\theta = \left\{X, \ \mathbf{d}, \ \boldsymbol{\phi},\ C, \ \mathcal{A}\right\}$ and point estimates of the deterministic GP parameters $\Omega = \left\{\tau, \ D \right\}$.

We do so by employing a variational inference scheme \citep{bishop_variational_1999,klami_group_2015}, in which we maximize the evidence lower bound (ELBO), $L(Q,\Omega)$, with respect to the approximate posterior distribution $Q(\theta)$ and the deterministic parameters $\Omega$, where
\begin{equation}
    \log P(Y) \geq L(Q,\Omega) = \mathbb{E}_Q[\log P(Y,\theta | \Omega)] - \mathbb{E}_Q[\log Q(\theta)] \label{eq:lb_time}
\end{equation}
We constrain $Q(\theta)$ so that it factorizes over the elements of $\theta$:
\begin{equation}
    Q(\theta) = Q_x(X) Q_d(\mathbf{d}) Q_{\phi}(\boldsymbol{\phi}) Q_c(C) Q_{\mathcal{A}}(\mathcal{A}) \label{eq:q_time}
\end{equation}
This factorization enables closed-form updates during optimization.
The ELBO can then be iteratively maximized via coordinate ascent of the factors of $Q(\theta)$ and the deterministic parameters $\Omega$. 
Here all hyperparameters were fixed to a very small value, $\beta, a_{\phi}, b_{\phi}, a_{\alpha}, b_{\alpha} = 10^{-12}$, to produce noninformative priors \citep{klami_group_2015}.
Throughout this work, we take estimates of the latent variables and model parameters to be the corresponding means of the posterior distributions comprising equation \ref{eq:q_time}.

The computational bottlenecks in mDLAG-time comprise three stages of the fitting procedure: (1) the update of the posterior distribution over latents, $Q_x(X)$, (2) the updates of the GP parameters, $\Omega$, via gradient ascent, and (3) evaluation of the ELBO, equation~\ref{eq:lb_time}.
From equation \ref{eq:xprior_time}, let
\begin{equation} \label{eq:Kj}
    K_j = \begin{bmatrix} K_{1,1,j} & \cdots & K_{1,M,j} \\
    \vdots & \ddots & \vdots \\
    K_{M,1,j} & \cdots & K_{M,M,j}\end{bmatrix} \in \mathbb{S}^{MT \times MT}
\end{equation}
Each of the three stages above require the inversion of the $MT \times MT$ covariance matrix $K_j$, and therefore require at least $\mathcal{O}(M^3 T^3)$ operations and $\mathcal{O}(M^2 T^2)$ storage.
The update of $Q_x(X)$, in fact, requires the inversion of a larger $pMT \times pMT$ matrix (see below).
The iterative nature of gradient ascent also requires potentially several re-evaluations of $K_j^{-1}$ within a single fitting iteration.

A complete set of update equations, including for the three stages above, can be found in \citet{gokcen_uncovering_2023}.
For the sake of concision, we highlight just the update of the posterior distribution over latents, $Q_x(X)$.
Let us first define several variables.
Construct $\mathbf{y}_{n,t} = [\mathbf{y}^{1\top}_{n,t} \cdots \mathbf{y}^{M\top}_{n,t}]^{\top} \in \mathbb{R}^q$, $q = \sum_m q_m$, by vertically concatenating the observed activity of groups $m = 1,\ldots,M$ at time $t$ on trial $n$.
Then construct $\bar{\mathbf{y}}_n = [\mathbf{y}^{\top}_{n,1} \cdots \mathbf{y}^{\top}_{n,T}]^{\top} \in \mathbb{R}^{qT}$ by vertically concatenating the observed activity $\mathbf{y}_{n,t}$ across all time points $t = 1,\ldots,T$.
For latents, define $\mathbf{x}_{n,t} = [\mathbf{x}^{1\top}_{n,:,t} \cdots \mathbf{x}^{M\top}_{n,:,t}]^{\top} \in \mathbb{R}^{pM}$ by vertically concatenating the $p$ latents of each group at time $t$ on trial $n$.
Then we vertically concatenate the latents $\mathbf{x}_{n,t}$ across all time points $t = 1,\ldots,T$ to give $\bar{\mathbf{x}}_n = [\mathbf{x}_{n,1}^{\top} \cdots \mathbf{x}_{n,T}^{\top}]^{\top} \in \mathbb{R}^{pMT}$.
Finally, we collect the parameters $C^m$, $\Phi^m$, and $\mathbf{d}^m$ across populations $m = 1,\ldots,M$ by defining $C = \text{diag}(C^1,\ldots,C^M) \in \mathbb{R}^{q \times pM}$, $\Phi = \text{diag}(\Phi^1,\ldots,\Phi^m) \in \mathbb{S}^{q \times q}$, and $\mathbf{d} = [\mathbf{d}^{1\top} \cdots \mathbf{d}^{M\top}]^{\top} \in \mathbb{R}^q$.

The update to the posterior distribution over latents, $Q_x(X)$, takes the same functional form as the prior distribution (equation \ref{eq:xprior_time}), a Gaussian distribution,
\begin{align}
    Q_x(X) &= \prod_{n=1}^N \mathcal{N}(\bar{\mathbf{x}}_n \ | \ \bar{\boldsymbol{\mu}}_{x_n}, \bar{\Sigma}_{x}) \label{eq:qx_time}
\end{align}
with trial-dependent posterior mean $\bar{\boldsymbol{\mu}}_{x_n} \in \mathbb{R}^{pMT}$ and trial-independent posterior covariance $\bar{\Sigma}_x \in \mathbb{S}^{pMT \times pMT}$.
The update for the posterior covariance is the computationally limiting step:
\begin{align}
    \bar{\Sigma}_{x} &= \bigl(\bar{K}^{-1} + \langle \overline{C^{\top} \Phi C}\rangle\bigr)^{-1} \label{eq:qx_cov_time}
\end{align}
Here we introduce the notation $\langle \cdot \rangle$ to indicate the expectation with respect to the approximate posterior distribution, $\mathbb{E}_{Q}[\cdot]$.
The term $\langle \overline{C^{\top} \Phi C}\rangle \in \mathbb{R}^{pMT \times pMT}$ is a block diagonal matrix comprising $T$ copies of the matrix $\langle C^{\top} \Phi C\rangle$.
The elements of the prior covariance matrix $\bar{K} \in \mathbb{R}^{pMT \times pMT}$ follow the structure of $\bar{x}_n$, and are computed using equations \ref{eq:k} and \ref{eq:delta_k}.
Evaluating equation~\ref{eq:qx_cov_time} on each mDLAG fitting iteration thus requires the inversion of a $pMT \times pMT$ matrix, costing $\mathcal{O}(p^3 M^3 T^3)$ operations and $\mathcal{O}(p^2 M^2 T^2)$ storage.
Ameliorating this computational cost and the remaining costs of the three bottleneck stages above is the focus of the rest of this work. 

\subsection{mDLAG with Inducing Variables (mDLAG-inducing)}
\label{sec:mdlag-induce}

The scaling challenges of mDLAG-time stem from the Gaussian process state model (equation~\ref{eq:xprior_time}), namely the $MT \times MT$ covariance matrix $K_j$ and the computation of its inverse and determinant.
Alternative formulations of the state model that mitigate or avoid these computations altogether are therefore desirable.
We first consider an approach that incorporates inducing variables \citep{titsias_variational_2009, alvarez_efficient_2010, duncker_temporal_2018}.
In essence, we might be able to improve scalability if, instead of the full set of latents across groups, we manipulate one latent representation common to all groups (thereby improving scaling with $M$) at a lower temporal resolution (thereby improving scaling with $T$).
This approach, mDLAG-inducing, requires no changes to the mDLAG-time observation model or ARD components (equations~\ref{eq:mdlag_obs1}--\ref{eq:alphaprior}), only the state model (\fig{method_intro}{c}).

\subsubsection{Gaussian Process State Model with Inducing Variables}
\label{sec:mdlag-induce_state}

For each latent $j$ on trial $n$ we define a corresponding inducing variable \\ $\mathbf{w}_{n,j,:} = [w_{n,j,1} \cdots w_{n,j,T_{\text{ind}}}]^{\top} \in \mathbb{R}^{T_{\text{ind}}}$.
Inducing variable values are defined over a set of inducing (time) points indexed by $t \in \{1,\ldots,T_{\text{ind}}\}$.
The number of inducing points per trial, $T_{\text{ind}}$, is a hyperparameter, and in this work we will choose $T_{\text{ind}}$ to be no greater than the number of time points per trial, $T$, so that $T_{\text{ind}} \leq T$ (in principle, $T_{\text{ind}}$ can also be chosen separately for each latent $j$, but we choose not to do so here).
While the number of inducing points is discrete, the times at which the inducing variables are defined are real-valued.
In detail, inducing variable $w_{n,j,t}$ at inducing point $t$, $t \in \{1,\ldots,T_{\text{ind}}\}$, is defined at time $\xi_t \in \mathbb{R}$.

These locations in time are themselves a design choice, and can even be treated as learnable model parameters, just as the GP parameters \citep{snelson_sparse_2005,titsias_variational_2009,duncker_temporal_2018}.
Throughout this work, we will fix the inducing points on a uniformly spaced grid, with the first and last inducing points fixed at the beginning and end of each trial, i.e., $\xi_1 = 1$ and $\xi_{T_{\text{ind}}} = T$.
We do not explore inducing points as learnable model parameters here to facilitate comparison with the other two methods (updating the inducing points adds a significant computational cost to the model fitting procedure), but we do include that option in accompanying code (see \nameref{sec:code}).

We then define a GP over all inducing points $\xi_1,\ldots,\xi_{T_{\text{ind}}}$ for each latent $j = 1,\ldots,p$ as follows:
\begin{equation}
    \mathbf{w}_{n,j,:} \sim \mathcal{N}\left(\mathbf{0}, K^w_j\right) \label{eq:wprior}
\end{equation}
where $K^w_j \in \mathbb{S}^{T_{\text{ind}} \times T_{\text{ind}}}$ is the inducing variable covariance matrix (\fig{method_intro}{d}, top, $K^w_j$).
Continuing with the squared exponential GP covariance function, element $(t_1, t_2)$ of $K^w_j$ is computed according to
\begin{align}
    k^w_j(\xi_{t_1},\xi_{t_2}) &= \left(1 - \sigma_j^2\right) \exp \Biggl(-\frac{\left( \Delta t \right)^2}{2 \tau_j^2}\Biggr) + \sigma_j^2 \cdot \delta_{\Delta t} \label{eq:kw} \\
    \Delta t &= \xi_{t_2} - \xi_{t_1} \label{eq:delta_kw}
\end{align}
The characteristic timescale, $\tau_j \in \mathbb{R}_{>0}$, and the GP noise variance, $\sigma_j^2 \in (0,1)$, are defined as in equation \ref{eq:k}.

We then define each latent $j$ across groups $m = 1,\ldots,M$, \\ $\mathbf{x}_{n,j,:} = [\mathbf{x}^{1\top}_{n,j,:} \cdots \mathbf{x}^{M\top}_{n,j,:}]^{\top} \in \mathbb{R}^{MT}$, in terms of the common inducing variable, $\mathbf{w}_{n,j,:}$, via the following conditional Gaussian distribution:
\begin{equation} \label{eq:xprior_induce}
    P\left(
    \mathbf{x}_{n,j,:} \ | \ \mathbf{w}_{n,j,:} \right) = \mathcal{N}\biggl(\mathbf{x}_{n,j,:} \ | \ K^{xw}_j (K^w_j)^{-1} \mathbf{w}_{n,j,:}, K_j - K^{xw}_j (K^w_j)^{-1} K^{wx}_j\biggr)
\end{equation}
where $K_j \in \mathbb{S}^{MT \times MT}$ is the covariance matrix defined in equation~\ref{eq:Kj} (\fig{method_intro}{b}), and $K^{xw}_j = (K^{wx}_j)^{\top} \in \mathbb{R}^{MT \times T_{\text{ind}}}$ is the cross-covariance matrix between latent $j$ and its inducing variable (\fig{method_intro}{d}, bottom, $K^{xw}_j$).
If we define $K^{xw}_{m,j} \in \mathbb{R}^{T \times T_{\text{ind}}}$ to be the $m$\textsuperscript{th} block of $K^{xw}_j = [K^{xw\top}_{1,j} \cdots K^{xw\top}_{M,j}]^{\top}$, then element $(t_1, t_2)$ of $K^{xw}_{m,j}$, where $t_1 \in \{1,\ldots,T\}$ and $t_2 \in \{1,\ldots,T_{\text{ind}}\}$, can be computed according to the covariance function
\begin{align}
    k^{xw}_j(t_1,\xi_{t_2}) &= \left(1 - \sigma_j^2\right) \exp \Biggl(-\frac{\left( \Delta t \right)^2}{2 \tau_j^2}\Biggr) + \sigma_j^2 \cdot \delta_{\Delta t} \label{eq:kxw} \\
    \Delta t &= \xi_{t_2} - (t_1 - D^m_j) \label{eq:delta_kxw}
\end{align}
with the same timescale parameter $\tau_j$ and GP noise variance $\sigma_j^2$ as defined as in equation \ref{eq:kw}.
The time delay parameter $D^m_j \in \mathbb{R}$ is defined in equation~\ref{eq:delta_k}.
Under this formulation, each $\mathbf{x}^{m}_{n,j,:}$ for latent $j$ and group $m$ can be viewed as an interpolated (upsampled) and time-delayed version of the common inducing variable $\mathbf{w}_{n,j,:}$.

\subsubsection{Posterior Inference, Fitting, and Computational Scaling}
\label{mdlag-induce_fit}

Let $Y$, $X$, and $W$ be collections of all observed variables, latent variables, and inducing variables, respectively, across all time points (or inducing points) and trials.
As with mDLAG-time, let $\mathbf{d}$, $\boldsymbol{\phi}$, $C$, $\mathcal{A}$, $\tau$, and $D$ be collections of the mean parameters, noise precisions, loading matrices, ARD parameters, GP timescales, and time delays, respectively.
From the observed activity, we seek to estimate posterior distributions over the probabilistic model components $\theta = \left\{X, \ W, \ \mathbf{d}, \ \boldsymbol{\phi},\ C, \ \mathcal{A}\right\}$, which now includes the inducing variables $W$, and point estimates of the deterministic GP parameters $\Omega = \left\{\tau, \ D \right\}$.

We again employ variational inference, in which we maximize the ELBO, \\ $L(Q,\Omega)$ (equation~\ref{eq:lb_time}), with respect to the approximate posterior distribution $Q(\theta)$ and the deterministic parameters $\Omega$.
For mDLAG-inducing, however, we follow \citet{titsias_variational_2009} and constrain $Q(\theta)$ so that it factorizes over the elements of $\theta$ as follows:
\begin{align}
    Q(\theta) &= Q_{xw}(X,W) Q_d(\mathbf{d}) Q_{\phi}(\boldsymbol{\phi}) Q_c(C) Q_{\mathcal{A}}(\mathcal{A}) \\
    &= P(X | W) Q_w(W) Q_d(\mathbf{d}) Q_{\phi}(\boldsymbol{\phi}) Q_c(C) Q_{\mathcal{A}}(\mathcal{A}) \label{eq:q_induce_main}
\end{align}
Here we have constrained the joint approximate posterior distribution over the latents and their inducing variables to factorize as $Q_{xw}(X,W) = P(X | W) Q_w(W)$, with generic distribution $Q_w(W)$ over the inducing variables and the conditional prior distribution $P(X | W)$ over the latents themselves.
From equation~\ref{eq:xprior_induce}, \\ $P(X | W) = \prod_{j} \prod_{n} P(\mathbf{x}_{n,j,:} \ | \ \mathbf{w}_{n,j,:})$.
Then as with mDLAG-time, this factorization enables closed-form updates during optimization.
The ELBO can be iteratively maximized via coordinate ascent of the factors of $Q(\theta)$ and the deterministic parameters $\Omega$. 

As with mDLAG-time in Section~\ref{sec:mdlag-time_fit}, the three key stages of the mDLAG-inducing fitting procedure are (1) the update of the posterior distribution over the latents and their inducing variables, $Q_{xw}(X,W)$, (2) the updates of the GP parameters, $\Omega$, via gradient ascent, and (3) evaluation of the variational lower bound, equation~\ref{eq:lb_time}.
A complete set of update equations, including for these three key stages, are provided in Appendix~\ref{app:mdlag-induce_fit}.
At all three stages, mDLAG-inducing achieves linear scaling in both the number of groups $M$ and the number of time points per trial $T$, albeit with superlinear scaling in the number of inducing points $T_{\text{ind}}$.
For example, the limiting computation in the update of $Q_{xw}(X,W)$ costs  $\mathcal{O}(p^3 M T T_{\text{ind}}^2)$ operations (see Appendix~\ref{app:mdlag-induce_fit}).

\subsection{mDLAG in the Frequency Domain (mDLAG-frequency)}
\label{sec:mdlag-freq}

Consider now the unitary discrete Fourier transform (DFT) of the time series of observations for unit $r$ in group $m$ on trial $n$: $\widetilde{\mathbf{y}}^m_{n,r,:} = U_T \mathbf{y}^m_{n,r,:}$, where $U_T \in \mathbb{C}^{T \times T}$ is the unitary DFT matrix.
Our goal is to develop a generative model for these observations while remaining entirely in the frequency domain (mDLAG-frequency, \fig{method_intro}{e}), with the hope of uncovering computational benefits.
Note that the conversion of each unit's activity from the time domain to the frequency domain has time complexity $\mathcal{O}(T \log T)$, assuming the DFT is implemented using a fast Fourier transform (FFT) algorithm.
Developing a generative model entirely in the frequency domain, and thereby procedures for posterior inference and fitting entirely in the frequency domain, requires this computation only once, as a preprocessing step.
Then, as we will show, each iteration of the fitting procedure will scale linearly in $T$ and $M$.

\subsubsection{Gaussian Process State Model}
\label{sec:mdlag-freq_state}

For the frequency domain to provide those computational benefits, it must improve the scalability of the limiting mDLAG-time model component, the GP state model.
We first consider the case with $M = 1$ group.
Taking the unitary DFT of the time course of latent $j$ on trial $n$, $\mathbf{x}_{n,j,:}$, we get
\begin{equation} \label{eq:xprior_dft}
    U_T \mathbf{x}_{n,j,:} \sim \mathcal{N}\left(\mathbf{0}, U_T K_j U_T^{\mathsf{H}}\right)
\end{equation}
a linear transformation of the state model ($U_T^{\mathsf{H}}$ is the conjugate transpose of $U_T$).
The key quantity in equation \ref{eq:xprior_dft} is the transformed covariance matrix $\widetilde{K}_j = U_T K_j U_T^{\mathsf{H}}$.
Unfortunately, like the time domain covariance matrix $K_j$, the frequency domain covariance matrix $\widetilde{K}_j$ is, in general, dense.
Consequently, any fitting procedure based on this state model will still scale cubicly in the number of time points per trial.
We will get no computational benefit over the bottleneck mDLAG-time updates (Section~\ref{sec:mdlag-time_fit}).

But critically, for large $T$ (and for a stationary GP), the transformed covariance matrix $\widetilde{K}_j$ approaches a diagonal matrix, a well-established result in spectral analysis (\citealp{kolmogorov_stationary_1941}; \citealp[Section 4.11]{priestley_spectral_1981}).
The diagonal elements of that limiting matrix are given by the power spectral density (PSD) of the Gaussian process, a function of frequency.
By the Wiener–Khinchin theorem, the PSD function of latent $j$, $s_j$, is given by the continuous-time Fourier transform of its covariance function, $k_j$ (treated as a function of the time difference $t_2 - t_1$).
For the squared exponential covariance function in equation \ref{eq:k}, the PSD function takes the following closed form:
\begin{align}
    s_j(f_{l}) &= \left(1 - \sigma_j^2\right) \sqrt{2 \pi} \tau_j \exp \Biggl(-\frac{1}{2}\left(2 \pi f_{l}\right)^2 \tau_j^2\Biggr) + \sigma_j^2 \label{eq:s}
\end{align}
where $f_l \in \mathbb{R}$ is a frequency value ($l$ indexes the components of a discretely sampled signal, see below).
The characteristic timescale, $\tau_j \in \mathbb{R}_{>0}$, and the GP noise variance, $\sigma_j^2 \in (0,1)$, are defined as in equation \ref{eq:k}.

A generative model for a GP with a diagonal covariance matrix would lead to significant runtime advantages.
Toward that end, let $\widetilde{\mathbf{x}}_{n,j,:} \in \mathbb{C}^T$ be the values of latent $j$ across frequencies $l = 1,\ldots,T$ on trial $n$.
Following the symmetry properties of the DFT, the index $l = 1$ corresponds to the zero frequency, $f_1 = 0$.
For even values of $T$, the indices $l = 2,\ldots,\frac{T}{2}+1$ correspond to increasing positive frequencies, where $l = \frac{T}{2}+1$ indexes the Nyquist frequency ($f_{\frac{T}{2}+1}$ is half the sampling rate of observations).
Indices $l = \frac{T}{2} + 2,\ldots,T$ correspond to negative frequencies increasing toward zero, with symmetry about the Nyquist rate such that $f_{\frac{T}{2}+2} = -f_{\frac{T}{2}}$, $f_{\frac{T}{2}+3} = -f_{\frac{T}{2}-1}$, up to $f_{T} = -f_{2}$.
Odd values of $T$ lead to similar frequency ordering but with subtle differences in bookkeeping.

Then as an approximation to the GP prior defined in equation \ref{eq:xprior_dft}, we define a (complex-valued) GP over all frequencies for each frequency domain latent $j = 1,\ldots,p$, where the covariance matrix is diagonal by construction (\fig{method_intro}{f}, top, $S_j$):
\begin{equation} \label{eq:xprior_freq}
    \widetilde{\mathbf{x}}_{n,j,:} \sim \mathcal{N}\left(\mathbf{0}, S_j \right)
\end{equation}
The $l$\textsuperscript{th} element of the diagonal ``PSD matrix'' $S_j \in \mathbb{S}^{T \times T}$ is given by the PSD function of the GP (equation \ref{eq:s}).
We explore the consequences of this approximation in Section~\ref{sec:bias}.
For now, we will continue to define the frequency domain generative model under this approximation, and extract any computational benefit that arises.

\subsubsection{Observation Model}
\label{sec:mdlag-freq_obs}

From here, we could define a linear observation model from latents to observed activity, akin to equation~\ref{eq:mdlag_obs1}.
However, we have omitted in the development of the model thus far any notion of time delays.
Let us revisit, for the multi-group ($M > 1$) case, the continuous-time Fourier transform of the cross-covariance function $k_{m_1,m_2,j}$ (equation~\ref{eq:k}) between groups $m_1$ and $m_2$:
\begin{align}
    s_{m_1,m_2,j}(f_{l}) &= \left(1 - \sigma_j^2\right) \sqrt{2 \pi} \tau_j \exp \Biggl(-\frac{1}{2}\left(2 \pi f_{l}\right)^2 \tau_j^2\Biggr) \notag\\
    &\quad \cdot \exp \Biggl(-i 2 \pi f_{l} (D^{m_1}_j - D^{m_2}_j) \Biggr) + \sigma_j^2 \label{eq:scross}
\end{align}
Here $i = \sqrt{-1}$ is the imaginary number.
This cross-spectral density (CSD) function is nearly identical in form to the PSD function of equation~\ref{eq:s}, but with the addition of a multiplicative complex exponential term, $\exp(-i 2 \pi f_{l} (D^{m_1}_j - D^{m_2}_j))$.
This term isolates the time delay parameters $D^{m_1}_j$ and $D^{m_2}_j$, representing them as a relative phase shift, $2 \pi f_l (D^{m_1}_j - D^{m_2}_j)$, at frequency $f_l$.
Note the separability of the complex exponential term by groups:
\begin{align}
    \exp \biggl(-i 2 \pi f_{l} (D^{m_1}_j - D^{m_2}_j) \biggr) &=  \exp \biggl(-i 2 \pi f_{l} D^{m_1}_j \biggr) \cdot \exp \biggl(i 2 \pi f_{l} D^{m_2}_j \biggr) \notag\\
    &= \exp \biggl(-i 2 \pi f_{l} D^{m_1}_j \biggr) \cdot \exp \biggl(-i 2 \pi f_{l} D^{m_2}_j \biggr)^* \notag\\
    &= h^{m_1}_{j,l} \cdot (h^{m_2}_{j,l})^* \label{eq:phase_sep}
\end{align}
where we have defined $h^{m}_{j,l} = \exp(-i 2 \pi f_{l} D^{m}_j)$ to be the phase shift term from latent $j$ to observed group $m$ at frequency $l$.

We can then construct a frequency domain analog, $\widetilde{\mathbf{x}}^m_{n,j,:} \in \mathbb{C}^T$, to the time delayed latent to group $m$, $\mathbf{x}^m_{n,j,:} \in \mathbb{R}^T$.
Collect the phase shift terms $h^{m}_{j,l}$ across all frequencies $l = 1,\ldots,T$ into the diagonal matrix $H^m_j = \text{diag}(h^{m}_{j,1}, \ldots h^{m}_{j,T}) \in \mathbb{C}^{T \times T}$ (\fig{method_intro}{f}, bottom, $H^m_j$).
Then we get $\widetilde{\mathbf{x}}^m_{n,j,:}$ by multiplication of the diagonal phase shift matrix $H^m_j$ with the frequency domain latent $j$ defined in equation~\ref{eq:xprior_freq}: $\widetilde{\mathbf{x}}^m_{n,j,:} = H^m_j \widetilde{\mathbf{x}}_{n,j,:}$.

We now have the conceptual basis to define a frequency domain observation model, from latents to observed activity.
For group $m$ (comprising $q_m$ units) at frequency $l$ on trial $n$, we define a linear relationship between observed activity, $\widetilde{\mathbf{y}}^m_{n,l} \in \mathbb{C}^{q_m}$, and the set of latents common to all groups, $\widetilde{\mathbf{x}}_{n,t} \in \mathbb{C}^p$ (\fig{method_intro}{e}):
\begin{align}
    \widetilde{\mathbf{y}}^m_{n,l} = C^m H_l^m \widetilde{\mathbf{x}}_{n,l} + \widetilde{\mathbf{d}}^m_l + \boldsymbol{\varepsilon}^m \label{eq:mdlag_obs1_freq} \\
    \boldsymbol{\varepsilon}^m \sim \mathcal{N}(\mathbf{0}, (\Phi^m)^{-1}) \label{eq:mdlag_obs2_freq}
\end{align}
Here the diagonal matrix $H^m_l \in \mathbb{C}^{p \times p}$ is a different collection of the phase shift terms $h^{m}_{j,l}$, across all latents $j = 1,\ldots,p$ at the single frequency $l$: $H^m_l = \text{diag}(h^{m}_{1,l}, \ldots h^{m}_{p,l})$.
The mean parameter term $\widetilde{\mathbf{d}}^m_l \in \mathbb{R}^{q_m}$ is shorthand for $\delta_{l-1} \cdot \sqrt{T} \mathbf{d}^m$, which is $\sqrt{T} \mathbf{d}^m$ for the zero frequency ($f_1 = 0$ for index $l = 1$), and $\mathbf{0}$ otherwise.

All other parameters are the same as defined in the mDLAG-time observation model (equations~\ref{eq:mdlag_obs1}--\ref{eq:mdlag_obs2}).
We further set the same prior distributions over the mean parameters $\mathbf{d}^m$ (equation~\ref{eq:dprior}) and noise precision parameters (equation~\ref{eq:phiprior}), and we maintain the ARD prior over the columns of each loading matrix $C^m$ (equations~\ref{eq:Cprior}--\ref{eq:alphaprior}).
Note that the PSD function given by equation~\ref{eq:s} is normalized so that $\int_{-\infty}^{+\infty} s_j(f) df = 1$, thereby removing---as was the case for mDLAG-time---model redundancy in the scaling of the latents and loading matrices $C^m$.

\subsubsection{Posterior Inference, Fitting, and Computational Scaling}
\label{sec:mdlag-freq_fit}

Fitting the mDLAG-frequency model proceeds analogously to fitting for mDLAG-time.
Let $\widetilde{Y}$ and $\widetilde{X}$ be collections of all observed and latent variables, respectively, across all frequencies and trials.
As before, let $\mathbf{d}$, $\boldsymbol{\phi}$, $C$, $\mathcal{A}$, $\tau$, and $D$ be collections of the mean parameters, noise precisions, loading matrices, ARD parameters, GP timescales, and time delays, respectively.
From frequency domain observations, we seek to estimate posterior distributions over the probabilistic model components \\ $\theta = \left\{\widetilde{X}, \ \mathbf{d}, \ \boldsymbol{\phi},\ C, \ \mathcal{A}\right\}$ and point estimates of the deterministic GP parameters \\ $\Omega = \left\{\tau, D \right\}$.

We again employ variational inference, in which we maximize the lower bound $\widetilde{L}(\widetilde{Q},\Omega)$ to the log quasi-likelihood $\log P(\widetilde{Y})$, with respect to the approximate posterior distribution $\widetilde{Q}(\theta)$ and the deterministic parameters $\Omega$:
\begin{equation}
    \log P(\widetilde{Y}) \geq \widetilde{L}(\widetilde{Q},\Omega) = \mathbb{E}_{\widetilde{Q}}[\log P(\widetilde{Y},\theta | \Omega)] - \mathbb{E}_{\widetilde{Q}}[\log \widetilde{Q}(\theta)] \label{eq:lb_freq}
\end{equation}
The quasi-likelihood $P(\widetilde{Y})$ is meant to approximate the time domain likelihood $P(Y)$ (equation~\ref{eq:lb_time}).
The two quantities converge as the number of time points per trial, $T$, becomes large \citep{whittle_hypothesis_1951}.

We again constrain $\widetilde{Q}(\theta)$ so that it factorizes over the elements of $\theta$:
\begin{equation}
    \widetilde{Q}(\theta) = Q_{\widetilde{x}}(\widetilde{X}) Q_d(\mathbf{d}) Q_{\phi}(\boldsymbol{\phi}) Q_c(C) Q_{\mathcal{A}}(\mathcal{A}) \label{eq:q_freq_main}
\end{equation}
In the same manner as mDLAG-time, this factorization enables closed-form updates during optimization, and the lower bound $\widetilde{L}(\widetilde{Q},\Omega)$ can be iteratively maximized via coordinate ascent of the factors of $\widetilde{Q}(\theta)$ and the deterministic parameters $\Omega$. 

As with the previous two methods, the three key stages of the mDLAG-frequency fitting procedure are (1) the update of the posterior distribution over latents, $Q_{\widetilde{x}}(\widetilde{X})$, (2) the updates of the GP parameters, $\Omega$, via gradient ascent, and (3) evaluation of the variational lower bound, equation~\ref{eq:lb_freq}.
At all three stages, mDLAG-frequency achieves linear scaling in both the number of groups $M$ and the number of time points per trial $T$.
A complete set of update equations, including for these three key stages, are provided in Appendix~\ref{app:mdlag-freq_fit}.

Here we highlight just the update of the posterior distribution over latents, $Q_{\widetilde{x}}(\widetilde{X})$.
The update to $Q_{\widetilde{x}}(\widetilde{X})$ takes the same functional form as the prior distribution (equation~\ref{eq:xprior_freq}), a Gaussian distribution,
\begin{align}
    Q_{\widetilde{x}}(\widetilde{X}) &= \prod_{n=1}^N \prod_{l=1}^T \mathcal{N}(\widetilde{\mathbf{x}}_{n,l} \ | \ \widetilde{\boldsymbol{\mu}}_{x_{n,l}}, \widetilde{\Sigma}_{x,l}) \label{eq:qx_freq}
\end{align}
with trial- and frequency-dependent posterior mean $\widetilde{\boldsymbol{\mu}}_{x_{n,l}} \in \mathbb{C}^{p}$ and frequency-dependent, but trial-independent, posterior PSD matrix $\widetilde{\Sigma}_{x,l} \in \mathbb{C}^{p \times p}$.
Note the independence in equation~\ref{eq:qx_freq} not only across trials, but also across frequencies.
This independence emerges naturally---we impose only the factorization in equation \ref{eq:q_freq_main}.
By contrast, the analogous updates for mDLAG-time are independent across trials only, not time points (equation~\ref{eq:qx_time}).
We can thus update the posterior PSD matrix, $\widetilde{\Sigma}_{x,l}$, separately for each frequency $l$:
\begin{align}
    \widetilde{\Sigma}_{x,l} &= \biggl(S_{l}^{-1} + \sum_{m=1}^M (H^m_l)^{\mathsf{H}}\langle (C^m)^{\top} \Phi^m C^m \rangle H^m_l \biggr)^{-1} \label{eq:qx_cov_freq}
\end{align}
The elements of the diagonal PSD matrix $S_l = \text{diag}(s_1(f_l), \ldots, s_p(f_l)) \in \mathbb{R}^{p \times p}$ are computed using equation~\ref{eq:s}.
The diagonal phase shift matrix $H^m_l \in \mathbb{C}^{p \times p}$ is defined as in equation~\ref{eq:mdlag_obs1_freq}.

Evaluation of equation~\ref{eq:qx_cov_freq} for all frequencies $l = 1,\ldots,T$ is significantly more efficient than the evaluation of the analogous equation~\ref{eq:qx_cov_time} for mDLAG-time.
In the first term, the PSD matrix $S_l$ is diagonal, and thus its inversion scales linearly in the number of latents, $p$.
In the second term, $H^m_l$ is diagonal, and thus the sum over groups $m = 1,\ldots,M$ costs $\mathcal{O}(p^2 M)$, and $\mathcal{O}(p^2 M T)$ in total when evaluated for all frequencies.
The final computation is the inversion of a $p \times p$ matrix, costing $\mathcal{O}(p^3 T)$ operations and $\mathcal{O}(p^2 T)$ storage, when evaluated for all frequencies.
Thus in total, posterior inference over the latents scales linearly in both the number of time points per trial $T$ and the number of groups $M$.

\section{Results}
\label{sec:results}

\subsection{Demonstration in Simulation}
\label{sec:demo}

We start with an illustrative simulation to demonstrate, across our three fitting methods, (1) their basic correctness, (2) their relative runtime performance, and (3) themes explored more deeply in subsequent sections.
Dataset characteristics are summarized in \tbl{simulated-summary} (Demo).

\clearpage
\maintablecaption{simulated-summary}{Summary of Simulated Datasets.}{}
\begin{table}[ht!]
  \footnotesize
  \caption{}
  \label{tab:simulated-summary}
  \centering
  \begin{tabular}{p{3.2cm}p{1.4cm}p{1.3cm}p{1.0cm}p{1.9cm}}
    \toprule
    Dataset     & Demo    & Scaling, $T$    & Scaling, $M$    & Model \quad \quad selection \\
    \midrule
    Independent runs    & 1    & 20    & 20    & 10 per SNR\\
    Training trials    & 100     & 100   
 & 100    & 50 \\
    Time points    & 100    & 10 -- 500    & 50    & 10 -- 200 \\
    Samp. period (ms)     & 20    & 20    & 20    & 20 \\
    Groups    & 2    & 2    & 1 -- 24    & 1 \\
    Units per group    & 10    & 12    & 24 -- 1    & 1 \\
    Latents    & 4    & 1    & 1    & 4 \\
    GP timescales (ms)    & 20 -- 120    & 100    & 100    & 50 \\
    GP time delays (ms)    & +12, -23    & 10    & 0 -- 20    & --- \\
    Signal-to-noise ratio    & 0.2    & 0.2    & 0.2    & 0.01 -- 10.0 \\
    \bottomrule
  \end{tabular}
\end{table}

We generated simulated activity from two groups ($M = 2$) according to the mDLAG generative model (equations \ref{eq:mdlag_obs1}--\ref{eq:delta_k}; \nameref{sec:methods}, Section~\ref{sec:data_generation}).
For the sake of illustration, we set 10 units in each group ($q_m = 10$).
We designed the loading matrices $C^m$ so that all types of inter-group interactions were represented (\fig{demo}{a}, left, ground truth): a feedforward interaction (A to B), a feedback interaction (B to A), and an interaction local to each group.
We also scaled the observation noise precision matrices $\Phi^m$ so that noise levels were representative of realistic neural activity.
Specifically, activity due to single-unit observation noise was 5 times stronger than activity due to latents (the ``signal-to-noise ratio'' $\text{tr}(C^m C^{m\top}) / \text{tr}((\Phi^m)^{-1}) = 0.2$ for each group).
Gaussian process timescales and time delays were also chosen within realistic ranges (\tbl{simulated-summary}, Demo).
With all model parameters specified, we then generated $N = 100$ independent and identically distributed trials.
Each trial was $2$ s in length, comprising $T = 100$ time points with a sampling period of 20 ms, or a rate of 50 Hz.

We then fit three mDLAG models to the simulated activity using each of the three fitting methods: mDLAG-time, mDLAG-inducing, and mDLAG-frequency.
For all fitting methods, we set the initial number of latents ($p = 8$) to be greater than the ground truth number ($p = 4$), to verify that these additional latents would be pruned during fitting (\nameref{sec:methods}, Section~\ref{sec:mdlag_init}).
For mDLAG-inducing we chose the number of inducing points to be half of the number of time points per trial ($T_{\text{ind}} = 50$; we will elaborate on the choice of this hyperparameter in subsequent sections).

\clearpage
\mainfigure{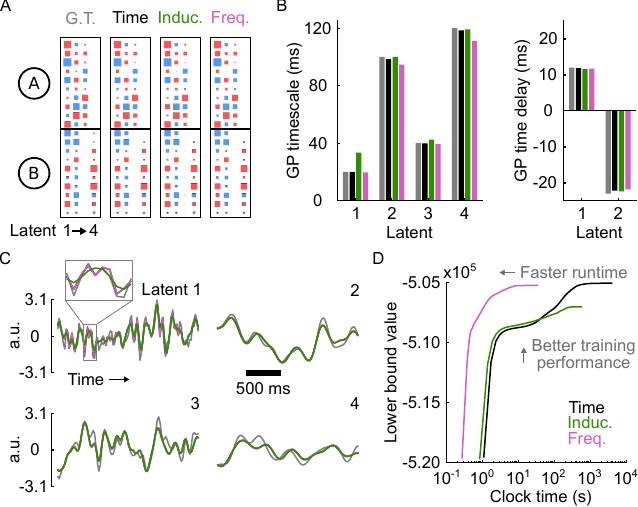}
\maincaption{demo}{Demonstrative simulation.}{\panel{a} Loading matrix estimates.
From left to right: Ground truth (G.T.), mDLAG-time (Time), mDLAG-inducing (Induc.), mDLAG-frequency (Freq.).
Here the loading matrices $C^1$ and $C^2$, for groups A and B, respectively, have been concatenated vertically.
Each element of each matrix is represented by a square: magnitude is represented by the square's area, and sign is represented by the square's color (red: positive; blue: negative).
Note that the sign and ordering of each loading matrix column is, in general, arbitrary.
We have therefore reordered and flipped the signs of the columns of the estimates to facilitate comparison with the ground truth.
\panel{b} Gaussian process (GP) parameter estimates.
Left: GP timescales. Right: GP time delays.
Time delays are only relevant for latents 1 and 2.
Latents 3 and 4 are local to groups A and B, respectively.
\panel{c} Single-trial latent time course estimates.
Each panel corresponds to the ground truth and estimated time course of a single latent variable.
All estimated time courses, regardless of the method used for model fitting, were computed using the mDLAG-time inference equation~\ref{eq:qx_time}.
Inset: zoomed-in view of latent 1 time courses.
For concision, for latents 1 and 2 we show only latents corresponding to group A ($\mathbf{x}^m_{n,j,:}$); the latents corresponding to group B are time-shifted versions of those shown here.
a.u.: arbitrary units.
\panel{d} Lower bound value versus elapsed clock time over the course of model fitting.
Note that each method optimizes a slightly different lower bound.
Lower bound values are therefore not directly comparable across methods.
Color scheme is consistent across panels \panel{a}--\panel{d}. Gray: ground truth; black: mDLAG-time; green: mDLAG-inducing; magenta: mDLAG-frequency.}
\clearpage

All fitting methods produced similar estimates.
The number of latent variables ($p = 4$) was correctly identified in all cases (\nameref{sec:methods}, Section~\ref{sec:model_selection}).
The loading matrices ($C$) and their group-wise sparsity patterns were also recovered with high accuracy (\fig{demo}{a}; normalized frobenius norm between ground truth loading matrix and estimates: mDLAG-time, 0.0462; mDLAG-inducing, 0.0538; mDLAG-frequency, 0.0467).
GP timescale (\fig{demo}{b}, left) and time delay (\fig{demo}{b}, right) estimates were all within 10\% of the ground truth, with one exception: mDLAG-inducing produced a significant overestimate of the timescale for latent 1 (\fig{demo}{b}, left, latent 1; compare green to gray).
Consequently, the trial-to-trial time course estimates of latent 1, based on the mDLAG-inducing fit, were overly smooth (\fig{demo}{c}, latent 1; see inset), leading to lower reconstruction accuracy than the other two fitting methods ($R^2$ between ground truth and estimated time courses: mDLAG-time, 0.8653; mDLAG-inducing, 0.8357; mDLAG-frequency, 0.8652).
We will elaborate on this phenomenon in subsequent sections.
The mDLAG-time and mDLAG-frequency methods produced practically indistinguishable estimates of latent time courses (\fig{demo}{c}, black and magenta traces coincide).

The runtime for mDLAG-frequency, however, was significantly faster than for the other two fitting methods (\fig{demo}{d}).
Per fitting iteration, mDLAG-frequency ran more than an order of magnitude faster than either mDLAG-time or mDLAG-inducing (wall clock time per iteration: mDLAG-time, 0.616 s; mDLAG-inducing, 0.464 s; mDLAG-frequency, 0.032 s).
Compared to mDLAG-time, mDLAG-frequency also required significantly fewer iterations to reach convergence (mDLAG-time, 6,952; mDLAG-inducing, 1,282; mDLAG-frequency, 1,091).
mDLAG-frequency appears to converge in fewer iterations than mDLAG-time because, at least in part, the GP parameters converge more quickly to their optimal values (\suppfig{param_progress_supp}).
The total runtime to convergence of mDLAG-frequency was thus faster by a factor of over 100 (total wall clock time: mDLAG-time, 4,283 s; mDLAG-inducing, 595 s; mDLAG-frequency, 35 s).

\subsection{Scaling with Number of Time Points per Trial}
\label{sec:scaling_T}

We next performed a more exhaustive characterization of the runtime performance of each fitting method, starting with runtime scaling as a function of the number of time points per trial.
We again generated simulated activity from two groups ($M = 2$) according to the mDLAG generative model (\nameref{sec:methods}, Section~\ref{sec:data_generation}), with a few key differences from above (\tbl{simulated-summary}, Scaling, $T$).
Without loss of generality, we fixed one global latent variable, shared across both groups ($p = 1$), and fixed the associated GP timescale and time delay throughout all experiments (100 ms and 10 ms for the GP timescale and time delay, respectively).
We generated datasets that comprised $T = 500$ time points per trial ($10$ s in length).
Beyond $T = 500$, mDLAG-time was prohibitively expensive to run, precluding method-to-method comparisons.
We then fit mDLAG models to increasingly long trial epochs (lengths spaced equally on a log scale from 10 to 500 time points per trial).
Here we chose the same number of estimated latent variables as the ground truth ($p = 1$).
Overall, we ran 20 independent experiments (``runs'' on independent datasets produced by different randomly generated mDLAG models) to gauge variability in performance.

Across nearly all trial lengths, $T$, the three fitting methods gave practically indistinguishable statistical performance (\fig{scalingT}{a}: black, green, and magenta traces visually indistinguishable).
All methods exhibited a leave-unit-out test $R^2$ value (Appendix~\ref{app:mdlag_pred_lgo}, Section~\ref{app:mdlag_pred_luo}) close to the noise ceiling (\fig{scalingT}{a}, black dashed line, $R^2 = 0.167$).
Below 50 time points per trial, mDLAG-frequency exhibited slightly worse performance (\fig{scalingT}{a}, inset; in the worst case, $R^2$ for mDLAG-frequency was 1\% worse than the other two methods at $T = 10$ time points per trial).

However, mDLAG-frequency exhibited dramatically better runtime scaling with the number of time points per trial.
In agreement with theory, per fitting iteration, mDLAG-time exhibited cubic scaling with $T$ (\fig{scalingT}{b}: compare mDLAG-time, black trace, to the cubic reference slope, gray dotted line), while mDLAG-frequency exhibited linear scaling or better (\fig{scalingT}{b}: compare mDLAG-frequency, magenta trace, to the linear reference slope, gray dashed line).
Fitting with mDLAG-frequency thus produced a 100x speed-up per iteration at $T = 500$ (mean clock time per iteration: mDLAG-time, $6.030 \pm 0.048$ s; mDLAG-inducing, $0.524 \pm 0.005$ s; mDLAG-frequency, $0.059 \pm 0.002$ s).
Because mDLAG-time also exhibited an increase in the number of iterations required for convergence (\fig{scalingT}{c}, at $T = 500$: mDLAG-time, $3,822 \pm 209$; mDLAG-inducing, $2,242 \pm 91$; mDLAG-frequency, $424 \pm 79$; see also \suppfig{param_progress_supp}), the computational advantage of mDLAG-frequency compounded: at $T = 500$, the overall runtime of mDLAG-frequency was 1,000x faster than that of mDLAG-time (\fig{scalingT}{d}, total clock time: mDLAG-time, $22,871 \pm 1,128$ s; mDLAG-inducing, $1,167 \pm 37.0$ s; mDLAG-frequency, $23.3 \pm 3.4$ s).

Note that mDLAG-inducing scaled superlinearly with $T$ (\fig{scalingT}{b}: compare mDLAG-inducing, green trace, to the linear reference slope, gray dashed line).
In theory, for a fixed number of inducing points, $T_{\text{ind}}$, mDLAG-inducing should scale linearly with the number of time points per trial.
Here, however, as $T$ increased, we also increased the number of inducing points to maintain it above the approximate Nyquist rate of the underlying latent time course (for a squared exponential GP with 100 ms timescale, a sampling rate of 12.5 Hz or greater captures essentially 100\% of the frequency content).
Without doing so, the performance of mDLAG-inducing suffers significantly (\suppfig{inducing_alias_supp}{}; see also \fig{demo}{c}, latent 1).
Consequently, the runtime benefit (with increasing $T$) of mDLAG-inducing over mDLAG-time is less significant than that of mDLAG-frequency.

\mainfigure{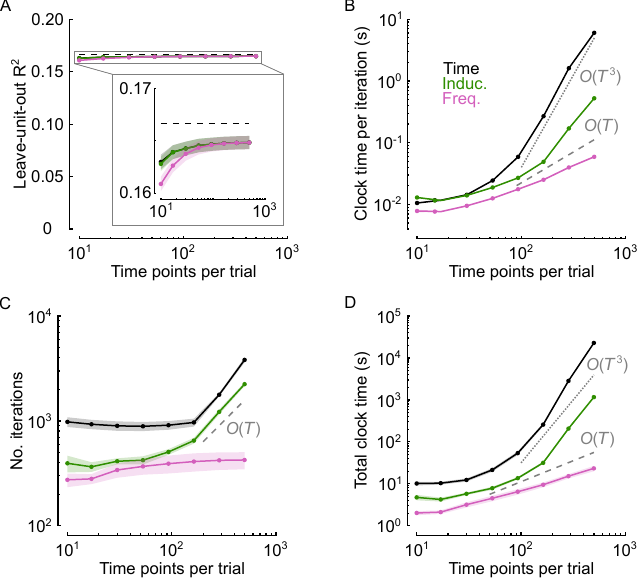}
\maincaption{scalingT}{Scaling of performance and runtime with number of time points per trial.}{\panel{a} Leave-unit-out $R^2$ versus number of time points per trial.
For each run, performance was evaluated on 100 held-out test trials.
Black dashed line: noise ceiling ($R^2 = 0.167$ for these simulations).
Inset: same data, but with vertical axis magnified to show the difference between methods.
\panel{b} Mean (across fitting iterations) clock time per iteration versus number of time points per trial.
Gray dashed line: reference slope for a linear scaling law, $O(T)$.
Gray dotted line: reference slope for a cubic scaling law, $O(T^3)$.
\panel{c} Number of iterations to convergence versus number of time points per trial.
Gray dashed line: reference slope for a linear scaling law, $O(T)$.
\panel{d} Total clock time to convergence versus number of time points per trial.
Gray dashed line: reference slope for a linear scaling law, $O(T)$.
Gray dotted line: reference slope for a cubic scaling law, $O(T^3)$.
Across panels \panel{a}--\panel{d}, solid traces indicate the mean, and shading (where visible) indicates standard error of the mean, computed across 20 runs.
Color scheme is consistent across panels.
Black: mDLAG-time (Time); green: mDLAG-inducing (Induc.); magenta: mDLAG-frequency (Freq.).}
\vspace{11pt}

\subsection{Scaling with Number of Groups}
\label{sec:scaling_M}

We performed a similar characterization of runtime scaling as a function of the number of observation groups (\tbl{simulated-summary}, Scaling, $M$).
Here, we fixed the total number of units throughout experiments at $q = 24$, but then considered a varying number of groups, from $M = 1$ (with 24 units belonging to one group) to $M = 24$ (with one ``group'' comprising one unit).
We again fixed one global latent variable ($p = 1$), with the same GP timescale as in Section~\ref{sec:scaling_T} (100 ms), but we selected the magnitudes of pairwise time delays, uniformly at random, between 0 ms and 20 ms.
The number of time points per trial was fixed at $T = 50$ ($1$ s in length).
The number of estimated latent variables was again chosen to match the ground truth ($p = 1$).
For mDLAG-inducing, we chose $T_{\text{ind}} = 13$ inducing points, the smallest number that maintained the approximate Nyquist rate of the latent signals (12--13 Hz for a squared exponential GP with 100 ms timescale).
Overall, we ran 20 independent experiments at each group number ($M$) to gauge variability in performance.

Across all group numbers, $M$, the three fitting methods gave practically indistinguishable statistical performance (\fig{scalingM}{a}: black, green, and magenta traces visually indistinguishable).
Both mDLAG-frequency and mDLAG-inducing, however, scaled significantly better than mDLAG-time.
In agreement with theory, per fitting iteration, mDLAG-time exhibited near-cubic scaling with the number of groups (\fig{scalingM}{b}: compare mDLAG-time, black trace, to the cubic reference slope, gray dotted line), while both mDLAG-frequency and mDLAG-inducing exhibited linear scaling or better (\fig{scalingM}{b}: compare mDLAG-frequency, magenta trace, and mDLAG-inducing, green trace, to the linear reference slope, gray dashed line).
Fitting with mDLAG-frequency and mDLAG-inducing thus produced a 285x and a 88x speed-up per iteration, respectively, at $M = 24$ (mean clock time per iteration: mDLAG-time, $7.200 \pm 0.091$ s; mDLAG-inducing, $0.0816 \pm  0.0013$ s; mDLAG-frequency, $0.0252 \pm 0.0004$ s).
The number of iterations required for mDLAG-time convergence also increased with the number of groups (\fig{scalingM}{c}, number of iterations at $M = 24$: mDLAG-time, $15,770 \pm 387$; mDLAG-inducing, $273 \pm 11$; mDLAG-frequency, $240 \pm 10$).
The computational advantages of mDLAG-frequency and mDLAG-inducing thus compounded, resulting in overall runtime speed-ups of over 18,000x and 5,000x for mDLAG-frequency and mDLAG-inducing, respectively, at $M = 24$ (\fig{scalingM}{d}, total clock time: mDLAG-time, $113,087 \pm 1,987$ s; mDLAG-inducing, $22.20 \pm 0.80$ s; mDLAG-frequency, $6.05 \pm 0.29$ s).

\clearpage
\mainfigure{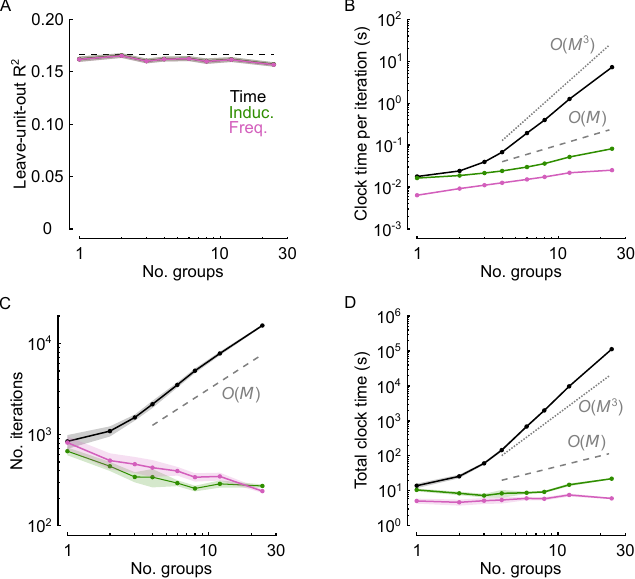}
\maincaption{scalingM}{Scaling of performance and runtime with number of groups.}{\panel{a} Leave-unit-out $R^2$ versus number of groups.
For each run, performance was evaluated on 100 held-out test trials.
Black dashed line: noise ceiling ($R^2 = 0.167$ for these simulations).
\panel{b} Mean (across fitting iterations) clock time per iteration versus number of groups.
Gray dashed line: reference slope for a linear scaling law, $O(M)$.
Gray dotted line: reference slope for a cubic scaling law, $O(M^3)$.
\panel{c} Number of iterations to convergence versus number of groups.
Gray dashed line: reference slope for a linear scaling law, $O(M)$.
\panel{d} Total clock time to convergence versus number of groups.
Gray dashed line: reference slope for a linear scaling law, $O(M)$.
Gray dotted line: reference slope for a cubic scaling law, $O(M^3)$.
Across panels \panel{a}--\panel{d}, solid traces indicate the mean, and shading (where visible) indicates standard error of the mean, computed across 20 runs.
Color scheme is consistent across panels.
Black: mDLAG-time (Time); green: mDLAG-inducing (Induc.); magenta: mDLAG-frequency (Freq.).}
\vspace{11pt}

\subsection{Neuropixels Recordings of Three Visual Cortical Areas}
\label{sec:npx}

We next sought to validate our accelerated methods beyond a simulated environment, on state-of-the-art neural recordings.
We considered electrophysiological recordings from neuronal populations in anesthetized macaque visual cortex, recorded using multiple Neuropixels probes (\nameref{sec:methods}, Section~\ref{sec:npx_methods}).
These recordings comprised hundreds of neurons (min: 231, max: 450) spanning three brain areas: V1, V2, and V3d.
Each brain area was treated as a separate group throughout our analyses below (i.e., $M = 3$).
Within a recording session, on each trial, animals were visually presented with a drifting sinusoidal grating (one of two possible orientations, 90\textdegree \ apart).
Each orientation was presented on 300 trials, for a total of 600 trials per recording session.

Overall, we analyzed seven recording sessions from three animals.
We further treated each grating stimulus orientation separately, giving a total of 14 ``datasets'' (a subset of which were previously analyzed in \citealp{gokcen_uncovering_2023}).
For each dataset, we allocated at random 225 trials as a training set and 75 trials as a test set on which to measure model performance (Appendix~\ref{app:mdlag_pred_lgo}).
We applied each fitting method (mDLAG-time, mDLAG-inducing, mDLAG-frequency) to spike counts (taken in 20 ms non-overlapping time bins) measured during the first 1,000 ms after stimulus onset (\fig{npx}{a}, trial-averaged responses for an example dataset).
As a preprocessing step, for each neuron, we subtracted the mean spike count across time bins within each trial to remove slow fluctuations beyond the timescale of a trial \citep{cowley_slow_2020}.
We started all fitting methods from the same initialization, with $p = 30$ latents (\nameref{sec:methods}, Section~\ref{sec:mdlag_init}).
In terms of scale, these recordings included many more trials and neurons than the simulations in Sections \ref{sec:scaling_T} and \ref{sec:scaling_M}, but the number of time points per trial ($T = 50$) and groups ($M = 3$) are within the ranges considered above.

mDLAG-frequency performed statistically as well as mDLAG-time (\fig{npx}{b}, magenta points lie on the diagonal), yet achieved a median 3x speed-up per fitting iteration, and a 25x speed-up overall (\fig{npx}{c}, median clock time per iteration across datasets: mDLAG-time, 5.88 s; mDLAG-frequency, 2.00 s; \fig{npx}{d}, median number of iterations across datasets: mDLAG-time, 30,737; mDLAG-frequency, 3,659; \fig{npx}{e}, median total clock time across datasets: mDLAG-time, 50.4 hrs; mDLAG-frequency, 1.8 hrs).
mDLAG-inducing, by contrast, presented a performance trade-off that depended on the choice of the number of inducing points.
At $T_{\text{ind}} = 20$ inducing points, mDLAG-inducing achieved a runtime similar to mDLAG-frequency (\fig{npx}{c}, median clock time per iteration across datasets: 2.03 s; \fig{npx}{d}, median number of iterations across datasets: 3,457; \fig{npx}{e}, median total clock time across datasets: 1.9 hrs).

\clearpage
\mainfigure{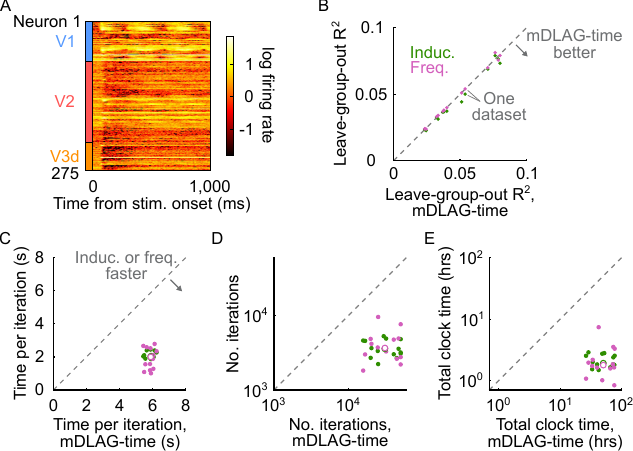}
\maincaption{npx}{Performance and runtime on Neuropixels recordings from macaque visual cortex.}{Here $M = 3$ groups, corresponding to the three brain areas V1, V2, and V3d. \panel{a} Temporally smoothed peristimulus time histograms during the stimulus presentation period, for an example session and stimulus condition.
\panel{b} Performance (leave-group-out $R^2$) of mDLAG-inducing (green points; Induc.) or mDLAG-frequency (magenta points; Freq.) versus performance of mDLAG-time.
Each data point represents one dataset.
For each dataset, performance was evaluated on 75 held-out test trials.
mDLAG-inducing (with $T_{\text{ind}} = 20$ inducing points) was significantly outperformed by mDLAG-time and mDLAG-frequency (one-sided paired sign tests: mDLAG-time better than mDLAG-inducing, $p = 0.0065$; mDLAG-frequency better than mDLAG-inducing, $p = 0.0009$).
\panel{c} Mean clock time per iteration, mDLAG-inducing or mDLAG-frequency versus mDLAG-time.
As in panel \panel{b}, each data point represents one dataset.
White-filled circles indicate the median of the green points (circle with green border) or of the magenta points (circle with magenta border).
\panel{d} Number of iterations to convergence, mDLAG-inducing or mDLAG-frequency versus mDLAG-time.
Same conventions as in panel \panel{c}.
\panel{e} Total clock time to convergence, mDLAG-inducing or mDLAG-frequency versus mDLAG-time.
Same conventions as in panel \panel{c}.}
\vspace{22pt}

However, $T_{\text{ind}} = 20$ inducing points corresponds to an effective sampling rate of 20 Hz.
At a 20 Hz sampling rate, aliasing becomes a potential issue for latents with (squared exponential) GP timescales of about 50 ms or shorter.
For instance, by mDLAG-time estimates, about 25\% of the GP timescales encountered across Neuropixels datasets were 35 ms or shorter.
By contrast, no mDLAG-inducing timescale estimates were shorter than 35 ms (\suppfig{npx_timescales_supp}).
Models fit by mDLAG-inducing consequently performed significantly worse than models fit by the other two methods (\fig{npx}{b}, one-sided paired sign tests, leave-group-out $R^2$: mDLAG-time better than mDLAG-inducing, $p = 0.0065$; mDLAG-frequency better than mDLAG-inducing, $p = 0.0009$).
This gap in mDLAG-inducing's statistical performance could be closed by increasing the number of inducing points (\suppfig{npx_supp_S32}{a}: mDLAG-inducing performance when we chose $T_{\text{ind}} = 32$ inducing points, corresponding to an approximate Nyquist rate for GP timescales of 30 ms or longer), but at the expense of runtime benefit (\suppfig{npx_supp_S32}{b}: mDLAG-frequency 55\% faster than mDLAG-inducing per iteration, and 72\% faster overall).

\subsection{Exploration of Biases Introduced by Frequency Domain Fitting}
\label{sec:bias}

Thus far we have shown that the frequency domain fitting approach provides orders of magnitude runtime speed-up with seemingly minimal loss in statistical performance relative to baseline.
Of course, no approximation comes without a cost.
We now explore the biases introduced by the frequency domain approach, particularly in the estimation of GP parameters and of latent dimensionality.

But first, why might there be biases in the first place?
The approximation to take the frequency domain PSD matrix as diagonal is equivalent to replacing the time domain covariance matrix with a circulant matrix (\fig{circulant}{a}, top: time domain covariance matrix; \fig{circulant}{b}, top: circulant approximation arising implicitly from the mDLAG-frequency model).
This circulant approximation affects both the generative model and posterior inference.
Latent time courses generated via the approximate frequency domain model (equation~\ref{eq:xprior_freq}, followed by an inverse DFT) are putatively periodic (\fig{circulant}{b}, bottom: note the continuity between the solid trace and its copies, the dotted traces), with correlations between the beginning of the trial and the end of the trial (\fig{circulant}{b}, top: note the large magnitude of covariance in the upper right and lower left quadrants of the circulant covariance matrix $\hat{K}$).

Like frequency-domain-generated time courses, frequency-domain-inferred time courses are similarly periodic, leading to edge effects (\fig{circulant}{c}).
In the middle of a trial, time domain and frequency domain inference produce practically indistinguishable estimates (\fig{circulant}{c}: compare black and magenta traces).
However, the estimates start to visibly deviate toward the beginning and end of the trial, as the frequency domain estimate curves away from the time domain estimate to satisfy its periodic boundary conditions.
The extent of edge effects, in both generation and inference, increases as the length of the GP timescale increases relative to the length of the trial.

\clearpage
\mainfigure{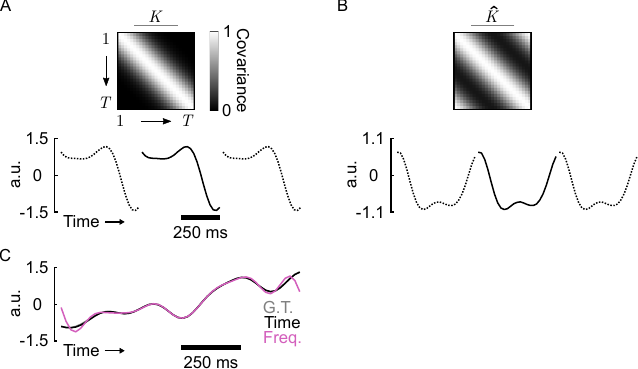}
\maincaption{circulant}{The circulant approximation induced by the frequency domain approach and its effects on generation and inference.}{\panel{a} Generation via the time domain. Top: An example Gaussian process (GP) covariance matrix ($K$), corresponding to the mDLAG-time model.
Bottom: An example GP time course (black solid trace) generated via the mDLAG-time model.
To emphasize the (lack of) boundary conditions, copies of that time course are displayed to its left and right (black dotted traces).
\panel{b} Generation via the frequency domain. Top: The circulant approximation of $K$ in panel \panel{a} ($\hat{K}$), arising implicitly from the mDLAG-frequency model.
Both $K$ and $\hat{K}$ shown here were generated with the same squared exponential GP timescale (100 ms) for trial length $T = 25$ time points and $M = 1$ group.
Bottom: An example GP time course (black solid trace) generated via the mDLAG-frequency model.
The signal was generated in the frequency domain, and then transformed into the time domain via the Discrete Fourier Transform (DFT).
To emphasize the periodic boundary conditions, copies of that time course are displayed to its left and right (black dotted traces).
\panel{c} Inference via the time domain versus the frequency domain.
A ground truth latent time course (G.T., gray; occluded by the black trace) was generated via the mDLAG-time model (trial length $T = 50$ time points, $M = 1$ group, and GP timescale 100 ms).
Then, time courses were inferred via the mDLAG-time model (Time, black; equation~\ref{eq:qx_time}) or the mDLAG-frequency model (Freq., magenta; equation~\ref{eq:qx_freq}, followed by DFT).
In either case, the ground truth parameters were used for inference.
a.u.: arbitrary units.}
\vspace{11pt}

\subsubsection{Biases in Estimation of Gaussian Process Parameters}
\label{sec:bias_gp}

To characterize the impact of the frequency domain approximation on GP parameter estimation, we returned to our simulated datasets from above.
Here we will focus on bias as a function of trial length (\tbl{simulated-summary}, Scaling, $T$), as the departure of the frequency domain approximation from the original time domain specification is ultimately a finite trial length effect.
We explore biases as a function of group number (\suppfig{bias_params_supp}{a}), sampling rate (\supptbl{simulated-summary-supp}, Sampling rate; \suppfig{bias_params_supp}{b}), and number of units (\supptbl{simulated-summary-supp}, Scaling, $q$; \suppfig{bias_params_supp}{c}) in the Supplement.

Comparing the GP timescales and time delays estimated via mDLAG-frequency to the ground truth values, we found that the magnitudes of both parameters were consistently underestimated (\fig{bias_params}{a,b}, magenta traces).
Biases were more severe for shorter trial lengths, and approached the ground truth for longer trial lengths.
In separate analyses, we re-fit models via mDLAG-frequency while keeping either the GP timescales or the GP time delays fixed throughout fitting.
Regardless of which parameter was held fixed, the bias remained in estimates of the other parameter, suggesting that these biases are decoupled effects (\suppfig{bias_params_supp_fix}{}).

Intuition for this effect can be gained by considering the time domain covariance matrix and its (approximate) circulant counterpart.
Suppose, for example, a ground truth latent variable with covariance matrix as depicted in \fig{circulant}{a}.
Consider also a frequency domain (circulant in the time domain) estimate with GP timescale that matches the ground truth value (\fig{circulant}{b}, top).
If we consider the error between the ground truth covariance matrix and the circulant estimate, then the error for elements close to the main diagonal will be low, while the error for elements in the upper right and lower left quadrants will be relatively high.
The error accumulated in these upper right and lower left quadrants could be reduced if the GP timescale estimate was shorter, but at the expense of added error for elements close to the main diagonal.
Ultimately, the optimal timescale balances these sources of error, settling on an estimate that is shorter than the ground truth, but not too short.
We validated this intuition empirically in simulation (\suppfig{cov_error_supp}{}).
Importantly, as the length of the trial increases, the bias approaches zero (\fig{bias_params}{a,b}, magenta traces): the bias becomes less concerning precisely as the runtime of mDLAG-frequency becomes more advantageous (\fig{scalingT}).

Why does this bias not lead to more severe negative impacts on model performance (\fig{scalingT}{a}, \fig{scalingM}{a}, \fig{npx}{b})?
An underestimate of GP timescale means that higher frequencies are overrepresented \textit{a priori}.
In principle, then, posterior estimates of the latent time courses are encouraged to maintain potentially spurious high-frequency content in held-out test trials, leading to overfitting.
In practice, however, given the consistently similar performance between mDLAG-frequency and mDLAG-time, overfitting was evidently not an issue for the simulated datasets and neural recordings considered here.
An overestimate of GP timescale, by contrast, would encourage underfitting, as meaningful high-frequency content is suppressed.
For the neural recordings considered here, that type of underfitting appeared to be the more salient issue, as evidenced by the relatively poor performance of mDLAG-inducing model fits (\fig{npx}{b}) paired with their consistent overestimates of GP timescales (\suppfig{npx_timescales_supp}).

\clearpage
\mainfigure{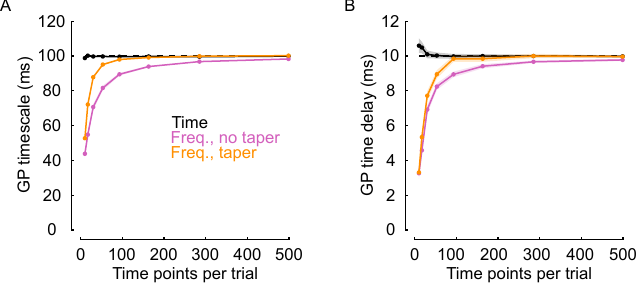}
\maincaption{bias_params}{Bias in Gaussian process (GP) parameter estimation with number of time points per trial.}{\panel{a} GP timescale estimates versus number of time points per trial.
\panel{b} GP time delay estimates versus number of time points per trial.
Across panels \panel{a} and \panel{b}, black dashed traces indicate ground truth parameter values.
Solid traces indicate the mean of estimates, and shading (where visible) indicates standard error of the mean, computed across 20 runs.
Black: mDLAG-time (Time); magenta: mDLAG-frequency without tapering (Freq., no taper); gold: mDLAG-frequency with tapering (Freq., taper).}
\vspace{11pt}

\subsubsection{Biases in Estimation of Dimensionality}
\label{sec:bias_dim}

We next explored potential biases in the estimation of latent dimensionality, i.e., model selection.
In the case of methods like GPFA \citep{yu_gaussian-process_2009} and DLAG \citep{gokcen_disentangling_2022}, model selection is accomplished via cross-validation and grid search to find optimal latent dimensionalities.
In the case of mDLAG, estimation of latent dimensionality is incorporated into the model fitting process via the ARD prior, which encourages group-wise sparsity of the loadings (equations \ref{eq:Cprior}, \ref{eq:alphaprior}).
Here we focus on estimation of latent dimensionality via ARD (\nameref{sec:methods}, Section~\ref{sec:model_selection}).

As we will demonstrate, the results of model selection are noise-level dependent.
We therefore generated (via the mDLAG generative model; \nameref{sec:methods}, Section~\ref{sec:data_generation}) 10 independent datasets at four different noise levels (SNR = 0.01, 0.1, 1.0, 10.0), for a total of 40 datasets (\tbl{simulated-summary}, Model Selection).
Each dataset comprised $N = 50$ trials, with 200 time points per trial (400 ms in length).
Here we were also interested in bias as a function of trial length, and thus fit mDLAG models to increasingly long trial epochs (lengths spaced equally on a log scale from 10 to 200 time points per trial).
For simplicity, we generated activity for just one group with 24 units.
The themes from model selection for one group are representative of themes that underlie model selection in multi-group contexts.
We set the ground truth number of latents to $p = 4$, but initialized all models with $p = 8$ latents to allow dimensionality to be freely estimated.
All ground truth GP timescales were set to 50 ms.

\clearpage
\mainfigure{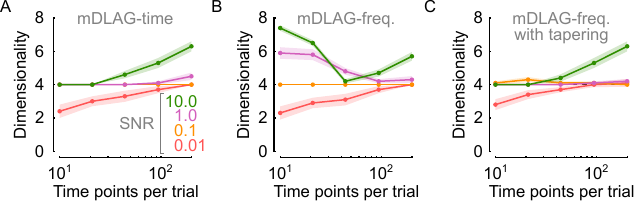}
\maincaption{bias_dim}{Bias in estimation of latent dimensionality with number of time points per trial.}{\panel{a} mDLAG-time estimates.
\panel{b} mDLAG-frequency (mDLAG-freq.) estimates.
\panel{c} mDLAG-frequency estimates with tapering as a preprocessing step.
Across panels \panel{a}--\panel{c}, black dashed lines indicate ground truth dimensionality ($p = 4$).
Solid traces indicate the mean of estimates, and shading (where visible) indicates standard error of the mean, computed across 10 runs.
SNR: Signal-to-noise ratio.
Red: SNR $= 0.01$; gold: SNR $= 0.1$; magenta: SNR $= 1.0$; green: SNR $= 10.0$.}
\vspace{22pt}

We first considered trends in dimensionality estimates via mDLAG-time as a baseline (\fig{bias_dim}{a}).
Biases in these estimates depended on trial length.
At low SNRs (\fig{bias_dim}{a}, red, SNR $= 0.01$), mDLAG-time produced conservative dimensionality estimates---underestimates that approached the ground truth with increasing trial length (and increasing training dataset size).
At mid-level SNRs (\fig{bias_dim}{a}, gold and magenta, SNR $= 0.1, 1.0$, respectively), dimensionality estimates were accurate at all trial lengths.
But interestingly, at high SNRs (\fig{bias_dim}{a}, green, SNR $= 10.0$), mDLAG-time produced overestimates of dimensionality at long trial lengths (large training dataset sizes).
This behavior has been reported previously for simpler dimensionality reduction approaches like factor analysis, in conjunction with cross-validation \citep{williamson_scaling_2016}: in low-noise, data-rich regimes, the effects of overfitting are small, and cross-validated performance remains high even for models with dimensionalities larger than the ground truth.

We then sought to characterize the extent to which dimensionality estimates via mDLAG-frequency matched or deviated from the trends described above (\fig{bias_dim}{b}).
At the lowest two SNRs (\fig{bias_dim}{b}, red and gold, SNR $= 0.01, 0.1$, respectively), mDLAG-frequency estimates were qualitatively indistinguishable from estimates by mDLAG-time (compare to \fig{bias_dim}{a}).
At the highest two SNRs (\fig{bias_dim}{b}, magenta and green, SNR $= 1.0, 10.0$, respectively), trends for the two fitting methods were notably different, at least for shorter trial lengths ($T < 44$).
In contrast with mDLAG-time estimates, mDLAG-frequency estimates were biased toward dimensionality values that were larger than the ground truth at shorter trial lengths, and approached the ground truth as trial length increased (up to $T = 44$).
Then as trial length continued to grow, the trends in mDLAG-frequency estimates matched the trends of mDLAG-time estimates, growing in magnitude above the ground truth value (for $T = 44$ and larger).

The trend for short trial lengths can be explained by the bias in GP timescale estimation (see previous section, \ref{sec:bias_gp}).
mDLAG-frequency tends to produce underestimates of GP timescales at short trial lengths---a mismatch with the data.
By employing additional latents, mDLAG-frequency could possibly compensate for any explanatory power lost by this parameter bias.
As trial length increases, the bias in GP timescale decreases; mDLAG-frequency estimates behave increasingly like mDLAG-time estimates.

\subsubsection{Bias Mitigation Strategies}
\label{sec:taper}

We have demonstrated empirically that the biases explored above have a minimal impact on model performance, and that such biases diminish as the runtime benefit of mDLAG-frequency grows.
Still, it might be of practical and scientific importance to mitigate these biases during the model fitting process.
Bias mitigation is a mature topic of study in spectral estimation, with many strategies to choose from \citep[Chapter 7]{priestley_spectral_1981}.
To demonstrate that this body of work is applicable to the present context, we considered one such strategy: tapering.

We repeated all simulated experiments with mDLAG-frequency (\fig{bias_params}{}, \fig{bias_dim}{}, \suppfig{bias_params_supp}{}).
Here, however, as a preprocessing step, we applied a taper (we chose a periodic Hamming window function; \nameref{sec:methods}, Section~\ref{sec:taper_method}) to each observed unit on each trial.
Tapering smoothly attenuates each observed time series toward zero at the beginning and end of each trial, effectively enforcing the periodic boundary conditions assumed by the frequency domain approach.

Across experiments, tapering was an effective bias mitigation strategy.
For instance, in trial length experiments, mDLAG-frequency (without tapering) achieved within 10\% error of the ground truth GP timescale by $T = 163$ (\fig{bias_params}{a}, magenta).
With tapering, that error was achieved by $T = 53$ (\fig{bias_params}{a}, gold).
In model selection experiments (\fig{bias_dim}{c}), tapering removed the tendency of mDLAG-frequency to overestimate dimensionality at short trial lengths (compare \fig{bias_dim}{c} to \fig{bias_dim}{b}, for trial lengths $T = 44$ and shorter).
Tapering was also effective for bias mitigation in our experiments scaling group number (\suppfig{bias_params_supp}{a}), sampling rate (\suppfig{bias_params_supp}{b}), and number of neurons (\suppfig{bias_params_supp}{c}).

As an alternative to tapering, one could use mDLAG-frequency to fit an initial set of model parameters, and then fine-tune those parameters using mDLAG-time for a limited number of iterations.
For example, we employed this approach on one of the Neuropixels datasets (\suppfig{npx_finetune_supp}).
After 500 iterations of fine-tuning with mDLAG-time, GP timescales and time delays settled into slightly different values from the initial mDLAG-frequency estimates.

\section{Discussion}
\label{sec:discussion}

In this work we developed two methods to accelerate model fitting and inference in multi-group GP factor models.
Our methods incorporate multi-group extensions of techniques drawn from two broader classes of approaches in the GP and spectral estimation fields: sparsity via inducing variables \citep{titsias_variational_2009, duncker_temporal_2018, alvarez_efficient_2010} and frequency domain representations \citep{whittle_hypothesis_1951, dowling_linear_2023, keeley_identifying_2020, paciorek_bayesian_2007, ulrich_gp_2015}.
We demonstrated that mDLAG-inducing and mDLAG-frequency can achieve linear scaling in both the number of time points per trial $T$ and the number of groups $M$, translating to orders of magnitude runtime speed-ups over the baseline approach, mDLAG-time, across simulations and neural recordings.
Notably, in neural recordings of hundreds of neurons across three brain areas, mDLAG-frequency produced a 25x speed-up over mDLAG-time without compromising statistical performance.
Prior to this work, the analysis of increasingly available recordings that span many brain areas, cortical layers, and cell types was prohibitive for the class of multi-group GP factor models.
With the methods developed here, that analysis is now feasible, opening the door to questions about multi-population interactions that were previously inaccessible.

In contrast with mDLAG-frequency, mDLAG-inducing often presented a trade-off between runtime and statistical performance.
Applied to neural recordings, mDLAG-inducing could achieve either equal runtime performance to mDLAG-frequency (\fig{npx}) or equal statistical performance to mDLAG-frequency (\suppfig{npx_supp_S32}), but not both.
In theory, mDLAG-inducing scales linearly in $T$ so long as the number of inducing points $T_{\text{ind}}$ is held fixed.
In practice, $T_{\text{ind}}$ must be kept large enough to avoid aliasing and excessively overestimating latent timescales (\fig{demo}{b}, \suppfig{inducing_alias_supp}, \suppfig{npx_timescales_supp}), hence tempering the runtime benefit of the approach (\fig{demo}{d}, \fig{scalingT}).
The need to search for the optimal hyperparameter $T_{\text{ind}}$, when the (potentially wide-ranging) latent timescales are not known \textit{a priori}, tempers that runtime benefit further.

Of course, mDLAG-frequency is not without trade-offs.
For small trial lengths, we demonstrated that mDLAG-frequency exhibits a tendency to underestimate the magnitudes of GP timescales and time delays (\fig{bias_params}, \suppfig{npx_timescales_supp}), and a tendency to overestimate dimensionality (\fig{bias_dim}{b}).
Both of these biases could leave models fit via mDLAG-frequency prone to overfitting.
For the simulated data and neural recordings considered here, however, overfitting did not appear to be an issue.
Importantly, as trial length increases, mDLAG-frequency's bias diminishes (\fig{bias_params}, \fig{bias_dim}) precisely as its runtime benefit grows (\fig{scalingT}).
For instance, in the simulations with $T = 500$ time points per trial, a 1,000x runtime speed-up (\fig{scalingT}{d}) with less than 2\% bias in GP parameter estimates (\fig{bias_params}) represents a remarkably good trade-off.

We explored tapering and fine-tuning in the time domain as straightforward yet effective bias mitigation strategies (Section~\ref{sec:taper}).
While these strategies worked well for the simulated and neural datasets considered here, they might not always be the best strategy for the problem at hand.
In other contexts, other strategies could be explored, such as zero-padding \citep{aoi_scalable_2017} or even debiased modifications to the variational lower bound \citep{sykulski_debiased_2019}, though likely at the expense of increased computational cost.

We have focused here on approaches to accelerating one particular dimensionality reduction method, mDLAG; but in fact, mDLAG is representative of a general class of dimensionality reduction methods.
Our approaches are therefore quite general, applicable to a wide range of dimensionality reduction methods commonly used in neuroscience.
In the case of two groups ($M = 2$), mDLAG is equivalent to a Bayesian formulation of DLAG \citep{gokcen_disentangling_2022}, and closely related to other multi-group approaches that incorporated GP state models \citep{keeley_identifying_2020, balzani_probabilistic_2023}.
In the case of one group ($M = 1$), and when all time delays are fixed to zero ($D^m_j = 0$), mDLAG becomes equivalent to a Bayesian formulation of Gaussian process factor analysis (GPFA; \citealp{yu_gaussian-process_2009, duncker_temporal_2018, jensen_scalable_2021}).
In the case where each ``group'' comprises one unit ($M = q$), mDLAG is akin to a time-delay GPFA \citep{lakshmanan_extracting_2015} or cross-spectral factor analysis \citep{gallagher_cross-spectral_2017} model with a sparsity prior on each coefficient of the loading matrix, $C$.
By removing temporal smoothing (i.e., in the limit as all GP timescale parameters $\tau_j$ approach $0$) mDLAG becomes equivalent to the static method group factor analysis \citep{klami_group_2015}.

The accelerated fitting methods developed here, in the Bayesian mDLAG setting, are readily applicable to non-Bayesian methods like GPFA and DLAG.
For these non-Bayesian methods, development would largely follow Sections \ref{sec:mdlag-induce} and \ref{sec:mdlag-freq}, but the mean parameter $\mathbf{d}^m$, the loading matrix $C^m$, and the noise precision matrix $\Phi^m$ would be treated as deterministic model parameters for which we seek point estimates, rather than probabilistic model parameters for which we seek posterior distributions.
In fact, we include an implementation of these methods in accompanying code (see \nameref{sec:code}).
We note also that we restricted our development here to models with a linear-Gaussian relationship between latents and observations (equations~\ref{eq:mdlag_obs1} and \ref{eq:mdlag_obs2}).
Our approaches could be adapted to nonlinear regimes \citep{wu_gaussian_2017, duncker_scalable_2023, gondur_multi-modal_2024} and non-Gaussian noise models \citep{keeley_efficient_2020, duncker_temporal_2018, zhao_variational_2017}, though again likely at the expense of increased computational cost.

Our investigations focused on the commonly used squared exponential GP covariance function with time delays (equation~\ref{eq:k}).
We emphasize, however, that all three fitting methods examined here are compatible with any stationary GP covariance function, while possessing the same computational scaling properties (\fig{scalingT}, \fig{scalingM}).
The mDLAG-frequency mathematical development, specifically, need not depend on the particular parametrization of the GP cross-spectral density (CSD) function given in equation~\ref{eq:scross}.
Rather, any GP CSD function that can be written in terms of an amplitude function and a phase function is compatible with the approach.
Concretely, for latent $j$ and groups $m_1$ and $m_2$, we can write the GP CSD function $s_{m_1,m_2,j}(f_l)$, evaluated at frequency $f_l$, in terms of the amplitude function $\zeta_j(f_l)$ and phase function $\eta_{m_1,m_2,j}(f_l)$:
\begin{equation}
    s_{m_1,m_2,j}(f_l) = \zeta_j(f_l) \cdot \exp\left( -i \eta_{m_1,m_2,j}(f_l) \right)
\end{equation}

For the squared exponential GP covariance function (equation~\ref{eq:k}), the corresponding amplitude function $\zeta_j(f_l)$ is the squared exponential in equation~\ref{eq:s}.
The phase function is a linear function of frequency, $\eta_{m_1,m_2,j}(f_l) = 2 \pi f_l (D^{m_1}_j - D^{m_2}_j)$, where the slope is determined by the difference between time delay parameters $D^{m_1}_j$ and $D^{m_2}_j$.
Alternative choices for the amplitude function can be drawn directly from the single-output GP literature \citep[e.g.,][]{wilson_gaussian_2013}.
In tandem, alternative choices of the phase function would lead to alternative multi-output GP kernels, for instance, constant (as in a simple phase shift; \citealp{ulrich_gp_2015}) or affine \citep{parra_spectral_2017} functions of frequency.
Different choices of GP kernel would lead to different behavior in terms of statistical performance and biases in estimation.
These differences ought to be explored on a case-by-case basis.
Still, the themes raised in Section~\ref{sec:bias} will be relevant to any alternative GP kernel choices.

Multi-group dimensionality reduction methods have historically faced a number of scaling challenges, from model selection to model fitting.
The challenge of model selection requires identifying both the number of latents across all observed groups, and which subset of groups each latent involves.
For example, in the case of $M = 2$ groups, the non-Bayesian DLAG \citep{gokcen_disentangling_2022} (and methods with similar group structure, e.g., \citealp{keeley_identifying_2020, balzani_probabilistic_2023}) has three hyperparameters: one for the number of shared latents, and two for the number of latents local to each group.
Grid search over just 10 candidate values for each of these three hyperparameters would result in $10^3$ candidate models to fit, and $k$-fold cross-validation would inflate that number further.
To generalize this model selection approach to $M$ groups would require the fitting of $p^{2^M-1}k$ candidate models, for a search over $p$ candidate values for each of $2^M-1$ hyperparameters with $k$-fold cross-validation.
With the incorporation of automatic relevance determination \citep{klami_group_2015}, mDLAG reduced the number of model fits to one.
The methods developed here then accelerate that model fit to overcome remaining computational barriers, thereby enabling a broad class of dimensionality reduction techniques to keep pace with the rapidly growing scale of multi-population neural recordings.

\section{Methods}
\label{sec:methods}

\subsection{Synthetic Data Generation via the Time and Frequency Domains}
\label{sec:data_generation}

Throughout our simulated experiments (\tbl{simulated-summary}, \supptbl{simulated-summary-supp}), we generated the simulated datasets using the mDLAG generative model.
In Section~\ref{sec:demo} (\fig{demo}; \tbl{simulated-summary}, Demo), generation could be performed straightforwardly via the mDLAG-time formulation, equations \ref{eq:mdlag_obs1}--\ref{eq:delta_k}.
For the remaining larger-scale experiments (\tbl{simulated-summary}: Scaling, $T$; Scaling, $M$; Model selection; \supptbl{simulated-summary-supp}: Sampling rate; Scaling, $q$), data generation via the mDLAG-time formulation was prohibitively expensive.
Sampling from the multivariate Gaussian distribution in equation~\ref{eq:xprior_time} requires $\mathcal{O}(M^3 T^3)$ operations, cubic in the number of groups, $M$, and the number of time points per trial, $T$.

For these remaining experiments, we therefore generated simulated datasets using the mDLAG-frequency formulation, equations \ref{eq:s}--\ref{eq:xprior_freq}, \ref{eq:mdlag_obs1_freq}--\ref{eq:mdlag_obs2_freq}, and \ref{eq:dprior}--\ref{eq:alphaprior} \citep{dietrich_fast_1997}.
The computational cost of sampling frequency domain observations is ameliorated to $\mathcal{O}(MT)$ operations.
Then, we convert these observations to the time domain by taking the inverse unitary DFT of the frequency domain activity for each unit $r = 1,\ldots,q_m$ in group $m$ on each trial $n$: $\mathbf{y}^m_{n,r,:} = U^{\mathsf{H}}_T \widetilde{\mathbf{y}}^m_{n,r,:}$.
Assuming the DFT is carried out using an FFT algorithm, this operation scales $\mathcal{O}(T \log T)$ in the number of time points per trial.

Yet time courses generated via mDLAG-frequency exhibit edge effects (Section~\ref{sec:bias}, \fig{circulant}{b}).
To mitigate the influence of these edge effects on our experimental results, we conservatively first generated trials that were three times longer than desired ($3T$ time points in length, when we desired length-$T$ trials).
Then we took only the middle third of each trial, and threw out the first and final thirds.
For our simulated experiments, in which latent time courses were generated with a squared exponential GP timescale of 100 ms, taking only the middle third of the trial was sufficient for edge effects to be rendered negligible.
Even with the generation of extra time points per trial, data generation via the mDLAG-frequency formulation was still markedly faster than data generation via the mDLAG-time formulation.

\subsection{Parameter Initialization}
\label{sec:mdlag_init}

We initialized each of the three methods, mDLAG-time, mDLAG-inducing, and mDLAG-frequency, in a similar fashion.
For any of the methods, we first specified the number of latents, $p$.
Through automatic relevance determination, insignificant latents are then effectively pruned.
Therefore, in general, $p$ should be chosen to be as small as possible (to minimize runtime) yet large enough that at least one of the initial latents is deemed insignificant (according to shared variance explained, see equation \ref{eq:varexp}, below), thereby ensuring that dimensionalities are not underestimated.
We indicate our chosen value of $p$ for each experiment throughout the Results.

To initialize the rest of the fitting procedure, we specified initial values for only the moments of the posterior factors $Q_d(\mathbf{d})$, $Q_{\phi}(\boldsymbol{\phi})$, $Q_c(C)$, and $Q_{\mathcal{A}}(\mathcal{A})$ (see, for example, equations \ref{eq:qd_induce}--\ref{eq:qalpha_induce}) that were needed to begin iteration.
The posterior distribution over the latents $Q_x(X)$ (for mDLAG-time, equation~\ref{eq:qx_time}; $Q_w(W)$ over inducing variables for mDLAG-inducing, equation~\ref{eq:qw_app}; $Q_{\widetilde{x}}(\widetilde{X})$ over frequency domain latents for mDLAG-frequency, equation~\ref{eq:qx_freq}) was then the first factor to be updated each iteration of the fitting procedure.
We specified noninformative priors by fixing all hyperparameters to a very small value \citep{klami_group_2015}, $\beta, a_{\phi}, b_{\phi}, a_{\alpha}, b_{\alpha} = 10^{-12}$.

For $Q_d(\mathbf{d})$, we initialized $\boldsymbol{\mu}_d^m$ at the sample mean of observed activity across all trials and time points.
For $Q_{\phi}(\boldsymbol{\phi})$, we initialized $\langle \phi^m_r \rangle^{-1}$ for each unit $r$ in group $m$ to the sample variance of that unit across all trials and time points.
For $Q_c(C)$, we first randomly initialized all first moments $\bar{\boldsymbol{\mu}}_{c_r}^m \in \mathbb{R}^{p}$ (for the $r$\textsuperscript{th} row of $C^m$ in group $m$, see equation~\ref{eq:Cmean}) with entries drawn from a zero-mean Gaussian distribution with variance chosen to match the scale of the data.
Then, we initialized the second moments $\langle \bar{\mathbf{c}}_r^m (\bar{\mathbf{c}}_r^m)^{\top} \rangle$ to the outer product of first moments $\bar{\boldsymbol{\mu}}_{c_r}^m \bar{\boldsymbol{\mu}}_{c_r}^{m\top}$.
For $Q_{\mathcal{A}}(\mathcal{A})$, we initialized $\langle \alpha^m_j \rangle$ for each latent $j$ in group $m$ to $q_m / \langle \lVert \mathbf{c}_j^m \rVert_2^2 \rangle$, which stems from equations \ref{eq:alpha-induce_update1} and \ref{eq:alpha-induce_update2}.

We initialized all delay parameters to zero, and all Gaussian process timescale parameters to the same value, equal to twice the sampling period or spike count bin width of the neural activity.
All Gaussian process noise variances were fixed to small values ($10^{-3}$), as in \citet{gokcen_uncovering_2023}.
Finally, for mDLAG-inducing specifically, we specified the number of inducing points $T_{\text{ind}}$, and fixed the inducing points on a uniformly spaced grid, with the first and last inducing points fixed at the beginning and end of each trial, i.e., $\xi_1 = 1$ and $\xi_{T_{\text{ind}}} = T$.
Fitting via all three methods proceeded iteratively, until their respective lower bounds improved from one iteration to the next by less than a present tolerance (here we used $10^{-8}$; see Appendices \ref{app:mdlag-induce_fit} and \ref{app:mdlag-freq_fit}).

\subsection{Choosing the Number of Significant Latents in Each Group}
\label{sec:model_selection}

mDLAG incorporates automatic relevance determination (ARD) to automatically determine, during model fitting, both the total number of latents and the subset of groups that each latent involves.
We sought an intuitive measure of the significance of each latent variable within a population, post-fitting, based on the amount of shared variance each latent explains.
The shared variance of latent $j$ in group $m$ is given by $\langle 
\lVert \mathbf{c}_j^m \rVert_2^2 \rangle$, the expected squared magnitude of the $j$\textsuperscript{th} column of the loading matrix $C^m$, with respect to the posterior distribution $Q_c(C)$.
Since the total shared variance can vary widely across observed groups, we considered a normalized metric, the fraction of shared variance explained by latent $j$ in group $m$:
\begin{equation}
    \nu_j^m = \frac{\langle 
\lVert \mathbf{c}_j^m \rVert_2^2 \rangle}{\sum_{k=1}^p \langle 
\lVert \mathbf{c}_k^m \rVert_2^2 \rangle}
\end{equation}
For small ARD hyperparameters, $a_\alpha$ and $b_\alpha$ (as we have chosen in this work), the fraction of shared variance can equivalently be computed using the estimated ARD parameters (see for example equations \ref{eq:alpha-induce_update1} and \ref{eq:alpha-induce_update2} of Appendix~\ref{app:mdlag-induce_fit}):
\begin{equation}
    \nu_j^m \approx \frac{\langle \alpha_j^m \rangle^{-1}}{\sum_{k=1}^p \langle \alpha_k^m \rangle^{-1}} \label{eq:varexp}
\end{equation}
If a latent does not significantly explain activity in a group, then $\nu_j^m$ will be close to zero.
Throughout this work, across all methods considered, we reported a latent as significant in a group if it explained at least 2\% of the shared variance within that group ($\nu_j^m \geq 0.02$).

\subsection{Neural Recordings}
\label{sec:npx_methods}

Animal procedures have been reported in previous work \citep{smith_spatial_2008, zandvakili_coordinated_2015}.
Briefly, animals (\textit{Macaca fascicularis}, 2--5 years old) were anesthetized with ketamine (10 mg kg\textsuperscript{-1}) and maintained on isoflurane during surgery.
All recordings were performed under sufentanil (6-18 $\mu$g kg\textsuperscript{-1} hr\textsuperscript{-1}) anesthesia.
Vecuronium bromide (150 $\mu$g kg\textsuperscript{-1} hr\textsuperscript{-1}) was used to prevent eye movements. 
All procedures were approved by the Institutional Animal Care and Use Committee of the Albert Einstein College of Medicine.

Recordings were performed using up to four Neuropixels 1.0 (IMEC, Belgium) probes spaced primarily in the mediolateral direction, with the most anterior probe placed roughly 3 mm posterior to the lunate sulcus. 
After the probes were lowered, the brain surface was covered with agar or Dura-Gel (Cambridge Neurotech LTD, United Kingdom) to prevent desiccation.
We carried out a total of seven recording sessions from three animals.
In one animal, we recorded from both hemispheres (in separate sessions).
For the analysis of each recording session, we excluded neurons that fired fewer than 0.5 spikes s\textsuperscript{-1}, on average, across all trials and all grating orientations, or exhibited a Fano factor greater than 5.
Average analyzed population sizes per recording session were 125 for V1 (range 55--191), 124 for V2 (range 63--166), and 53 for V3d (range 8--111).

\subsection{Bias Mitigation: Tapering}
\label{sec:taper_method}

Tapering is a well established technique for bias mitigation in spectral estimation \citep[Chapter 7]{priestley_spectral_1981}.
In this work (Section~\ref{sec:taper}, \fig{bias_params}, \fig{bias_dim}{c}), we used tapering as a preprocessing step to bring observed time series (in training data only; we left test data unaltered to faithfully measure performance) in line with the periodic boundary conditions assumed by the frequency domain approach.
Toward that end, let $v_t \in \mathbb{R}$, $t = 1,\ldots,T$, be a set of weights, to be applied to a time series of length $T$.
We chose the commonly used periodic Hamming window function to compute each weight according to
\begin{equation}
    v_t = 0.54 - 0.46 \cos \left( 2 \pi \cdot \frac{t-1}{T} \right)
\end{equation}
The act of tapering then involves the multiplication of each weight $v_t$ with each $y_{n,r,t} \in \mathbb{R}$, the observation at time $t$ on trial $n$ of unit $r$ (here group identity does not matter, and we suppress any dependency).
The same weights $v_t$ are used for all trials and units.

We took additional care to preserve the sample mean and variance (over time points and trials) of each unit's activity pre- and post-tapering.
Let $\mu_{r} \in \mathbb{R}$ be the sample mean of the observations for unit $r$, and let $\sigma_{r} \in \mathbb{R}_{>0}$ be its standard deviation.
Then we apply the taper to normalized observations as follows:
\begin{equation}
    y^{\prime}_{n,r,t} = v_t \cdot \frac{y_{n,r,t} - \mu_r}{\sigma_r}
\end{equation}
The sample mean and standard deviation (over time points and trials) of the modified observations $y^{\prime}_{n,r,t}$, $\mu^{\prime}_{r}$ and $\sigma^{\prime}_{r}$, are no longer the same as the original sample mean and standard deviation, $\mu_{r}$ and $\sigma_{r}$.
We restore these original statistics with an additional processing step:
\begin{equation}
    y^{\prime\prime}_{n,r,t} = \frac{\sigma_r}{\sigma^{\prime}_{r}} \cdot (y^{\prime}_{n,r,t} - \mu^{\prime}_r) + \mu_r
\end{equation}
Models were then fit via mDLAG-frequency to the fully preprocessed observations $y^{\prime\prime}_{n,r,t}$.

\clearpage
\subsection*{Reproducibility, Code Availability, and Data Availability}
\label{sec:code}

All numerical and statistical analyses described in this work were performed in MATLAB (MathWorks; version 2024a).
Numerical results were obtained on a Red Hat Enterprise Linux machine (release 7.9, 64-bit) with 219.68 GB of RAM, on an Intel Xeon Gold 6140 CPU (2.3 GHz).
An implementation of mDLAG-time is currently available on \href{http://github.com/egokcen/mDLAG}{GitHub} and on \href{https://doi.org/10.5281/zenodo.10048163}{Zenodo} \citep{gokcen_mdlag_2023}.
Implementations of mDLAG-inducing and mDLAG-frequency (and a frequency domain fitting approach to the non-Bayesian DLAG) will be made available upon publication.
Code and data to reproduce figures, including any reported $p$-values, will also be made available.

\subsection*{Acknowledgements}

We thank Alison Xu for assistance with data collection.
This work was supported by the Dowd Fellowship (E.G.), ECE Benjamin Garver Lamme/Westinghouse Fellowship (E.G.), Simons Collaboration on the Global Brain 542999 and 2794-08 (A.K.), 543009 and 2794-04 (C.K.M.), 543065 and 3241-05 (B.M.Y.), NIH R01 EY035896 (B.M.Y., A.K., C.K.M.), NIH RF1 NS127107 (A.K., C.K.M., B.M.Y.), NSF NCS DRL 2124066 (B.M.Y.), NIH R01 NS129584 (B.M.Y.).

\subsection*{Author Contributions}

E.G.: conceptualization, data curation (simulations), formal analysis, investigation, methodology, software, validation, visualization, writing -- original draft.
A.I.J.: data curation (Neuropixels recordings), resources (Neuropixels recordings), writing -- review and editing.
A.K.: data curation (Neuropixels recordings), funding acquisition, resources (Neuropixels recordings), supervision, writing -- review and editing.
C.K.M.: funding acquisition, supervision, writing -- review and editing.
B.M.Y: funding acquisition, resources (computing), supervision, writing -- review and editing.

\subsection*{Competing Interests}

The authors declare no competing interests.

{

}

\clearpage

\setcounter{figure}{0}
\setcounter{table}{0}
\renewcommand{\figurename}{Supplementary Figure}

\section*{Supplementary Tables and Figures}

\suppfigure{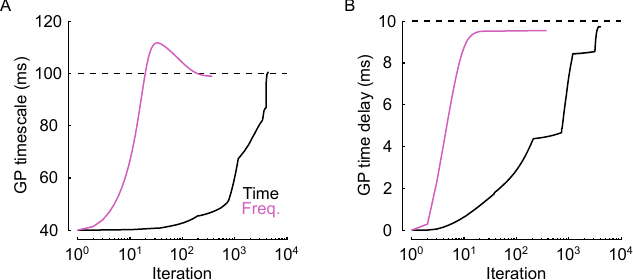}
\suppcaption{param_progress_supp}{Progress of Gaussian process (GP) parameter estimates with each fitting iteration (related to \fig{scalingT}{c}, an example run with $T = 500$).}{\panel{a} GP timescale progress with each fitting iteration.
\panel{b} GP time delay progress with each fitting iteration.
Black: mDLAG-time (Time); magenta: mDLAG-frequency (Freq.).
Black dashed lines indicate ground truth parameter values.
GP parameter estimates for mDLAG-frequency converge toward the optimum in significantly fewer iterations than the GP parameter estimates for mDLAG-time, partially explaining the faster convergence of the mDLAG-frequency method.}
\clearpage

\suppfigure{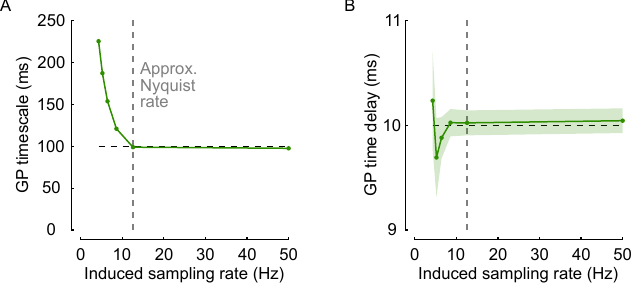}
\suppcaption{inducing_alias_supp}{Effects of induced sampling rate (determined by the chosen number of inducing points) on mDLAG-inducing performance.}{For the same simulated data as in \fig{scalingT}{}, we chose a fixed trial length ($T = 163$ time points), and then re-fit mDLAG-inducing models with varying numbers of inducing points ($T_{\text{ind}} \in \{14, 17, 21, 28, 41, 163\}$).
Each number of inducing points corresponded to a different induced sampling rate.
\panel{a} GP timescale estimates versus induced sampling rate.
\panel{b} GP time delay estimates versus induced sampling rate.
Horizontal black dashed lines indicate ground truth parameter values.
Solid traces indicate the mean of estimates, and shading (where visible) indicates standard error of the mean, computed across 20 runs.
Performance degrades below an induced sampling rate of 12-13 Hz, the approximate Nyquist rate of a squared exponential GP with 100 ms timescale (vertical gray dashed lines).}
\clearpage

\suppfigure{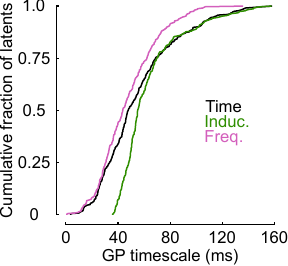}
\suppcaption{npx_timescales_supp}{Empirical cumulative distribution functions (eCDFs) of estimated GP timescales across all Neuropixels datasets.}{Black: mDLAG-time (Time); green: mDLAG-inducing (Induc.); magenta: mDLAG-frequency (Freq.).
In contrast with mDLAG-time and mDLAG-frequency, mDLAG-inducing (with $T_{\text{ind}} = 20$ inducing points) did not capture fast timescales.
For instance, by mDLAG-time estimates, about 25\% of the GP timescales encountered across Neuropixels datasets were 35 ms or shorter (34\%, according to mDLAG-frequency).
No mDLAG-inducing timescale estimates were shorter than 35 ms.
This discrepancy in estimated timescales explains the discrepancy in performance between mDLAG-inducing and the other two methods (\fig{npx}{b}).
eCDFs are computed over 324, 309, and 323 significant latents for mDLAG-time, mDLAG-inducing, and mDLAG-frequency, respectively.
These numbers are pooled across all 14 Neuropixels datasets.}
\clearpage

\suppfigure{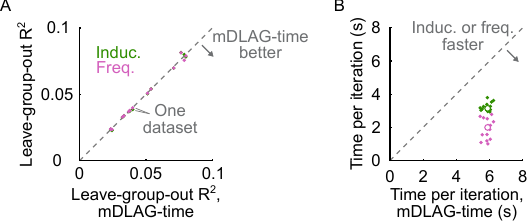}
\suppcaption{npx_supp_S32}{Neuropixels results (related to \fig{npx}{}), but with $T_{\text{ind}} = 32$ inducing points for mDLAG-inducing, rather than $T_{\text{ind}} = 20$.}{\panel{a} Performance (leave-group-out $R^2$) of mDLAG-inducing (green points; Induc.) or mDLAG-frequency (magenta points; Freq.) versus performance of mDLAG-time.
Each data point represents one dataset.
For each dataset, performance was evaluated on 75 held-out test trials.
In \fig{npx}{b}, mDLAG-inducing (with $T_{\text{ind}} = 20$ inducing points) was outperformed by both mDLAG-time and mDLAG-frequency.
Here, mDLAG-inducing (with $T_{\text{ind}} = 32$ inducing points) performed similarly to the other methods (one-sided paired sign tests: mDLAG-time better than mDLAG-inducing, $p = 0.3953$; mDLAG-frequency better than mDLAG-inducing, $p = 0.0898$).
\panel{b} Mean clock time per iteration, mDLAG-inducing or mDLAG-frequency versus mDLAG-time.
As in panel \panel{a}, each data point represents one dataset.
White-filled circles indicate the median of the green points (circle with green border) or of the magenta points (circle with magenta border).
In \fig{npx}{c-e}, mDLAG-inducing (with $T_{\text{ind}} = 20$ inducing points) gave a similar runtime (and speed-up over mDLAG-time) to mDLAG-frequency.
Here, while mDLAG-inducing (with $T_{\text{ind}} = 32$ inducing points) still provides a significant speed-up over mDLAG-time, mDLAG-frequency is now 55\% faster than mDLAG-inducing per iteration, and 72\% faster overall.
Median clock time per iteration across datasets: mDLAG-time, 5.88 s; mDLAG-inducing, 3.15 s; mDLAG-frequency, 2.00 s.
Median number of iterations across datasets (not shown): mDLAG-time, 30,737; mDLAG-inducing, 3,221; mDLAG-frequency, 3,659; for mDLAG-time, we cut off fitting at 50,000 iterations.
Median total clock time across datasets (not shown): mDLAG-time, 50.4 hrs; mDLAG-inducing, 2.9 hrs; mDLAG-frequency, 1.8 hrs.}
\clearpage

\supptablecaption{simulated-summary-supp}{Summary of Supplementary Simulated Datasets.}{}
\begin{table}[ht!]
  \footnotesize
  \caption{}
  \label{tab:simulated-summary-supp}
  \centering
  \begin{tabular}{p{3.2cm}p{1.5cm}p{1.5cm}}
    \toprule
    Dataset     & Sampling rate    & Scaling, $q$ \\
    \midrule
    Independent runs    & 10    & 5 \\
    Training trials    & 100    & 100 \\
    Time points    & 7 -- 100    & 100 \\
    Samp. period (ms)     & 10 -- 160    & 20 \\
    Groups    & 2    & 2 \\
    Units per group    & 12    & 10 -- 500 \\
    Latents    & 1    & 1 \\
    GP timescales (ms)    & 100    & 100 \\
    GP time delays (ms)    & 10    & 10 \\
    Signal-to-noise ratio    & 0.2    & 0.2 \\
    \bottomrule
  \end{tabular}
\end{table}
\clearpage

\suppfigure{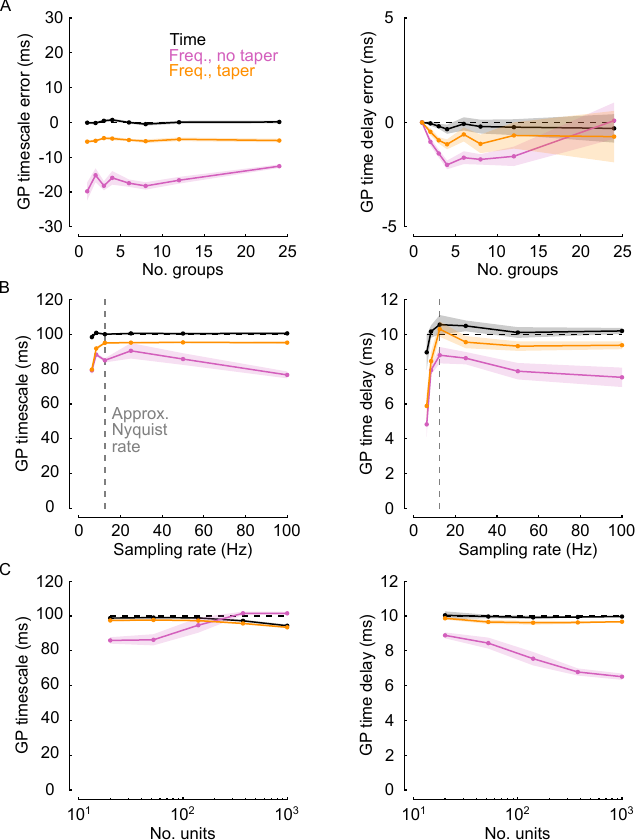}
\suppcaption{bias_params_supp}{Additional characterizations of bias in Gaussian process (GP) parameter estimation.}{\panel{a} Estimation bias as a function of number of groups (same data as in \fig{scalingM}{}; \tbl{simulated-summary}, Scaling, $M$).
Left: GP timescale estimation error versus number of groups.
Right: GP time delay estimation error versus number of groups.
Black dashed lines indicate zero error.
Solid traces indicate the mean of estimates, and shading (where visible) indicates standard error of the mean, computed across 20 runs.
Black: mDLAG-time (Time); magenta: mDLAG-frequency without tapering (Freq., no taper); gold: mDLAG-frequency with tapering (Freq., taper).
Group number has no consistent effect on estimation bias.
Tapering (gold) was an effective bias mitigation strategy.
\panel{b} Estimation bias as a function of sampling rate.
We generated 10 additional datasets (\supptbl{simulated-summary-supp}, Sampling rate; see also Section~\ref{sec:data_generation}), where each trial was $1,000$ ms in length, sampled at 1 ms resolution.
We then fit mDLAG models to these data, downsampled to different rates (6.25 -- 100 Hz).
Regardless of downsampling, trials always spanned $1,000$ ms.
Left: GP timescale estimates versus sampling rate.
Right: GP time delay estimates versus sampling rate.
Solid traces indicate the mean of estimates, and shading (where visible) indicates standard error of the mean, computed across 10 runs.
Below the approximate Nyquist rate for the ground truth latent signals (vertical gray dashed lines; 12 -- 13 Hz for a squared exponential GP with 100 ms timescale), all methods exhibit a decline in performance, as sampling rate decreases, due to aliasing.
mDLAG-frequency (magenta) increasingly underestimated both GP timescales and time delays as sampling rates significantly surpassed the approximate Nyquist rate.
Increasingly fast sampling rates likely exacerbate edge effects due to finite trial length and the periodic boundary conditions of the frequency domain approach.
Discontinuities between the beginning and ends of each trial could introduce spurious high-frequency content.
Tapering (gold), which attenuates observations at the beginning and ends of each trial toward zero, was an effective bias mitigation strategy, eliminating this trend with increasing sampling rate.
\panel{c} Estimation bias as a function of number of units (neurons).
We generated 5 additional datasets (\supptbl{simulated-summary-supp}, Scaling, $q$; see also Section~\ref{sec:data_generation}), comprising $M = 2$ groups with $q_m = 500$ units each.
We then fit mDLAG models to observations from increasing subsets of units per group (10 to 500).
Left: GP timescale estimates versus number of units (the total across both groups).
Right: GP time delay estimates versus number of units.
Solid traces indicate the mean of estimates, and shading (where visible) indicates standard error of the mean, computed across 5 runs.
Increasing the number of units had distinct effects on GP timescale estimation and GP time delay estimation.
GP timescales were underestimated for smaller numbers of units (20 -- 140), but slightly overestimated for larger number of units (188 -- 500).
GP time delays were increasingly underestimated with increasing number of units.
It has been reported previously that methods that incorporate Whittle quasi-likelihoods can experience increased bias in high-dimensional regimes \citep{guillaumin_debiased_2022}.
Tapering (gold) was an effective bias mitigation strategy.}
\clearpage

\suppfigure{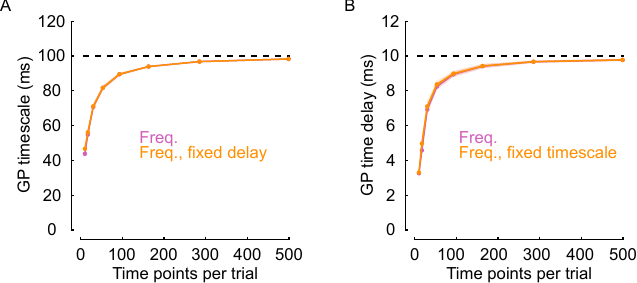}
\suppcaption{bias_params_supp_fix}{Biases in Gaussian process (GP) timescale estimation and GP time delay estimation are decoupled phenomena.}{\panel{a} GP timescale estimates versus number of time points per trial. Magenta: mDLAG-frequency estimates with all parameters freely estimated (Freq.; same as in \fig{bias_params}{a}). Gold: mDLAG-frequency estimates with GP time delays fixed at the ground truth value (Freq., fixed delay).
\panel{b} Magenta: mDLAG-frequency estimates with all parameters freely estimated (Freq.; same as in \fig{bias_params}{b}). Gold: mDLAG-frequency estimates with GP timescales fixed at the ground truth value (Freq., fixed timescale).
Across panels \panel{a} and \panel{b}, black dashed lines indicate ground truth parameter values.
Solid traces indicate the mean of estimates, and shading (where visible) indicates standard error of the mean, computed across 20 runs.
Regardless of which parameter was held fixed, the bias remained in estimates of the other parameter, suggesting that these biases are decoupled effects.}
\clearpage

\suppfigure{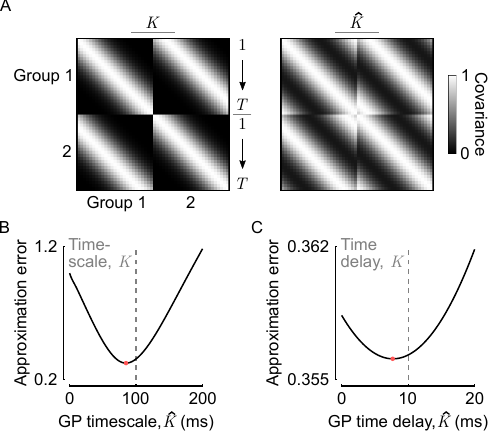}
\suppcaption{cov_error_supp}{Understanding biases in Gaussian process (GP) parameter estimation as a covariance matrix approximation problem.}{\panel{a} Left: An example GP covariance matrix ($K$), corresponding to the mDLAG-time model.
Right: The corresponding circulant approximation of $K$ ($\hat{K}$), arising implicitly from the mDLAG-frequency model.
Both $K$ and $\hat{K}$ shown here were generated with the same squared exponential GP timescale (100 ms) and time delay (80 ms), for trial length $T = 25$ time points (with 20 ms sampling period) and $M = 2$ groups.
\panel{b} Approximation error between the covariance matrix $K$ and its circulant approximation $\hat{K}$ as a function of the GP timescale used to generate $\hat{K}$.
Gray dashed line: The ``ground truth'' GP timescale used to generate $K$.
Red point: The GP timescale of the circulant approximation $\hat{K}$ that minimizes approximation error.
\panel{c} Approximation error between the covariance matrix $K$ and its circulant approximation $\hat{K}$ as a function of the GP time delay used to generate $\hat{K}$.
Gray dashed line: The ``ground truth'' GP time delay used to generate $K$.
Red point: The GP time delay of the circulant approximation $\hat{K}$ that minimizes approximation error.
For panels \panel{b} and \panel{c}, matrices were generated with $T = 25$ time points and $M = 2$ groups.
Approximation error was computed as the normalized Frobenius norm between $K$ and $\hat{K}$, $\lVert K - \hat{K} \rVert_{\text{F}} / \lVert K \rVert_{\text{F}}$.}
\clearpage

\suppfigure{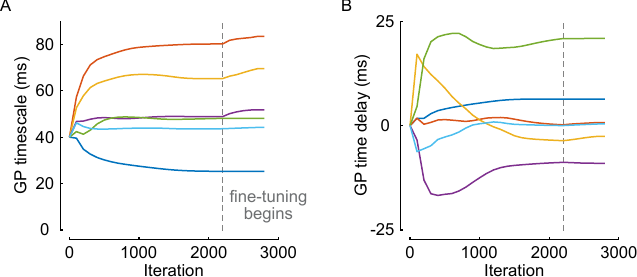}
\suppcaption{npx_finetune_supp}{Using mDLAG-time to fine-tune a model fit by mDLAG-frequency (example Neuropixels dataset).}{\panel{a} GP timescale progress with each fitting iteration.
\panel{b} GP time delay progress with each fitting iteration.
Each colored trace corresponds to one latent variable.
Colors are consistent across panels, so that purple represents the same latent variable in panel \panel{a} and \panel{b}, and so on.
For visual clarity, we show only latents found to be significantly shared between areas V1 and V2.
Vertical dashed lines indicate the iteration at which mDLAG-frequency converged and fine-tuning with mDLAG-time began.
Fine-tuning then took place for another 500 iterations, at which point GP parameters settled into slightly different optimal values.
After 500 iterations of fine-tuning with mDLAG-time, GP timescales and time delays settled into slightly different values from the values at which mDLAG-frequency converged.
Fine-tuning is a straightforward strategy to mitigate parameter estimation biases.}
\clearpage

\appendix

\setcounter{table}{0}

\section*{Appendix}

\section{Mathematical Notation}
\label{app:notation}

To disambiguate each variable or parameter across the three models we consider here, we need to keep track of up to four labels that indicate their associated (1) trial; (2) unit (neuron) or latent variable index; (3) sample in time or frequency; or (4) group or subpopulation (\apptbl{notation-shape}).
We indicate the first three labels via subscripts.
Trials are indexed by $n = 1,\ldots,N$.
Units (neurons) are indexed by $r = 1,\ldots,q$, and latent variables (latents) are indexed by $j = 1,\ldots,p$.
In the time domain, we index a trial of length $T$ samples by $t = 1,\ldots,T$.
In the frequency domain, we must also consider $T$ frequency components, but to emphasize the indexing over frequency rather than time, we switch to the subscript $l = 1,\ldots,T$.
Where relevant, we indicate the group (subpopulation) to which a variable or parameter pertains via $m = 1,\ldots,M$ (most commonly a superscript).

Putting these four labels together, we define the observed activity of unit $r$ (out of $q_m$) in group $m$ at time $t$ on trial $n$ as $y^m_{n,r,t} \in \mathbb{R}$ (\apptbl{notation-time}).
Similarly, we define latent $j$ (out of $p$) in group $m$ at time $t$ on trial $n$ as $x^m_{n,j,t} \in \mathbb{R}$.
To indicate a collection of all variables along a particular index, we replace that index with a colon.
Hence we represent the simultaneous activity of $q_m$ units observed in group $m$ at time $t$ on trial $n$ as the vector $\mathbf{y}^m_{n,:,t} \in \mathbb{R}^{q_m}$.
Similarly, we represent the collection of all $p$ latents in group $m$ at time $t$ on trial $n$ as the vector $\mathbf{x}^m_{n,:,t} \in \mathbb{R}^p$.
For concision, where a particular index is either not applicable or not immediately relevant, we omit it.
The identities of the remaining indices should be clear from context.
For example, we might rewrite $\mathbf{y}^m_{n,:,t}$ as $\mathbf{y}^m_{n,t}$.

It is conceptually helpful to understand the above notation as taking cross-sections of three-dimensional arrays (\apptbl{notation-time} -- \apptbl{notation-freq}).
For example, observed activity in group $m$ on trial $n$ can be collected into the matrix (two-dimensional array) $Y^m_n = [\mathbf{y}^m_{n,1} \cdots \mathbf{y}^m_{n,T}] \in \mathbb{R}^{q_m \times T}$.
Hence each $\mathbf{y}^m_{n,t}$ is a column of $Y^m_n$.
Then we can form the three-dimensional array $Y^m$ by concatenating the matrices $Y^m_1,\ldots,Y^m_N$ across trials along a third dimension.
Similarly, the latents in group $m$ on trial $n$ can be collected into the matrix $X^m_n = [\mathbf{x}^m_{n,:,1} \cdots \mathbf{x}^m_{n,:,T}] \in \mathbb{R}^{p \times T}$.
We represent a row of $X^m_n$ (i.e., the values of a single latent $j$ at all time points on trial $n$) as $\mathbf{x}^m_{n,j,:} \in \mathbb{R}^T$.
Then we can form the three-dimensional array $X^m$ by concatenating the matrices $X^m_1,\ldots,X^m_N$ across trials along a third dimension.

For all variables with a frequency domain counterpart, we indicate that counterpart using a tilde (\apptbl{notation-freq}).
The complex-valued $\widetilde{y}^m_{n,r,l} \in \mathbb{C}$ is thus the activity of unit $r$ in group $m$ at frequency $l$ on trial $n$.
Similarly, $\widetilde{x}_{n,j,l} \in \mathbb{C}$ is the value of latent $j$ at frequency $l$ on trial $n$ (in the frequency domain, we do not assign latents to each group $m$, see Section~\ref{sec:mdlag-freq}).
We collect these frequency domain quantities into structures analogous to their time domain counterparts, for example, $\widetilde{\mathbf{y}}^m_{n,l} \in \mathbb{C}^{q_m}$ as the analog to $\mathbf{y}^m_{n,t}$, defined above.

\apptablecaption{notation-shape}{Data Shape, Indices, and Other Constants.}{}
\begin{table}[ht!]
  \footnotesize
  \caption{}
  \label{tab:notation-shape}
  \centering
  \begin{tabular}{p{0.5in}p{3.4in}}
    \toprule
    Symbol     & Description  \\
    \midrule
    $N$    & Number of trials  \\
    $n$    & Index over trials, $n = 1,\ldots,N$ \\
    $T$    & Number of time points (or frequencies) per trial  \\
    $t$    & Index over time points or inducing points, $t = 1,\ldots,T$  \\
    $l$    & Index over frequency components, $l = 1,\ldots,T$  \\
    $f_l$  & The real-valued frequency of component $l$  \\
    $M$    & Number of observed groups  \\
    $m$    & Index over groups, $m = 1,\ldots,M$  \\
    $q_m$  & Number of units (neurons) in group $m$  \\
    $q$    & Total number of units across all groups, $\sum_m q_m$  \\
    $r$    & Index over units, $r = 1,\ldots,q_m$  \\
    $p$    & Number of latent variables (same for all groups)  \\
    $j$    & Index over latent variables, $j = 1,\ldots,p$  \\
    $i$    & The imaginary number, $\sqrt{-1}$  \\
    \bottomrule
  \end{tabular}
\end{table}
\clearpage

\apptablecaption{notation-time}{mDLAG-time and Symbols Common Across Methods.}{}
\begin{table}[ht!]
  \footnotesize
  \caption{}
  \label{tab:notation-time}
  \centering
  \begin{tabular}{p{0.5in}p{3.4in}}
    \toprule
    Symbol     & Description  \\
    \midrule
    $Y^m_n$    & $q_m \times T$ matrix of time domain observations in group $m$ on trial $n$  \\
    $\mathbf{y}^m_{n,t}$    & $q_m \times 1$ vector of observations in group $m$ at time $t$ on trial $n$; the $t$\textsuperscript{th} column of $Y^m_n$ \\
    $\mathbf{y}^m_{n,r,:}$  & $T \times 1$ vector of observations of unit $r$ in group $m$ over time on trial $n$; the $r$\textsuperscript{th} row of $Y^m_n$  \\
    $y^m_{n,r,t}$    & The activity of unit $r$ in group $m$ at time $t$ on trial $n$; the $r$\textsuperscript{th} element of $\mathbf{y}^m_{n,t}$  \\
    $X^m_n$    & $p \times T$ matrix of time domain latents in group $m$ on trial $n$  \\
    $X_n$    & The collection, across groups, of time domain latents on trial $n$, $\{X^1_n, \ldots, X^M_n\}$  \\
    $\mathbf{x}^m_{n,:,t}$    & $p \times 1$ vector of latents in group $m$ at time $t$ on trial $n$; the $t$\textsuperscript{th} column of $X^m_n$  \\
    $\mathbf{x}^m_{n,j,:}$    & $T \times 1$ vector of values of latent $j$ in group $m$ over time on trial $n$; the $j$\textsuperscript{th} row of $X^m_n$  \\
    $x^m_{n,j,t}$    & The value of latent $j$ in group $m$ at time $t$ on trial $n$; the $j$\textsuperscript{th} element of $\mathbf{x}^m_{n,:,t}$ and the $t$\textsuperscript{th} element of $\mathbf{x}^m_{n,j,:}$  \\
    $C^m$    & $q_m \times p$ loading matrix for group $m$  \\
    $\alpha^m_j$    & automatic relevance determination (ARD) parameter for group $m$ and latent $j$  \\
    $\mathbf{d}^m$    & $q_m \times 1$ mean parameter for group $m$  \\
    $\boldsymbol{\phi}^m$    & $q_m \times 1$ observation noise precision parameter for group $m$, $\boldsymbol{\phi}^m = [\phi^m_1 \cdots \phi^m_{q_m}]^{\top}$  \\
    $\beta$    & precision (hyper)parameter of the Gaussian prior over each mean parameter $\mathbf{d}^m$; fixed to a small value  \\
    $a_{\phi}, \ b_{\phi}$    & shape and rate (hyper)parameters, respectively, of the Gamma prior over each noise precision parameter $\phi^m_r$ for unit $r$ in group $m$; fixed to small values  \\
    $a_{\alpha}, \ b_{\alpha}$    & shape and rate (hyper)parameters, respectively, of the Gamma prior over each ARD parameter $\alpha^m_j$ for group $m$ and latent $j$; fixed to small values  \\
    $D^m_j$   & GP time delay between group $m$ and latent $j$  \\
    $\tau_j$    & GP timescale for latent $j$  \\
    $\sigma_j$  & GP noise parameter for latent $j$; fixed to a small value  \\
    $K_{m_1,m_2,j}$    & $T \times T$ covariance matrix for latent $j$, between groups $m_1$ and $m_2$  \\
    $k_{m_1,m_2,j}$    & covariance function for latent $j$, between groups $m_1$ and $m_2$  \\
    \bottomrule
  \end{tabular}
\end{table}
\clearpage

\apptablecaption{notation-inducing}{mDLAG-inducing.}{}
\begin{table}[ht!]
  \footnotesize
  \caption{}
  \label{tab:notation-inducing}
  \centering
  \begin{tabular}{p{0.5in}p{3.4in}}
    \toprule
    Symbol     & Description  \\
    \midrule
    $T_{\text{ind}}$    & Number of inducing points per trial \\
    $W_n$    & $p \times T_{\text{ind}}$ matrix of inducing variables (common to all groups) on trial $n$  \\
    $\mathbf{w}_{n,j,:}$    & $T_{\text{ind}} \times 1$ vector of inducing variable values for latent $j$ (common to all groups) on trial $n$; the $j$\textsuperscript{th} row of $W_n$  \\
    $w_{n,j,t}$    & Inducing variable for latent $j$ at inducing point $t$ on trial $n$; the $t$\textsuperscript{th} element of $\mathbf{w}_{n,j,:}$  \\
    $\xi_t$    & inducing point; the real-valued time at which inducing variable $w_{n,j,t}$ is defined  \\
    $K^w_{j}$    & $T_{\text{ind}} \times T_{\text{ind}}$ inducing variable covariance matrix for latent $j$  \\
    $k^w_{j}$    & covariance function for the inducing variable of latent $j$  \\
    $K^{xw}_{j}$    & $MT \times T_{\text{ind}}$ covariance matrix between latent $j$ and its inducing variable  \\
    $k^{xw}_{j}$    & covariance function between latent $j$ and its inducing variable  \\
    \bottomrule
  \end{tabular}
\end{table}

\apptablecaption{notation-freq}{mDLAG-frequency.}{}
\begin{table}[ht!]
  \footnotesize
  \caption{}
  \label{tab:notation-freq}
  \centering
  \begin{tabular}{p{0.5in}p{3.4in}}
    \toprule
    Symbol     & Description  \\
    \midrule
    $\widetilde{Y}^m_n$    & $q_m \times T$ matrix of frequency domain observations in group $m$ on trial $n$  \\
    $\widetilde{\mathbf{y}}^m_{n,l}$    & $q_m \times 1$ vector of observations in group $m$ at frequency $l$ on trial $n$; the $l$\textsuperscript{th} column of $\widetilde{Y}^m_n$  \\
    $\widetilde{\mathbf{y}}^m_{n,r,:}$  & $T \times 1$ vector of observations of unit $r$ in group $m$ across frequencies on trial $n$; the $r$\textsuperscript{th} row of $\widetilde{Y}^m_n$  \\
    $\widetilde{y}^m_{n,r,l}$    & The value of unit $r$ in group $m$ at frequency $l$ on trial $n$; the $r$\textsuperscript{th} element of $\widetilde{\mathbf{y}}^m_{n,l}$ and the $l$\textsuperscript{th} element of $\widetilde{\mathbf{y}}^m_{n,r,:}$  \\
    $\widetilde{X}_n$    & $p \times T$ matrix of frequency domain latents (common to all groups) on trial $n$  \\
    $\widetilde{\mathbf{x}}_{n,:,l}$    & $p \times 1$ vector of latents (common to all groups) at frequency $l$ on trial $n$; the $l$\textsuperscript{th} column of $\widetilde{X}_n$  \\
    $\widetilde{\mathbf{x}}_{n,j,:}$    & $T \times 1$ vector of values of latent $j$ (common to all groups) across frequencies on trial $n$; the $j$\textsuperscript{th} row of $\widetilde{X}_n$  \\
    $\widetilde{x}_{n,j,l}$    & The value of latent $j$ at frequency $l$ on trial $n$; the $j$\textsuperscript{th} element of $\widetilde{\mathbf{x}}_{n,:,l}$ and the $l$\textsuperscript{th} element of $\widetilde{\mathbf{x}}_{n,j,:}$  \\
    $\widetilde{\mathbf{d}}^m_l$    & shorthand for $\delta_{l-1} \cdot \sqrt{T} \mathbf{d}^m$: $\sqrt{T} \mathbf{d}^m$ for the zero frequency, and $\mathbf{0}$ otherwise  \\
    $H_{j}^m$    & $T \times T$ diagonal phase-shift operator matrix, from latent $j$ to group $m$  \\
    $H_{l}^m$    & $p \times p$ diagonal phase-shift operator matrix at frequency $l$, from latents to group $m$  \\
    $S_{j}$    & $T \times T$ diagonal power spectral density (PSD) matrix for latent $j$  \\
    $S_{l}$    & $p \times p$ diagonal PSD matrix at frequency $l$  \\
    $s_{j}$    & PSD function for latent $j$  \\
    $U_T$    & $T \times T$ unitary discrete Fourier transform matrix  \\
    \bottomrule
  \end{tabular}
\end{table}
\clearpage

\section{Posterior Inference and Fitting via mDLAG-inducing}
\label{app:mdlag-induce_fit}

\subsection{Variational Inference}
\label{app:mdlag-induce_updates}

Let $Y$, $X$, and $W$ be collections of all observed variables, latent variables, and inducing variables, respectively, across all time points (or inducing points) and trials.
Similarly, let $\mathbf{d}$, $\boldsymbol{\phi}$, $C$, $\mathcal{A}$, $\tau$, and $D$ be collections of the mean parameters, noise precisions, loading matrices, ARD parameters, GP timescales, and time delays, respectively.
From the observed activity, we seek to estimate posterior distributions over $X$, $W$, and the remaining probabilistic model components
\begin{equation}
    \theta = \left\{\mathbf{d}, \ \boldsymbol{\phi},\ C, \ \mathcal{A}\right\} \label{eq:mdlag-induce_theta}
\end{equation}
and point estimates of the deterministic GP parameters $\Omega = \left\{\tau, \ D \right\}$.

In the case of methods like GPFA \citep{yu_gaussian-process_2009} and DLAG \citep{gokcen_disentangling_2022}, the linear-Gaussian structure of the model enables an exact expectation-maximization (EM) algorithm.
With the introduction of prior distributions over model parameters (to enable automatic relevance determination), the family of mDLAG models, including mDLAG-inducing, loses this property.
The complete likelihood of the mDLAG-inducing model (\fig{method_intro}{c}),
\begin{align}
    P(Y,X,W,\theta | \Omega) &= P(\mathbf{d}) P(\boldsymbol{\phi}) P(C | \mathcal{A}) P(\mathcal{A}) P(Y | X, C, \mathbf{d}, \boldsymbol{\phi}) P(X|W,\Omega) P(W|\tau) \notag\\
    &= \prod_{m=1}^M \Biggl[ P(\mathbf{d}^m) \Biggl[ \prod_{r=1}^{q_m} P(\phi^m_r) \Biggr] \Biggl[ \prod_{j=1}^{p} P(\mathbf{c}_j^m \ | \ \alpha_j^m) P(\alpha_j^m) \Biggr] \notag\\ 
    &\quad \cdot \Biggl[ \prod_{n=1}^{N} \prod_{t=1}^{T} P(\mathbf{y}^m_{n,t} \ | \ \mathbf{x}^m_{n,t}, C^m, \mathbf{d}^m, \boldsymbol{\phi}^m) \Biggr] \Biggr] \notag\\
    &\quad \cdot \prod_{n=1}^{N} \prod_{j=1}^{p} \Biggl[ P\bigl(\mathbf{x}_{n,j,:} \ | \ \mathbf{w}_{n,j,:}, \tau_j, \{ D^m_j \}_{m=1}^{M}\bigr) P\bigl(\mathbf{w}_{n,j,:} \ | \ \tau_j\bigr) \Biggr] \label{eq:mdlag-induce_completeLL}
\end{align}
is no longer Gaussian.
Then a hypothetical EM E-step (evaluation of the posterior distribution $P(X,W,\theta|Y,\Omega)$) becomes prohibitive, as it relies on the analytically intractable marginalization of equation \ref{eq:mdlag-induce_completeLL} with respect to $X$, $W$, and $\theta$.

We therefore employ a variational inference scheme \citep{bishop_variational_1999,klami_group_2015,gokcen_uncovering_2023}, in which we maximize the lower bound $L(Q,\Omega)$ to the log-likelihood $\log P(Y)$, with respect to the approximate posterior distribution $Q(X,W,\theta)$ and the deterministic parameters $\Omega$:
\begin{equation}
    \log P(Y) \geq L(Q,\Omega) = \mathbb{E}_Q[\log P(Y,X,W,\theta | \Omega)] - \mathbb{E}_Q[\log Q(X,W,\theta)] \label{eq:lb_induce}
\end{equation}
We constrain $Q(X,W,\theta)$ so that it factorizes over $X$, $W$, and the elements of $\theta$:
\begin{equation}
    Q(X,W,\theta) = Q_{xw}(X,W) Q_d(\mathbf{d}) Q_{\phi}(\boldsymbol{\phi}) Q_c(C) Q_{\mathcal{A}}(\mathcal{A})
\end{equation}
We then follow \citet{titsias_variational_2009}, and further constrain $Q_{xw}(X,W)$, the joint approximate posterior distribution over the latents and their inducing variables, to factorize as $Q_{xw}(X,W) = P(X | W, \Omega) Q_w(W)$, with generic distribution $Q_w(W)$ over the inducing variables and the conditional prior distribution $P(X | W, \Omega)$ (equation~\ref{eq:mdlag-induce_completeLL}) over the latents themselves.
The final constrained form of $Q(X,W,\theta)$ is thus
\begin{equation}
    Q(X,W,\theta) = P(X | W, \Omega) Q_w(W) Q_d(\mathbf{d}) Q_{\phi}(\boldsymbol{\phi}) Q_c(C) Q_{\mathcal{A}}(\mathcal{A}) \label{eq:q_induce_app}
\end{equation}

This particular choice of factorization leads to an important simplification of the lower bound $L(Q,\Omega)$, ultimately enabling efficient (and closed-form) updates of the latents $X$ and inducing variables $W$ during optimization.
We first rewrite the lower bound by explicitly writing out the definition of the expectation:
\begin{align}
     L(Q,\Omega) &= \mathbb{E}_Q[\log P(Y,X,W,\theta | \Omega)] - \mathbb{E}_Q[\log Q(X,W,\theta)] \\
     &= \int Q(X,W,\theta) \log P(Y,X,W,\theta | \Omega) dX dW d\theta \notag\\
     &\quad - \int Q(X,W,\theta) \log Q(X,W,\theta) dX dW d\theta \\
     &= \int Q(X,W,\theta) \log \frac{P(Y,X,W,\theta | \Omega)}{Q(X,W,\theta)} dX dW d\theta \\
     &= \int Q(X,W,\theta) \notag\\
     &\quad \cdot \log \frac{P(\mathbf{d}) P(\boldsymbol{\phi}) P(C | \mathcal{A}) P(\mathcal{A}) P(Y | X, C, \mathbf{d}, \boldsymbol{\phi}) P(X|W,\Omega) P(W|\tau)}{P(X | W, \Omega) Q_w(W) Q_d(\mathbf{d}) Q_{\phi}(\boldsymbol{\phi}) Q_c(C) Q_{\mathcal{A}}(\mathcal{A})} \notag\\
     &\quad \cdot dX dW d\theta
\end{align}
where in the last line we substituted equations~\ref{eq:mdlag-induce_completeLL} and \ref{eq:q_induce_app} into the numerator and denominator, respectively, of the log term.
Notice the common term $P(X | W, \Omega)$ between the numerator and denominator.
This term cancels, resulting in the following simplified expression for the lower bound:
\begin{align}
     L(Q,\Omega) &= \int Q(X,W,\theta) \notag\\
     &\quad \cdot \log \frac{P(\mathbf{d}) P(\boldsymbol{\phi}) P(C | \mathcal{A}) P(\mathcal{A}) P(Y | X, C, \mathbf{d}, \boldsymbol{\phi}) P(W|\tau)}{Q_w(W) Q_d(\mathbf{d}) Q_{\phi}(\boldsymbol{\phi}) Q_c(C) Q_{\mathcal{A}}(\mathcal{A})} dX dW d\theta
\end{align}
To simplify notation further, define the joint prior distribution \\ $P(W,\theta|\tau) = P(\mathbf{d}) P(\boldsymbol{\phi}) P(C | \mathcal{A}) P(\mathcal{A}) P(W|\tau)$ and the joint posterior distribution $Q_{w\theta}(W,\theta) = Q_w(W) Q_d(\mathbf{d}) Q_{\phi}(\boldsymbol{\phi}) Q_c(C) Q_{\mathcal{A}}(\mathcal{A})$.
Then substituting in these expressions, along with $Q(X,W,\theta) = P(X|W,\Omega) Q_{w\theta}(W,\theta)$, we get
\begin{align}
     L(Q,\Omega) &= \int P(X|W,\Omega) Q_{w\theta}(W,\theta) \log \frac{P(Y | X, C, \mathbf{d}, \boldsymbol{\phi}) P(W,\theta|\tau)}{Q_{w\theta}(W,\theta)} dX dW d\theta \\
     &= \int Q_{w\theta}(W,\theta) \biggl[ P(X|W,\Omega) \cdot \log P(Y | X, C, \mathbf{d}, \boldsymbol{\phi}) dX \biggr] dW d\theta \notag\\
     &\quad + \int Q_{w\theta}(W,\theta) \biggl[  P(X|W,\Omega) \cdot \log \frac{P(W,\theta|\tau)}{Q_{w\theta}(W,\theta)} dX \biggr] dW d\theta
\end{align}

Now let $\mathbb{E}_{P_{x|w}}[\cdot]$ and $\mathbb{E}_{Q_{w\theta}}[\cdot]$ be expectations with respect to the conditional prior distribution $P(X|W,\Omega)$ and joint posterior distribution $Q_{w\theta}(W,\theta)$, respectively.
We can then rewrite the lower bound as
\begin{align}
     L(Q,\Omega) &= \mathbb{E}_{Q_{w\theta}}\bigl[ \mathbb{E}_{P_{x|w}}\bigl[ \log P(Y | X, C, \mathbf{d}, \boldsymbol{\phi}) \bigr] \bigr] \notag\\
     &\quad + \mathbb{E}_{Q_{w\theta}}\biggl[ \mathbb{E}_{P_{x|w}}\biggl[\log \frac{P(W,\theta|\tau)}{Q_{w\theta}(W,\theta)} \biggr] \biggr] \\
     &= \mathbb{E}_{Q_{w\theta}}\bigl[ \mathbb{E}_{P_{x|w}}\bigl[ \log P(Y | X, C, \mathbf{d}, \boldsymbol{\phi}) \bigr] \bigr] + \mathbb{E}_{Q_{w\theta}}\biggl[ \log \frac{P(W,\theta|\tau)}{Q_{w\theta}(W,\theta)} \biggr] \label{eq:lb_induce1}
\end{align}
The term $\mathbb{E}_{P_{x|w}}[ \log P(Y | X, C, \mathbf{d}, \boldsymbol{\phi})]$ integrates out any dependency on the latents $X$, and is therefore only a function of observations $Y$, inducing variables $W$, and parameters in $\theta$ and $\Omega$.
To emphasize these dependencies, we define $\log G(Y,W,\theta | \Omega) = \mathbb{E}_{P_{x|w}}[ \log P(Y | X, C, \mathbf{d}, \boldsymbol{\phi})]$.
Furthermore, note that the second term in equation~\ref{eq:lb_induce1} is the negative KL-divergence between the joint posterior distribution $Q_{w\theta}(W,\theta)$ and the joint prior distribution $P(W,\theta|\tau)$.

We then arrive at the final re-expression of the lower bound:
\begin{align}
     L(Q,\Omega) &= \mathbb{E}_{Q_{w\theta}}\bigl[ \log G(Y,W,\theta|\Omega) \bigr] - \text{KL}\bigl(Q_{w\theta}(W,\theta) \Vert P(W,\theta|\tau) \bigr) \label{eq:lb_induce_final}
\end{align}
This re-expression of the lower bound can then be iteratively maximized via coordinate ascent of the factors of $Q_{w\theta}(W,\theta)$ and the deterministic parameters $\Omega$.
Each factor or deterministic parameter is updated in turn while the remaining factors or parameters are held fixed.
These updates are repeated until the lower bound, which is guaranteed to be non-decreasing, improves from one iteration to the next by less than a present tolerance (here we used $10^{-8}$).

\subsubsection{Posterior Distribution Updates}
\label{sec:mdlag-induce_updates}

Maximizing the lower bound, $L(Q,\Omega)$, with respect to the $k$\textsuperscript{th} factor of $Q_{w\theta}(W,\theta)$, $\hat{Q}_k$, results in the following update rule:
\begin{equation}
    \log \hat{Q}_k = \langle \log G(Y,W,\theta|\Omega) + \log P(W,\theta|\tau) \rangle_{-k} + \text{const.} \label{eq:q_update}
\end{equation}
Here we introduce the notation $\langle \cdot \rangle$ to indicate the expectation, $\mathbb{E}_{Q_{w\theta}}[\cdot]$, with respect to the posterior distribution $Q_{w\theta}(W,\theta)$, and $\langle \cdot \rangle_{-k}$ specifically indicates the expectation with respect to all but the $k$\textsuperscript{th} factor of $Q_{w\theta}(W,\theta)$.

Because of the choice of Gaussian and conjugate Gamma priors in Section~\ref{sec:mdlag-induce}, evaluation of equation \ref{eq:q_update} for each factor $\hat{Q}_k$ leads to factors with the same functional form as their corresponding priors (equations \ref{eq:wprior}, \ref{eq:mdlag_obs1}--\ref{eq:alphaprior}):
\begin{align}
    Q_w(W) &= \prod_{n=1}^N \mathcal{N}(\bar{\mathbf{w}}_n \ | \ \bar{\boldsymbol{\mu}}_{w_n}, \bar{\Sigma}_{w}) \label{eq:qw_app} \\
    Q_{d}(\mathbf{d}) &= \prod_{m=1}^M \mathcal{N}(\mathbf{d}^m \ | \ \boldsymbol{\mu}_d^m, \Sigma_d^m) \label{eq:qd_induce} \\
    Q_{\phi}(\boldsymbol{\phi}) &= \prod_{m=1}^M \prod_{r=1}^{q_m} \Gamma(\phi_r^m \ | \ \bar{a}_{\phi}, \bar{b}_{\phi,r}^m) \label{eq:qphi_induce} \\
    Q_c(C) &= \prod_{m=1}^M \prod_{r=1}^{q_m} \mathcal{N}(\bar{\mathbf{c}}_r^m \ | \ \bar{\boldsymbol{\mu}}_{c_r}^m, \Sigma_{c_r}^m) \label{eq:qc_induce} \\
    Q_{\mathcal{A}}(\mathcal{A}) &= \prod_{m=1}^{M} \prod_{j=1}^{p} \Gamma(\alpha_j^m \ | \ \bar{a}_{\alpha}^m, \bar{b}_{\alpha,j}^m ) \label{eq:qalpha_induce}
\end{align}
Here $\bar{\mathbf{w}}_n \in \mathbb{R}^{pT_{\text{ind}}}$ is a collection of all inducing variables on trial $n$ (see below), and $\bar{\mathbf{c}}_r^m \in \mathbb{R}^p$ is the $r$\textsuperscript{th} row of $C^m$, the loading matrix for group $m$.
Any additional factorization in equations \ref{eq:qw_app}--\ref{eq:qalpha_induce} emerge naturally---we impose only the factorization in equation \ref{eq:q_induce_app}.

To express the updates for $Q_{w}(W)$, let us first define several variables.
First construct $\bar{\mathbf{w}}_n = [\mathbf{w}^{\top}_{n,1,:} \cdots \mathbf{w}^{\top}_{n,p,:}]^{\top} \in \mathbb{R}^{pT_{\text{ind}}}$ by vertically concatenating the inducing variables $\mathbf{w}_{n,j,:}$ for latents $j = 1,\ldots,p$ on trial $n$.
Collect the inducing variable covariance matrices $K^w_j$ for $j = 1,\ldots,p$ (equation~\ref{eq:wprior}; \fig{method_intro}{d}, top) into the block diagonal matrix $\bar{K}^w = \text{diag}(K^w_1, \ldots, K^w_p) \in \mathbb{S}^{pT_{\text{ind}} \times pT_{\text{ind}}}$.
Then, let $\mathbf{k}^{xw}_{m,j,t} \in \mathbb{R}^{T_{\text{ind}}}$ be the $t$\textsuperscript{th} row of $K^{xw}_{m,j} \in \mathbb{R}^{T \times T_{\text{ind}}}$, the $m$\textsuperscript{th} block (for group $m$) of the covariance matrix between latent $j$ and its inducing variable, $K^{xw}_j$ (equation~\ref{eq:xprior_induce}; \fig{method_intro}{d}, bottom).
For time point $t$, collect each $\mathbf{k}^{xw}_{m,j,t}$ for latents $j = 1,\ldots,p$ into the block diagonal matrix $\bar{K}^{xw}_{m,t} = \text{diag}(\mathbf{k}^{xw\top}_{m,1,t},\ldots,\mathbf{k}^{xw\top}_{m,p,t}) \in \mathbb{R}^{p \times pT_{\text{ind}}}$.

Posterior estimates of the inducing variables $W$ are independent across trials.
We can thus update $Q_w(W)$ by evaluating the posterior covariance, $\bar{\Sigma}_w \in \mathbb{S}^{pT_{\text{ind}} \times pT_{\text{ind}}}$, and mean, $\bar{\boldsymbol{\mu}}_{w_n} \in \mathbb{R}^{pT_{\text{ind}}}$, of $\bar{\mathbf{w}}_n$ for each trial $n$:
\begin{align}
    \bar{\Sigma}_w &= \left((\bar{K}^w)^{-1} + (\bar{K}^w)^{-1} \left[\sum_{m=1}^M \sum_{t=1}^T \bar{K}^{wx}_{m,t} \langle (C^m)^{\top} \Phi^m C^m \rangle \bar{K}^{xw}_{m,t} \right] (\bar{K}^w)^{-1} \right)^{-1} \label{eq:qw_cov_app}\\
    \bar{\boldsymbol{\mu}}_{w_n} &= \bar{\Sigma}_{w} (\bar{K}^w)^{-1} \left[ \sum_{m=1}^M \sum_{t=1}^T \bar{K}^{wx}_{m,t} \langle C^m \rangle^{\top} \langle \Phi^m \rangle \bigl( \mathbf{y}^m_{n,t} - \langle \mathbf{d}^m \rangle \bigr) \right] \label{eq:qw_mean_app}
\end{align}
where $\bar{K}^{wx}_{m,t} = (\bar{K}^{xw}_{m,t})^{\top}$.
Note that (1) The update for the posterior covariance, $\bar{\Sigma}_w$, is identical for trials of the same length. This computation can therefore be reused efficiently across trials.
(2) Under the posterior distribution, the inducing variables for latents $j = 1,\ldots,p$ are no longer independent, as they are under the prior distribution (equations \ref{eq:wprior}, \ref{eq:mdlag-induce_completeLL}).

For multiple groups ($M > 1$) and a sufficiently small number of inducing points $T_{\text{ind}}$, evaluation of equation~\ref{eq:qw_cov_app} is significantly more efficient than the evaluation of the analogous equation~\ref{eq:qx_cov_time} for mDLAG-time.
The quantity $\bar{K}^{wx}_{m,t} \langle (C^m)^{\top} \Phi^m C^m \rangle \bar{K}^{xw}_{m,t}$ costs $\mathcal{O}(p^3 T_{\text{ind}}^2)$ operations, for a total cost of $\mathcal{O}(p^3 M T T_{\text{ind}}^2)$ when evaluated for all groups and time points.
The inversion of the block diagonal $\bar{K}^w$ costs $O(p T_{\text{ind}}^3)$ operations.
Then the final inversion of a $pT_{\text{ind}} \times pT_{\text{ind}}$ matrix costs $\mathcal{O}(p^3 T_{\text{ind}}^3)$ operations and $\mathcal{O}(p^2 T_{\text{ind}}^2)$ storage.
Thus in total, posterior inference over the inducing variables scales linearly in both the number of time points per trial $T$ and the number of groups $M$, albeit superlinearly in the number of inducing points $T_{\text{ind}}$.

The posterior mean and covariance of the latents, $X$, are computed from their corresponding inducing variables using the conditional prior distribution $P(X|W,\Omega)$ (equation~\ref{eq:xprior_induce}).
The posterior mean of latent $j$ across groups $m = 1,\ldots,M$, \\ $\mathbf{x}_{n,j,:} = [\mathbf{x}^{1\top}_{n,j,:} \cdots \mathbf{x}^{M\top}_{n,j,:}]^{\top} \in \mathbb{R}^{MT}$ is given by
\begin{align}
    \langle \mathbf{x}_{n,j,:} \rangle &= \mathbb{E}_{Q_{w\theta}}\bigl[ \mathbb{E}_{P_{x|w}}\bigl[ \mathbf{x}_{n,j,:} \bigr] \bigr] \\
    &= \mathbb{E}_{Q_{w\theta}}\bigl[ K^{xw}_j (K^w_j)^{-1} \mathbf{w}_{n,j,:} \bigr] \\
    &= K^{xw}_j (K^w_j)^{-1} \langle \mathbf{w}_{n,j,:} \rangle \label{eq:qx_mean_induce_app}
\end{align}
This computation requires $\mathcal{O}(T_{\text{ind}}^3)$ operations for the inversion of $K^w_j$, followed by $\mathcal{O}(M T T_{\text{ind}} + T_{\text{ind}}^2)$ operations for the remaining matrix multiplications.

Let us next construct $\hat{\mathbf{x}}_n = [\mathbf{x}_{n,1,:}^{\top} \cdots \mathbf{x}_{n,p,:}^{\top}]^{\top} \in \mathbb{R}^{pMT}$ by vertically concatenating latents $j = 1,\ldots,p$ on trial $n$.
Also collect the covariance matrices between each latent $j = 1,\ldots,p$ and its inducing variable, $K^{xw}_j \in \mathbb{R}^{MT \times T_{\text{ind}}}$ (equation~\ref{eq:xprior_induce}; \fig{method_intro}{d}, bottom), into the block diagonal matrix $\bar{K}^{xw} = \text{diag}(K^{xw}_1, \ldots, K^{xw}_p) \in \mathbb{R}^{pMT \times pT_{\text{ind}}}$.
Similarly, collect the latent covariance matrices, $K_j \in \mathbb{S}^{MT \times MT}$ (equation~\ref{eq:Kj}; \fig{method_intro}{b}), into the block diagonal matrix $\hat{K} = \text{diag}(K_1,\ldots,K_p) \in \mathbb{S}^{pMT \times pMT}$ ($\hat{K}$ is simply a reorganization of $\bar{K}$ in equation~\ref{eq:qx_cov_time}).
Then the full posterior covariance of the latents, $\hat{\Sigma}_x \in \mathbb{S}^{pMT \times pMT}$ is given by, 
\begin{align}
    \hat{\Sigma}_x &= \mathbb{E}_{Q_{w\theta}}\bigl[ \mathbb{E}_{P_{x|w}}\bigl[ \hat{\mathbf{x}}_n \hat{\mathbf{x}}_n^{\top} \bigr] \bigr] - \mathbb{E}_{Q_{w\theta}}\bigl[ \mathbb{E}_{P_{x|w}}\bigl[ \hat{\mathbf{x}}_n \bigr] \bigr] \mathbb{E}_{Q_{w\theta}}\bigl[ \mathbb{E}_{P_{x|w}}\bigl[ \hat{\mathbf{x}}_n \bigr] \bigr]^{\top} \\
    &= \langle \hat{\mathbf{x}}_n \hat{\mathbf{x}}_n^{\top} \rangle - \langle \hat{\mathbf{x}}_n \rangle \langle \hat{\mathbf{x}}_n \rangle^{\top} \\
    &= \hat{K} - \bar{K}^{xw} (\bar{K}^w)^{-1} \bar{K}^{wx} + \bar{K}^{xw} (\bar{K}^w)^{-1} \langle \bar{\mathbf{w}}_n \bar{\mathbf{w}}_n^{\top} \rangle (\bar{K}^w)^{-1} \bar{K}^{wx} \notag\\
    &\quad - \bar{K}^{xw} (\bar{K}^w)^{-1} \langle \bar{\mathbf{w}}_n \rangle \langle \bar{\mathbf{w}}_n \rangle^{\top} (\bar{K}^w)^{-1} \bar{K}^{wx} \\
    &= \hat{K} - \bar{K}^{xw} (\bar{K}^w)^{-1} \bar{K}^{wx} + \bar{K}^{xw} (\bar{K}^w)^{-1} \bar{\Sigma}_w (\bar{K}^w)^{-1} \bar{K}^{wx} \label{eq:qx_cov_induce_app}
\end{align}
where the inducing variable covariance matrix $\bar{\Sigma}_w$ is defined in equation~\ref{eq:qw_cov_app}.

The prior covariance matrix $\hat{K}$, the posterior covariance matrix $\hat{\Sigma}_x$, or the posterior second moment $\langle \hat{\mathbf{x}}_n \hat{\mathbf{x}}_n^{\top} \rangle$ never need to be computed in full during the mDLAG-inducing fitting procedure.
Only subsets of these quantities are needed, thereby maintaining linear scaling in $M$ and $T$ and runtime gains over mDLAG-time.
From here, the updates for $Q_d(\mathbf{d})$, $Q_{\phi}(\boldsymbol{\phi})$, $Q_c(C)$, and $Q_{\mathcal{A}}(\mathcal{A})$ look identical in form to the analogous updates in mDLAG-time \citep{gokcen_uncovering_2023}.
We include them here for completeness.

Posterior estimates of the mean parameters $\mathbf{d}$ are independent across groups (and, in fact, units).
We can thus update $Q_d(\mathbf{d})$ by evaluating the posterior covariance, $\Sigma_d^m \in \mathbb{S}^{q_m \times q_m}$, and mean, $\boldsymbol{\mu}_d^m \in \mathbb{R}^{q_m}$, of mean parameter $\mathbf{d}^m$ for each group $m$:
\begin{align}
    \Sigma_d^m &= \bigl(\beta I_{q_m} + NT \langle \Phi^{m} \rangle \bigr)^{-1} \\
    \boldsymbol{\mu}_d^m &= \Sigma_d^m \langle \Phi^{m} \rangle \sum_{n=1}^N \sum_{t=1}^T \bigl(\mathbf{y}_{n,t}^m - \langle C^m \rangle \langle \mathbf{x}_{n,t}^m \rangle \bigr)
\end{align}

Posterior estimates of precision parameters $\boldsymbol{\phi}$ are independent across groups and units.
We can thus update $Q_{\phi}(\boldsymbol{\phi})$ by evaluating the posterior parameters $\bar{a}_{\phi}$ and $\bar{b}_{\phi,r}^m$ of parameter $\phi^m_r$ for each unit $r$ in group $m$:
\begin{align}
    \bar{a}_{\phi} &= a_{\phi} + \frac{NT}{2} \label{eq:phi-induce_update1} \\
    \bar{b}_{\phi,r}^m &= b_{\phi} + \frac{1}{2} \sum_{n=1}^N \sum_{t=1}^T \biggl[ (y_{n,r,t}^m)^2 + \langle 
    (d_r^m)^2 \rangle + \text{tr}\left(\langle \bar{\mathbf{c}}_r^m (\bar{\mathbf{c}}_r^m)^{\top} \rangle \langle \mathbf{x}_{n,t}^m (\mathbf{x}_{n,t}^m)^{\top} \rangle\right) \notag\\
    &\quad - 2 \langle\bar{\mathbf{c}}_r^m \rangle^{\top} \langle \mathbf{x}_{n,t}^m \rangle \bigl(y_{n,r,t}^m - \langle d_r^m \rangle \bigr) - 2 y_{n,r,t}^m \langle d_r^m \rangle \biggr] \label{eq:phi-induce_update2}
\end{align}
Here $\bar{\mathbf{c}}_r^m \in \mathbb{R}^p$ is again the $r$\textsuperscript{th} row of $C^m$, the loading matrix for group $m$.

Posterior estimates of loading matrices $C$ are independent across groups and units, i.e., across the rows of each $C^m$.
We can thus update $Q_c(C)$ by evaluating the posterior covariance, $\Sigma_{c_r}^m \in \mathbb{S}^{p \times p}$, and mean, $\bar{\boldsymbol{\mu}}_{c_r}^m \in \mathbb{R}^{p}$, of the $r$\textsuperscript{th} row of $C^m$:
\begin{align}
    \Sigma_{c_r}^m &= \biggl(\langle \mathcal{A}^m \rangle + \langle \phi_r^m \rangle \sum_{n=1}^N \sum_{t=1}^T \langle \mathbf{x}_{n,t}^m (\mathbf{x}_{n,t}^m)^{\top}\rangle \biggr)^{-1} \\
    \bar{\boldsymbol{\mu}}_{c_r}^m &= \Sigma_{c_r}^m \langle \phi_r^m \rangle \sum_{n=1}^N \sum_{t=1}^T \langle \mathbf{x}_{n,t}^m \rangle \bigl(y_{n,r,t}^m - \langle d_r^m \rangle \bigr) \label{eq:Cmean}
\end{align}
Here $\mathcal{A}^m = \text{diag}(\alpha^m_1, \ldots, \alpha^m_p)$.
Note that the posterior independence over the rows of each $C^m$ contrasts with the prior independence over the columns of each $C^m$ (equations \ref{eq:Cprior}, \ref{eq:mdlag-induce_completeLL}).

Finally, posterior estimates of ARD parameters $\mathcal{A}$ are independent across groups and latents.
We can thus update $Q_{\mathcal{A}}(\mathcal{A})$ by evaluating the posterior parameters $\bar{a}_{\alpha}^m$ and $\bar{b}_{\alpha,j}^m$ of parameter $\alpha^m_j$ for each group $m$ and latent $j$:
\begin{align}
    \bar{a}_{\alpha}^m &= a_{\alpha} + \frac{q_m}{2} \label{eq:alpha-induce_update1} \\
    \bar{b}_{\alpha,j}^m &= b_{\alpha} + \frac{1}{2} \langle 
\lVert \mathbf{c}_j^m \rVert_2^2 \rangle \label{eq:alpha-induce_update2}
\end{align}
where $\mathbf{c}_j^m \in \mathbb{R}^{q_m}$ is the $j$\textsuperscript{th} column of the loading matrix $C^m$.
All moments $\langle \cdot \rangle$ can be readily computed from the approximate posterior distributions given in equations \ref{eq:qw_app}--\ref{eq:qalpha_induce}.

\subsubsection{Gaussian Process Parameter Updates}
There are no closed-form solutions for the Gaussian process parameter updates, but we can compute gradients and perform gradient ascent.
To optimize timescales and time delays, we rewrite the lower bound to show the terms that depend on the covariance matrices that define the latents and their inducing variables, $K_j$, $K^{xw}_j$, and $K^w_j$ for latents $j = 1,\ldots,p$.
First we define several intermediate variables.
Construct $\hat{\mathbf{y}}^m_n = [\mathbf{y}^{m\top}_{n,1,:} \cdots \mathbf{y}^{m\top}_{n,q_m,:}]^{\top} \in \mathbb{R}^{q_m T}$ by vertically concatenating the observed activity of each unit $r = 1,\ldots,q_m$ in group $m$ across time on trial $n$, $\mathbf{y}^{m\top}_{n,r,:} \in \mathbb{R}^T$.
Then construct $\hat{\mathbf{y}}_n = [\hat{\mathbf{y}}^{1\top}_n \cdots \hat{\mathbf{y}}^{M\top}_n]^{\top} \in \mathbb{R}^{qT}$, $q = \sum_m q_m$, by vertically concatenating the $\hat{\mathbf{y}}^m_n$ across groups $m = 1,\ldots,M$.
Vertically concatenate all mean parameters $\mathbf{d}^m \in \mathbb{R}^{q_m}$ across groups $m = 1,\ldots,M$ to give the vector $\mathbf{d} = [\mathbf{d}^{1\top} \cdots \mathbf{d}^{M\top}]^{\top} \in \mathbb{R}^q$.
And finally, let $\hat{C}_j = \text{diag}(\mathbf{c}^1_j,\ldots,\mathbf{c}^M_j) \in \mathbb{R}^{q \times M}$ be the block diagonal matrix where each block is the $j$\textsuperscript{th} column of the loading matrix to group $m$, $\mathbf{c}^m_j \in \mathbb{R}^q$.

Then the terms of the lower bound $L(Q,\Omega)$ that depend on the $j$\textsuperscript{th} latent are given by
\begin{align}
    L_j &= \sum_{n=1}^N \biggl[ \frac{1}{2} \log |(K^w_j)^{-1}| - \frac{1}{2} \text{tr}\left((K^w_j)^{-1} \langle \mathbf{w}_{n,j,:} \mathbf{w}_{n,j,:}^{\top} \rangle \right) \notag\\
    &\quad + \langle \mathbf{w}_{n,j,:} \rangle^{\top} (K^w_j)^{-1} K^{wx}_j \bigl( \langle \hat{C}_j \rangle^{\top} \langle \Phi \rangle \otimes I_T \bigr) \bigl( \hat{\mathbf{y}}_n - \langle \mathbf{d} \rangle \otimes \mathbf{1}_T \bigr) \notag\\
    &\quad - \frac{1}{2} \text{tr} \biggl( \bigl( \langle \hat{C}_j^{\top} \Phi \hat{C}_j \rangle \otimes I_T \bigr) \bigl(  K_j - K^{xw}_j (K^w_j)^{-1} K^{wx}_j \bigr) \biggr) \notag\\
    &\quad - \frac{1}{2} \text{tr} \biggl( \bigl( \langle \hat{C}_j^{\top} \Phi \hat{C}_j \rangle \otimes I_T \bigr) K^{xw}_j (K^w_j)^{-1} \langle \mathbf{w}_{n,j,:} \mathbf{w}_{n,j,:}^{\top} \rangle (K^w_j)^{-1} K^{wx}_j \biggr) \notag\\
    &\quad - \sum_{k \neq j} \text{tr} \biggl( \bigl( \langle \hat{C}_k^{\top} \Phi \hat{C}_j \rangle \otimes I_T \bigr) K^{xw}_j (K^w_j)^{-1} \langle \mathbf{w}_{n,j,:} \mathbf{w}_{n,k,:}^{\top} \rangle (K^w_k)^{-1} K^{wx}_k \biggr) \biggr]
\end{align}
where $\otimes$ is the Kronecker product, $I_T \in \mathbb{S}^{T \times T}$ is the $T \times T$ identity matrix, and $\mathbf{1}_T \in \mathbb{R}^T$ is the length-$T$ column vector of ones.
The overall lower bound can then be re-expressed as
\begin{equation}
    L(Q,\Omega) = \sum_{j=1}^p L_j + \text{const.}
\end{equation}

From here we omit explicit expressions for the gradients of the lower bound $L(Q,\Omega)$ with respect to each of the timescale and time delay parameters.
However, we do write out intermediate expressions for the gradients starting from the chain rule, to illustrate the dependencies of each gradient update on the covariance matrices $K_j$, $K^{xw}_j$, and $K^w_j$.
The gradients with respect to timescale $\tau_j$ for latent $j$ are given by
\begin{align}
    \frac{\partial L}{\partial \tau_j} &= \text{tr}\left( \left( \frac{\partial L_j}{\partial K_j} \right)^{\top} \left( \frac{\partial K_j}{\partial \tau_j}\right)\right) \notag\\
    &\quad + \text{tr}\left( \left( \frac{\partial L_j}{\partial K^{xw}_j} \right)^{\top} \left( \frac{\partial K^{xw}_j}{\partial \tau_j}\right)\right) + \text{tr}\left( \left( \frac{\partial L_j}{\partial K^{wx}_j} \right)^{\top} \left( \frac{\partial K^{wx}_j}{\partial \tau_j}\right)\right) \notag\\
    &\quad + \text{tr}\left( \left( \frac{\partial L_j}{\partial K^w_j} \right)^{\top} \left( \frac{\partial K^w_j}{\partial \tau_j}\right)\right)
\end{align}
However, because of special structure in the partial derivatives $\frac{\partial L_j}{\partial K_j}$ (diagonal) and $\frac{\partial K_j}{\partial \tau_j}$ (zero along the diagonal), the first trace term evaluates exactly to zero.
Furthermore, because $K^{wx}_j = (K^{xw}_j)^{\top}$, we can simplify:
\begin{align}
    \frac{\partial L}{\partial \tau_j} &= 2 \cdot \text{tr}\left( \left( \frac{\partial L_j}{\partial K^{xw}_j} \right)^{\top} \left( \frac{\partial K^{xw}_j}{\partial \tau_j}\right)\right) + \text{tr}\left( \left( \frac{\partial L_j}{\partial K^w_j} \right)^{\top} \left( \frac{\partial K^w_j}{\partial \tau_j}\right)\right)
\end{align}
leaving dependence only on the smaller covariance matrices $K^{xw}_j$ and $K^w_j$, and not on the large covariance matrix $K_j$.
The gradients with respect to time delay $D^m_j$ for latent $j$ and group $m$ can be similarly simplified:
\begin{align}
    \frac{\partial L}{\partial D^m_j} &= 2 \cdot \text{tr}\left( \left( \frac{\partial L_j}{\partial K^{xw}_j} \right)^{\top} \left( \frac{\partial K^{xw}_j}{\partial D^m_j}\right)\right)
\end{align}
Since $K^w_j$ does not involve time delays (equation~\ref{eq:delta_kw}), the partial derivative $\frac{\partial L}{\partial D^m_j}$ does not depend on $K^w_j$.

\subsection{Evaluation of the Lower Bound}
\label{app:mdlag-induce_lb}

To monitor the progress of the fitting procedure, we evaluate the variational lower bound on each iteration:
\begin{align}
     L(Q,\Omega) &= \mathbb{E}_{Q_{w\theta}}\bigl[ \log G(Y,W,\theta|\Omega) \bigr] - \text{KL}\bigl(Q_{w\theta}(W,\theta) \Vert P(W,\theta|\tau) \bigr)
\end{align}
Due to the factorized forms of $Q_{w\theta}(W,\theta)$ and $P(W,\theta|\tau)$, $L(Q,\Omega)$ becomes
\begin{align}
    L(Q,\Omega) &= \mathbb{E}_{Q_{w\theta}}\bigl[ \log G(Y,W,\theta|\Omega) \bigr] - \text{KL}(Q_w(W) \Vert P(W | \tau)) - \text{KL}(Q_c(C) \Vert P(C | \mathcal{A})) \notag\\
    &\quad - \text{KL}(Q_{\mathcal{A}}(\mathcal{A}) \Vert P(\mathcal{A}))
    - \text{KL}(Q_{\phi}(\boldsymbol{\phi}) \Vert P(\boldsymbol{\phi})) - \text{KL}(Q_d(\mathbf{d}) \Vert P(\mathbf{d})) \label{eq:lb_induce_expanded}
\end{align}
This form of the lower bound provides insight into the nature of the optimization procedure for fitting mDLAG-inducing models.
The first term is essentially an expected log-likelihood (with respect to the approximate posterior $Q_{w\theta}(W,\theta)$) of the observations $Y$, given the latest estimates of the inducing variables $W$ and model parameters $\theta$ and $\Omega$.
This term encourages mDLAG-inducing models to explain the observed activity as well as possible.
The KL-divergence terms, on the other hand, penalize deviations of each factor of the fitted posterior from its corresponding prior distribution, and hence act as a form of regularization.

Using the posterior updates in Section~\ref{sec:mdlag-induce_updates} and the prior definitions in Section~\ref{sec:mdlag-induce}, each term of the lower bound can be computed as follows:
\begin{align}
    \mathbb{E}_{Q_{w\theta}}\bigl[ \log G(Y,W,\theta|\Omega) \bigr] &= -\frac{qNT}{2}\log(2\pi) + \frac{NT}{2} \sum_{m=1}^M \sum_{r=1}^{q_m} \langle \log \phi_r^m \rangle \notag \\
    &\quad- \sum_{m=1}^M \sum_{r=1}^{q_m} (\bar{a}_{\phi} - \langle \phi_r^m \rangle b_{\phi}) \\
    -\text{KL}(Q_w(W) \Vert P(W|\tau)) &= \frac{pNT_{\text{ind}}}{2} + \frac{1}{2} \sum_{n=1}^N \biggl[ \log|\bar{\Sigma}_{w}| \notag\\
    &\quad - \sum_{j=1}^p \bigl[ \log|K^w_j| + \text{tr}((K^w_j)^{-1} \langle \mathbf{w}_{n,j,:} \mathbf{w}_{n,j,:}^{\top} \rangle) \bigr] \biggr] \\
    -\text{KL}(Q_c(C) \Vert P(C | \mathcal{A})) &= \frac{qp}{2} + \sum_{m=1}^M \biggl[ \frac{1}{2} \sum_{r=1}^{q_m} \log|\Sigma_{c_r}^m| \notag\\ 
    &\quad + \sum_{j=1}^p \biggl[ \frac{q_m}{2} \langle \log \alpha_j^m \rangle - \frac{1}{2} \langle \alpha_j^m \rangle \langle \lVert \mathbf{c}^m_j \rVert_2^2 \rangle \biggr] \biggr] \\
    -\text{KL}(Q_{\mathcal{A}}(\mathcal{A}) \Vert P(\mathcal{A})) &= \sum_{m=1}^M \sum_{j=1}^{p} \biggl[ -\bar{a}_{\alpha}^m \log \bar{b}_{\alpha,j}^m + a_{\alpha} \log b_{\alpha} + \log \frac{\Gamma(\bar{a}_{\alpha}^m )}{\Gamma(a_{\alpha})} \notag \\
    &\ \ - b_{\alpha} \langle \alpha_j^m \rangle + \bar{a}_{\alpha}^m + (a_{\alpha} - \bar{a}_{\alpha}^m)(\Psi(\bar{a}_{\alpha}^m) - \log \bar{b}_{\alpha,j}^m) \biggr] \\
    -\text{KL}(Q_{\phi}(\boldsymbol{\phi}) \Vert P(\boldsymbol{\phi})) &= \sum_{m=1}^M \sum_{r=1}^{q_m} \biggl[ -\bar{a}_{\phi} \log \bar{b}_{\phi,r}^m + a_{\phi} \log b_{\phi} + \log \frac{\Gamma(\bar{a}_{\phi})}{\Gamma(a_{\phi})} \notag\\
    &\quad - b_{\phi} \langle \phi_r^m \rangle + \bar{a}_{\phi} + (a_{\phi} - \bar{a}_{\phi})(\Psi(\bar{a}_{\phi}) - \log \bar{b}_{\phi,r}^m) \biggr] \\
    -\text{KL}(Q_d(\mathbf{d}) \Vert P(\mathbf{d})) &= \frac{q}{2} + \frac{q}{2} \log \beta + \frac{1}{2} \log|\Sigma_d| - \frac{1}{2} \beta \langle \lVert \mathbf{d} \rVert_2^2 \rangle
\end{align}
Here, $\Gamma(\cdot)$ is the gamma function, and $\Psi(\cdot)$ is the digamma function.
All moments $\langle \cdot \rangle$ can be readily computed from the approximate posterior distributions given in equations \ref{eq:qw_app}--\ref{eq:qalpha_induce}.

\clearpage

\section{Posterior Inference and Fitting via mDLAG-frequency}
\label{app:mdlag-freq_fit}

\subsection{Variational Inference}
\label{app:mdlag-freq_updates}

Let $\widetilde{Y}$ and $\widetilde{X}$ be collections of all observed neural activity and latent variables, respectively, across all frequencies and trials.
Similarly, let $\mathbf{d}$, $\boldsymbol{\phi}$, $C$, $\mathcal{A}$, $\tau$, and $D$ be collections of the mean parameters, noise precisions, loading matrices, ARD parameters, GP timescales, and time delays, respectively.
From frequency domain observations, we seek to estimate posterior distributions over the probabilistic model components
\begin{equation}
    \theta = \left\{\widetilde{X}, \ \mathbf{d}, \ \boldsymbol{\phi},\ C, \ \mathcal{A}\right\} \label{eq:mdlag-freq_theta}
\end{equation}
and point estimates of the deterministic GP parameters $\Omega = \left\{\tau, D \right\}$.

In the case of methods like GPFA \citep{yu_gaussian-process_2009} and DLAG \citep{gokcen_disentangling_2022}, the linear-Gaussian structure of the model enables an exact expectation-maximization (EM) algorithm.
With the introduction of prior distributions over model parameters (to enable automatic relevance determination), the family of mDLAG models, including mDLAG-frequency, loses this property.
The complete quasi-likelihood of the mDLAG-frequency model (\fig{method_intro}{e}),
\begin{align}
    P(\widetilde{Y},\theta | \Omega) &= P(\mathbf{d}) P(\boldsymbol{\phi}) P(C | \mathcal{A}) P(\mathcal{A}) P(\widetilde{Y} | \widetilde{X}, C, \mathbf{d}, \boldsymbol{\phi}, D) P(\widetilde{X}|\tau) \notag\\
    &= \prod_{m=1}^M \Biggl[ P(\mathbf{d}^m) \Biggl[ \prod_{r=1}^{q_m} P(\phi^m_r) \Biggr] \Biggl[ \prod_{j=1}^{p} P(\mathbf{c}_j^m \ | \ \alpha_j^m) P(\alpha_j^m) \Biggr] \Biggr] \notag\\ 
    &\cdot \prod_{n=1}^{N} \prod_{l=1}^{T} \Biggl[ \Biggl[ \prod_{m=1}^M P(\widetilde{\mathbf{y}}^m_{n,l} \ | \ \widetilde{\mathbf{x}}_{n,l}, C^m, \mathbf{d}^m, \boldsymbol{\phi}^m, \{ D^m_j \}_{j=1}^{p}) \Biggr] \Biggl[ \prod_{j=1}^{p} P(x_{n,j,l} \ | \ \tau_j) \Biggr] \Biggr] \label{eq:mdlag-freq_completeLL}
\end{align}
is no longer Gaussian.
Then a hypothetical EM E-step (evaluation of the posterior distribution $P(\theta|\widetilde{Y},\Omega)$) becomes prohibitive, as it relies on the analytically intractable marginalization of equation \ref{eq:mdlag-freq_completeLL} with respect to $\theta$.

We therefore employ instead a variational inference scheme \citep{bishop_variational_1999,klami_group_2015,gokcen_uncovering_2023}, in which we maximize the lower bound $\widetilde{L}(\widetilde{Q},\Omega)$ to the log quasi-likelihood $\log P(\widetilde{Y})$, with respect to the approximate posterior distribution $\widetilde{Q}(\theta)$ and the deterministic parameters $\Omega$:
\begin{equation}
    \log P(\widetilde{Y}) \geq \widetilde{L}(\widetilde{Q},\Omega) = \mathbb{E}_{\widetilde{Q}}[\log P(\widetilde{Y},\theta | \Omega)] - \mathbb{E}_{\widetilde{Q}}[\log \widetilde{Q}(\theta)] \label{eq:lb_freq_app}
\end{equation}
The quasi-likelihood $P(\widetilde{Y})$ is meant to approximate the time domain likelihood $P(Y)$ (equation~\ref{eq:lb_time}).
The two quantities converge as the number of time points per trial, $T$, becomes large \citep{whittle_hypothesis_1951}.

We constrain $\widetilde{Q}(\theta)$ so that it factorizes over the elements of $\theta$:
\begin{equation}
    \widetilde{Q}(\theta) = Q_{\widetilde{x}}(\widetilde{X}) Q_d(\mathbf{d}) Q_{\phi}(\boldsymbol{\phi}) Q_c(C) Q_{\mathcal{A}}(\mathcal{A}) \label{eq:q_freq_app}
\end{equation}
This factorization enables closed-form updates during optimization (see below).
The lower bound $\widetilde{L}(\widetilde{Q},\Omega)$ can then be iteratively maximized via coordinate ascent of the factors of $\widetilde{Q}(\theta)$ and the deterministic parameters $\Omega$: each factor or deterministic parameter is updated in turn while the remaining factors or parameters are held fixed.
These updates are repeated until the lower bound, which is guaranteed to be non-decreasing, improves from one iteration to the next by less than a present tolerance (here we used $10^{-8}$).

\subsubsection{Posterior Distribution Updates}
\label{sec:mdlag-freq_updates}

Maximizing the lower bound, $\widetilde{L}(\widetilde{Q},\Omega)$, with respect to the $k$\textsuperscript{th} factor of $\widetilde{Q}$, $\hat{Q}_k$, results in the following update rule \citep{bishop_variational_1999}:
\begin{equation}
    \log \hat{Q}_k(\theta_k) = \langle \log P(\widetilde{Y},\theta | \Omega) \rangle_{-k} + \text{const.} \label{eq:q-freq_update}
\end{equation}
Here we introduce the notation $\langle \cdot \rangle$ to indicate the expectation with respect to the approximate posterior distribution, $\mathbb{E}_{\widetilde{Q}}[\cdot]$, and $\langle \log P(\widetilde{Y},\theta | \Omega) \rangle_{-k}$ specifically indicates the expectation of the complete log likelihood with respect to all but the $k$\textsuperscript{th} factor of $\widetilde{Q}$.

Because of the choice of Gaussian and conjugate Gamma priors in Section~\ref{sec:mdlag-freq}, evaluation of equation \ref{eq:q-freq_update} for each factor $\hat{Q}_k$ leads to factors with the same functional form as their corresponding priors (equations~\ref{eq:xprior_freq}, \ref{eq:dprior}--\ref{eq:alphaprior}):
\begin{align}
    Q_{\widetilde{x}}(\widetilde{X}) &= \prod_{n=1}^N \prod_{l=1}^T \mathcal{N}(\widetilde{\mathbf{x}}_{n,l} \ | \ \widetilde{\boldsymbol{\mu}}_{x_{n,l}}, \widetilde{\Sigma}_{x,l}) \label{eq:qx_freq_app} \\
    Q_{d}(\mathbf{d}) &= \prod_{m=1}^M \mathcal{N}(\mathbf{d}^m \ | \ \boldsymbol{\mu}_d^m, \Sigma_d^m) \label{eq:qd_freq_app} \\
    Q_{\phi}(\boldsymbol{\phi}) &= \prod_{m=1}^M \prod_{r=1}^{q_m} \Gamma(\phi_r^m \ | \ \bar{a}_{\phi}, \bar{b}_{\phi,r}^m) \label{eq:qphi_freq_app} \\
    Q_c(C) &= \prod_{m=1}^M \prod_{r=1}^{q_m} \mathcal{N}(\bar{\mathbf{c}}_r^m \ | \ \bar{\boldsymbol{\mu}}_{c_r}^m, \Sigma_{c_r}^m) \label{eq:qc_freq_app} \\
    Q_{\mathcal{A}}(\mathcal{A}) &= \prod_{m=1}^{M} \prod_{j=1}^{p} \Gamma(\alpha_j^m \ | \ \bar{a}_{\alpha}^m, \bar{b}_{\alpha,j}^m ) \label{eq:qalpha_freq_app}
\end{align}
Here $\bar{\mathbf{c}}_r^m \in \mathbb{R}^p$ is the $r$\textsuperscript{th} row of $C^m$, the loading matrix for group $m$.
Any additional factorization in equations \ref{eq:qx_freq_app}--\ref{eq:qalpha_freq_app} emerge naturally---we impose only the factorization in equation \ref{eq:q_freq_app}.

Posterior estimates of the frequency domain latents $\widetilde{X}$ are independent across trials and frequencies.
We can thus update $Q_{\widetilde{x}}(\widetilde{X})$ by evaluating the posterior covariance, $\widetilde{\Sigma}_{x,l} \in \mathbb{C}^{p \times p}$, and mean, $\widetilde{\boldsymbol{\mu}}_{x_{n,l}} \in \mathbb{C}^{p}$, of $\widetilde{\mathbf{x}}_{n,l}$ for each trial $n$ and frequency $l$:
\begin{align}
    \widetilde{\Sigma}_{x,l} &= \biggl(S_{l}^{-1} + \sum_{m=1}^M (H^m_l)^{\mathsf{H}}\langle (C^m)^{\top} \Phi^m C^m \rangle H^m_l \biggr)^{-1} \label{eq:qx_cov_freq_app} \\
    \widetilde{\boldsymbol{\mu}}_{x_{n,l}} &= \widetilde{\Sigma}_{x,l} \sum_{m=1}^M (H^m_l)^{\mathsf{H}} \langle C^m \rangle^{\top} \langle \Phi^m \rangle \bigl( \widetilde{\mathbf{y}}^m_{n,l} - \langle \widetilde{\mathbf{d}}^m_l \rangle \bigr) \label{eq:qx_mean_freq_app}
\end{align}
Recall from equation~\ref{eq:mdlag_obs1_freq} that the diagonal matrix $H^m_l \in \mathbb{C}^{p \times p}$ is a collection of the phase shift terms $h^{m}_{j,l} = \exp(-i 2 \pi f_{l} D^{m}_j)$, across all latents $j = 1,\ldots,p$ at the single frequency $l$: $H^m_l = \text{diag}(h^{m}_{1,l}, \ldots h^{m}_{p,l})$.
The mean parameter term $\widetilde{\mathbf{d}}^m_l \in \mathbb{R}^{q_m}$ is shorthand for $\delta_{l-1} \cdot \sqrt{T} \mathbf{d}^m$, which is $\sqrt{T} \mathbf{d}^m$ for the zero frequency, and $\mathbf{0}$ otherwise.
The elements of the diagonal PSD matrix $S_l = \text{diag}(s_1(f_l), \ldots, s_p(f_l)) \in \mathbb{R}^{p \times p}$ are computed using equation~\ref{eq:s}.
Note that (1) The update for the posterior covariance, $\widetilde{\Sigma}_{x,l}$, is identical for trials of the same length.
This computation can therefore be reused efficiently across trials.
(2) Under the posterior distribution, latent variables $j = 1,\ldots,p$ are no longer independent, as they are under the prior distribution (equations \ref{eq:xprior_freq}, \ref{eq:mdlag-freq_completeLL}).

Posterior estimates of the mean parameters $\mathbf{d}$ are independent across groups (and, in fact, observed units).
We can thus update $Q_d(\mathbf{d})$ by evaluating the posterior covariance, $\Sigma_d^m \in \mathbb{S}^{q_m \times q_m}$, and mean, $\boldsymbol{\mu}_d^m \in \mathbb{R}^{q_m}$, of mean parameter $\mathbf{d}^m$ for each population $m$:
\begin{align}
    \Sigma_d^m &= \bigl(\beta I_{q_m} + NT \langle \Phi^{m} \rangle \bigr)^{-1} \\
    \boldsymbol{\mu}_d^m &= \Sigma_d^m \langle \Phi^{m} \rangle \sum_{n=1}^N 
 \sqrt{T} \bigl(\widetilde{\mathbf{y}}_{n,1}^m - \langle C^m \rangle \langle \widetilde{\mathbf{x}}_{n,1} \rangle \bigr)
\end{align}
Note that updates to the mean, $\boldsymbol{\mu}_d^m$, depend only on observed and latent variables at the zero frequency, at index $l = 1$.

Posterior estimates of precision parameters $\boldsymbol{\phi}$ are independent across groups and units.
We can thus update $Q_{\phi}(\boldsymbol{\phi})$ by evaluating the posterior parameters $\bar{a}_{\phi}$ and $\bar{b}_{\phi,r}^m$ of parameter $\phi^m_r$ for each unit $r$ in group $m$:
\begin{align}
    \widetilde{a}_{\phi} &= a_{\phi} + \frac{NT}{2} \label{eq:phi-freq_update1} \\
    \widetilde{b}_{\phi,i}^m &= b_{\phi} + \frac{1}{2} \sum_{n=1}^N \sum_{l=1}^T \Re \biggl\{ \biggl[ |\widetilde{y}_{n,r,l}^m|^2 + \delta_{l-1} T \langle 
    (d_r^m)^2 \rangle \notag\\
    &\quad + \text{tr}\left(\langle \bar{\mathbf{c}}_r^m (\bar{\mathbf{c}}_r^m)^{\top} \rangle H^m_{l} \langle \widetilde{\mathbf{x}}_{n,l} \widetilde{\mathbf{x}}_{n,l}^{\mathsf{H}} \rangle H^{m\mathsf{H}}_{l} \right) \notag\\
    &\quad - 2 \langle\bar{\mathbf{c}}_r^m \rangle^{\top} H^m_{l} \langle \widetilde{\mathbf{x}}_{n,l} \rangle \bigl(\widetilde{y}_{n,r,l}^m - \delta_{l-1} \sqrt{T} \langle d_r^m \rangle \bigr)^* - 2 \delta_{l-1} \sqrt{T} \widetilde{y}_{n,r,l}^m \langle d_r^m \rangle \biggr] \biggr\} \label{eq:phi-freq_update2}
\end{align}
Here $\bar{\mathbf{c}}_r^m \in \mathbb{R}^p$ is again the $r$\textsuperscript{th} row of $C^m$, the loading matrix for group $m$.
The term $\delta_{l-1}$ is the Kronecker delta: $1$ for the zero frequency ($l = 1$) and $0$ otherwise.
$\Re\{\cdot\}$ indicates the real part of the enclosed expression.

Posterior estimates of loading matrices $C$ are independent across groups and units, i.e., across the rows of each $C^m$.
We can thus update $Q_c(C)$ by evaluating the posterior covariance, $\Sigma_{c_r}^m \in \mathbb{S}^{p \times p}$, and mean, $\bar{\boldsymbol{\mu}}_{c_r}^m \in \mathbb{R}^{p}$, of the $r$\textsuperscript{th} row of $C^m$:
\begin{align}
    \Sigma_{c_r}^m &= \biggl(\langle \mathcal{A}^m \rangle + \langle \phi_r^m \rangle \Re \biggl\{ \sum_{n=1}^N \sum_{l=1}^T H^m_{l} \langle \widetilde{\mathbf{x}}_{n,l} \widetilde{\mathbf{x}}_{n,l}^{\mathsf{H}}\rangle H^{m\mathsf{H}}_{l} \biggr\} \biggr)^{-1} \\
    \bar{\boldsymbol{\mu}}_{c_r}^m &= \Sigma_{c_r}^m \langle \phi_r^m \rangle \Re \biggl\{ \sum_{n=1}^N \sum_{l=1}^T H^m_{l} \langle \widetilde{\mathbf{x}}_{n,l} \rangle \bigl(\widetilde{y}_{n,r,l}^m - \delta_{l-1} \sqrt{T} \langle d_r^m \rangle \bigr)^{*}\biggr\}
\end{align}
Here $\mathcal{A}^m = \text{diag}(\alpha^m_1, \ldots, \alpha^m_p)$.
Note that the posterior independence over the rows of each $C^m$ contrasts with the prior independence over the columns of each $C^m$ (equations \ref{eq:Cprior}, \ref{eq:mdlag-freq_completeLL}).

Finally, posterior estimates of ARD parameters $\mathcal{A}$ are independent across groups and latents.
We can thus update $Q_{\mathcal{A}}(\mathcal{A})$ by evaluating the posterior parameters $\bar{a}_{\alpha}^m$ and $\bar{b}_{\alpha,j}^m$ of parameter $\alpha^m_j$ for each group $m$ and latent $j$:
\begin{align}
    \bar{a}_{\alpha}^m &= a_{\alpha} + \frac{q_m}{2} \label{eq:alpha-freq_update1} \\
    \bar{b}_{\alpha,j}^m &= b_{\alpha} + \frac{1}{2} \langle 
\lVert \mathbf{c}_j^m \rVert_2^2 \rangle \label{eq:alpha-freq_update2}
\end{align}
where $\mathbf{c}_j^m \in \mathbb{R}^{q_m}$ is the $j$\textsuperscript{th} column of the loading matrix $C^m$.
All moments $\langle \cdot \rangle$ can be readily computed from the approximate posterior distributions given in equations \ref{eq:qx_freq_app}--\ref{eq:qalpha_freq_app}.

\subsubsection{Gaussian Process Parameter Updates}

There are no closed-form solutions for the Gaussian process parameter updates, but we can compute gradients and perform gradient ascent.
To optimize timescales, we rewrite the lower bound, $\widetilde{L}(\widetilde{Q},\Omega)$, to show the terms that depend on the PSD matrix $S_j$ (from equation~\ref{eq:xprior_freq}).
Let
\begin{equation}
    \widetilde{L}(\widetilde{Q},\Omega) = \sum_{j=1}^p \left[ \frac{N}{2} \log |S_j^{-1}| - \frac{1}{2} \sum_{n=1}^N \text{tr}(S_j^{-1} \langle \widetilde{\mathbf{x}}_{n,j,:} \widetilde{\mathbf{x}}_{n,j,:}^{\mathsf{H}} \rangle) \right] + \text{const.}
\end{equation}
We further make the change of variables $\gamma_j = 1/\tau_j^2$.
The variable $\gamma_j$ is simpler to work with.
We then optimize with respect to $\gamma_j$.
The $\gamma_j$ gradients for latent $j$ are then given by the chain rule:
\begin{equation}
    \frac{\partial \widetilde{L}}{\partial \gamma_j} = \text{tr}\left( \left( \frac{\partial \widetilde{L}}{\partial S_j} \right)^{\top} \left( \frac{\partial S_j}{\partial \gamma_j} \right)\right) \label{eq:grad_gamma_freq_app}
\end{equation}
where
\begin{equation}
    \frac{\partial \widetilde{L}}{\partial S_j} = -\frac{N}{2} S_j^{-1} + \frac{1}{2} S_j^{-1} \biggl( \sum_{n=1}^N \langle \widetilde{\mathbf{x}}_{n,j,:} \widetilde{\mathbf{x}}_{n,j,:}^{\mathsf{H}} \rangle \biggr) S_j^{-1} \label{eq:grad_S_app}
\end{equation}
Because $S_j$ is a diagonal matrix, so is $\partial S_j / \partial \gamma_j$.
The $l$\textsuperscript{th} element along the diagonal of $\partial S_j / \partial \gamma_j$ is then given by
\begin{equation}
    \frac{\partial s_{j}(f_{l})}{\partial \gamma_j} = \left(1 - \sigma_j^2\right) \sqrt{\frac{\pi}{2}} \exp\left(\frac{-(2 \pi f_{l})^2}{2 \gamma_j}\right) \left[ (2 \pi f_{l})^2 \gamma_j^{-5/2} - \gamma_j^{-3/2} \right]
\end{equation}
To optimize $\gamma_j$ while respecting non-negativity constraints, we perform the change of variables $\gamma_j = \exp(\hat{\gamma}_j)$, and then perform unconstrained gradient ascent with respect to $\hat{\gamma}_j$.

Na\"ive computation of the posterior second moment $\langle \widetilde{\mathbf{x}}_{n,j,:} \widetilde{\mathbf{x}}_{n,j,:}^{\mathsf{H}} \rangle$ in equation~\ref{eq:grad_S_app} would lead to a gradient evaluation (equation~\ref{eq:grad_gamma_freq_app}) that costs $\mathcal{O}(T^2)$ operations.
This cost can be improved to $\mathcal{O}(T)$ by exploiting the diagonal structure of $S_j$ and its gradient $\partial S_j / \partial \gamma_j$.
Substituting equation~\ref{eq:grad_S_app} into equation~\ref{eq:grad_gamma_freq_app}, we get
\begin{align}
    \frac{\partial \widetilde{L}}{\partial \gamma_j} &= \text{tr}\left( \left( -\frac{N}{2} S_j^{-1} + \frac{1}{2} S_j^{-1} \biggl( \sum_{n=1}^N \langle \widetilde{\mathbf{x}}_{n,j,:} \widetilde{\mathbf{x}}_{n,j,:}^{\mathsf{H}} \rangle \biggr) S_j^{-1} \right)^{\top} \left( \frac{\partial S_j}{\partial \gamma_j} \right)\right) \notag \\
    &= -\frac{N}{2} \text{tr}\left(S_j^{-1} \frac{\partial S_j}{\partial \gamma_j}\right) + \frac{1}{2} \text{tr}\left( \sum_{n=1}^N \langle \widetilde{\mathbf{x}}_{n,j,:} \widetilde{\mathbf{x}}_{n,j,:}^{\mathsf{H}} \rangle^{\top} S_j^{-1} \frac{\partial S_j}{\partial \gamma_j} S_j^{-1} \right) 
    \label{eq:grad_tau_freq1}
\end{align}
Because of the diagonal structure of $S_j$ and its gradient $\partial S_j / \partial \gamma_j$, however, both trace terms can be computed efficiently, and only the diagonal elements of $\langle \widetilde{\mathbf{x}}_{n,j,:} \widetilde{\mathbf{x}}_{n,j,:}^{\mathsf{H}} \rangle$ are needed:
\begin{align}
    \frac{\partial \widetilde{L}}{\partial \gamma_j} &= \sum_{l=1}^T \left( s^{-1}_j(f_l) \frac{\partial s_j(f_l)}{\partial \gamma_j} + \left( \sum_{n=1}^N \langle |  \widetilde{x}_{n,j,l} |^2 \rangle \right) s^{-1}_j(f_l) \frac{\partial s_j(f_l)}{\partial \gamma_j} s^{-1}_j(f_l) \right) \label{eq:grad_tau_freq2}
\end{align}
The evaluation of GP timescale gradients therefore requires only $\mathcal{O}(T)$ operations.

The diagonal structure of the phase shift matrix $H^m_l$ can be similarly exploited to produce efficient time delay parameter updates.
To optimize time delays, we rewrite the lower bound to show the terms that depend on $H^m_{l}$.
Let
\begin{align}
    \widetilde{L}^m_{n,l} = &-\frac{1}{2} \text{tr} \biggl( \langle C^{m\top} \Phi^m C^m \rangle H^m_{l} \langle \widetilde{\mathbf{x}}_{n,l} \widetilde{\mathbf{x}}_{n,l}^{\mathsf{H}} \rangle H^{m\mathsf{H}}_{l} \biggr) \notag \\
    &+ \frac{1}{2} \text{tr} \biggl( H^m_{l} \langle \widetilde{\mathbf{x}}_{n,l} \rangle (\widetilde{\mathbf{y}}^m_{n,l} - \langle \widetilde{\mathbf{d}}^m_{l} \rangle)^{\mathsf{H}} \langle \Phi^m \rangle \langle C^m \rangle \biggr) \notag \\
    &+ \frac{1}{2} \text{tr} \biggl( \langle C^m \rangle^{\top} \langle \Phi^m \rangle (\widetilde{\mathbf{y}}^m_{n,l} - \langle \widetilde{\mathbf{d}}^m_{l} \rangle) \langle \widetilde{\mathbf{x}}_{n,l} \rangle^{\mathsf{H}} H^{m\mathsf{H}}_{l} \biggr)
\end{align}
Then
\begin{align}
    \widetilde{L}(\widetilde{Q},\Omega) = \sum_{n=1}^N \sum_{l=1}^T \sum_{m=1}^M \widetilde{L}^m_{n,l} + \text{const.}
\end{align}

Since $H^m_{l}$ is generally complex-valued, we follow the Wirtinger calculus (\citealp[Appendix 2]{schreier_statistical_2010}; \citealp{wirtinger_zur_1927}), and treat the lower bound $\widetilde{L}(\widetilde{Q},\Omega)$ formally as a function of both $H^m_{l}$ and its complex conjugate $(H^m_{l})^*$.
Then delay gradients for group $m$ and latent $j$ are given by the Wirtinger chain rule, which requires partial derivatives with respect to both $H^m_{l}$ and $(H^m_{l})^*$:
\begin{align}
    \frac{\partial \widetilde{L}}{\partial D^m_j} &= \sum_{n=1}^N \sum_{l=1}^T \frac{\partial \widetilde{L}^m_{n,l}}{\partial D^m_j} \\
    &= \sum_{n=1}^N \sum_{l=1}^T \Biggl[ \text{tr}\Biggl( \biggl( \frac{\partial \widetilde{L}^m_{n,l}}{\partial H^{m}_{l}} \biggr)^{\top} \biggl( \frac{\partial H^{m}_{l}}{\partial D^m_j} \biggr)\Biggr) + \text{tr}\Biggl( \biggl( \frac{\partial \widetilde{L}^m_{n,l}}{\partial (H^{m}_{l})^*} \biggr)^{\top} \biggl( \frac{\partial (H^{m}_{l})^*}{\partial D^m_j} \biggr)\Biggr)\Biggr]
\end{align}
The partial derivatives $\frac{\partial H^{m}_{l}}{\partial D^m_j} \in \mathbb{C}^{p \times p}$ and $\frac{\partial (H^{m}_{l})^*}{\partial D^m_j} \in \mathbb{C}^{p \times p}$ are each single-entry matrices with 
\begin{equation}
\frac{\partial H^{m}_{l}}{\partial D^m_j} = \text{diag}\bigl(0, \ldots, -i 2 \pi f_{l} \cdot h^{m}_{j,l}, \ldots, 0\bigr)
\end{equation}
and 
\begin{equation}
\frac{\partial (H^{m}_{l})^*}{\partial D^m_j} = \text{diag}\bigl(0, \ldots, i 2 \pi f_{l} \cdot (h^{m}_{j,l})^*, \ldots, 0\bigr)
\end{equation}
where $h^{m}_{j,l} = \exp(-i 2 \pi f_{l} D^{m}_j)$.
Each $\frac{\partial \widetilde{L}^m_{n,l}}{\partial H^{m}_{l}} \in \mathbb{C}^{p \times p}$ and $\frac{\partial \widetilde{L}^m_{n,l}}{\partial (H^{m}_{l})^*} \in \mathbb{C}^{p \times p}$ are given by
\begin{align}
    \frac{\partial \widetilde{L}^m_{n,l}}{\partial H^{m}_{l}} &= -\frac{1}{2} \langle C^{m\top} \Phi^m C^m \rangle (H^m_{l})^* \langle \widetilde{\mathbf{x}}_{n,l} \widetilde{\mathbf{x}}_{n,l}^{\mathsf{H}} \rangle^{\top} + \frac{1}{2}  \langle C^m \rangle^{\top} \langle \Phi^m \rangle (\widetilde{\mathbf{y}}^m_{n,l} - \langle \widetilde{\mathbf{d}}^m_{l} \rangle)^* \langle \widetilde{\mathbf{x}}_{n,l} \rangle^{\top} \\
    \frac{\partial \widetilde{L}^m_{n,l}}{\partial (H^{m}_{l})^*} &= -\frac{1}{2} \langle \widetilde{\mathbf{x}}_{n,l} \widetilde{\mathbf{x}}_{n,l}^{\mathsf{H}} \rangle^{\top} H^{m}_{l} \langle C^{m\top} \Phi^m C^m \rangle + \frac{1}{2} \langle \widetilde{\mathbf{x}}_{n,l} \rangle^{*} (\widetilde{\mathbf{y}}^m_{n,l} - \langle \widetilde{\mathbf{d}}^m_{l} \rangle)^{\top} \langle \Phi^m \rangle \langle C^m \rangle
\end{align}
The overall delay gradient $\frac{\partial \widetilde{L}}{\partial D^m_j}$ then evaluates to
\begin{align}
     \frac{\partial \widetilde{L}}{\partial D^m_j} = \sum_{n=1}^N \sum_{l=1}^T \Re \biggl\{ i2\pi f_l \cdot h^m_{j,l} \cdot \biggl( &\sum_{k=1}^p \langle \widetilde{x}_{n,j,l} \widetilde{x}_{n,k,l}^* \rangle \cdot (h^m_{k,l})^* \cdot \langle \mathbf{c}^{m\top}_k \Phi^m \mathbf{c}^{m}_j \rangle \notag \\
     &- \langle \widetilde{x}_{n,j,l} \rangle (\widetilde{\mathbf{y}}^m_{n,l} - \langle \widetilde{\mathbf{d}}^m_{l} \rangle)^{\mathsf{H}} \langle \Phi^m \rangle \langle \mathbf{c}^m_j \rangle \biggr) \biggr\}
\end{align}
which costs $\mathcal{O}(T)$ operations.

In practice, we fix all delay parameters for group $1$ at $0$ to ensure identifiability.
Similar to the timescales, one might wish to constrain the delays within some physically realistic range, such as the length of an experimental trial, so that $-D_{\text{max}} \leq D^m_j \leq D_{\text{max}}$.
Toward that end, we make the change of variables $D^m_j = D_{\text{max}} \cdot \tanh(\frac{\hat{D}^m_j}{2})$ and perform unconstrained gradient ascent with respect to $\hat{D}^m_j$.
Here we chose $D_{\text{max}}$ to be half the length of a trial.

\subsection{Evaluation of the Lower Bound}
\label{app:mdlag-freq_lb}

To monitor the progress of the fitting procedure, we evaluate the lower bound, $\widetilde{L}(\widetilde{Q},\Omega)$, on each iteration.
To evaluate the lower bound, we can rewrite it as follows:
\begin{equation}
    \widetilde{L}(\widetilde{Q},\Omega) = \mathbb{E}_{\widetilde{Q}}[\log P(\widetilde{Y} | \theta, \Omega)] - \text{KL}(\widetilde{Q}(\theta) \Vert P(\theta | \Omega))
\end{equation}
$\text{KL}(\widetilde{Q}(\theta) \Vert P(\theta|\Omega))$ is the KL-divergence between the approximate posterior distribution $\widetilde{Q}(\theta)$ and prior distribution $P(\theta | \Omega)$.
Due to the factorized forms of $\widetilde{Q}(\theta)$ and $P(\theta|\Omega)$, $\widetilde{L}(\widetilde{Q},\Omega)$ becomes
\begin{align}
    \widetilde{L}(\widetilde{Q},\Omega) &= \mathbb{E}_{\widetilde{Q}}[\log P(\widetilde{Y} | \theta, \Omega)] - \text{KL}(Q_{\widetilde{x}}(\widetilde{X}) \Vert P(\widetilde{X} | \Omega)) - \text{KL}(Q_c(C) \Vert P(C | \mathcal{A})) \notag\\
    &\quad - \text{KL}(Q_{\mathcal{A}}(\mathcal{A}) \Vert P(\mathcal{A}))
    - \text{KL}(Q_{\phi}(\boldsymbol{\phi}) \Vert P(\boldsymbol{\phi})) - \text{KL}(Q_d(\mathbf{d}) \Vert P(\mathbf{d}))
\end{align}
This form of the ELBO provides insight into the nature of the optimization procedure for fitting via mDLAG-frequency.
The first term is the expected log-quasi-likelihood (with respect to the approximate posterior $\widetilde{Q}(\theta)$) of the frequency domain observations, $\widetilde{Y}$, given the latest model parameters, $\theta$ and $\Omega$.
This term encourages mDLAG-frequency models to explain the frequency domain observations as well as possible.
The KL-divergence terms, on the other hand, penalize deviations of each factor of the fitted posterior from its corresponding prior distribution, and hence act as a form of regularization.

Using the posterior updates in Section~\ref{sec:mdlag-freq_updates} and the prior definitions in Section~\ref{sec:mdlag-freq}, each term of the lower bound can be computed as follows:
\begin{align}
    \mathbb{E}_{\widetilde{Q}}[\log P(\widetilde{Y} | \theta, \Omega)] &= -\frac{qNT}{2}\log(2\pi) + \frac{NT}{2} \sum_{m=1}^M \sum_{r=1}^{q_m} \langle \log \phi_r^m \rangle \notag \\
    &\quad - \sum_{m=1}^M \sum_{r=1}^{q_m} (\bar{a}_{\phi} - \langle \phi_r^m \rangle b_{\phi}) \\
    -\text{KL}(Q_{\widetilde{x}}(\widetilde{X}) \Vert P(\widetilde{X}|\Omega)) &= \frac{pNT}{2} + \frac{N}{2} \sum_{l=1}^T \log|\widetilde{\Sigma}_{x,l}| - \frac{N}{2} \sum_{l=1}^T \sum_{j=1}^p \log(s_j(f_l)) \notag \\
    &\quad -\frac{1}{2} \sum_{l=1}^T \sum_{j=1}^p s^{-1}_j(f_l) \sum_{n=1}^N \langle |  \widetilde{x}_{n,j,l} |^2 \rangle \\
    -\text{KL}(Q_c(C) \Vert P(C | \mathcal{A})) &= \frac{qp}{2} + \sum_{m=1}^M \biggl[ \frac{1}{2} \sum_{r=1}^{q_m} \log|\Sigma_{c_r}^m| \notag\\ 
    &\quad + \sum_{j=1}^p \biggl[ \frac{q_m}{2} \langle \log \alpha_j^m \rangle - \frac{1}{2} \langle \alpha_j^m \rangle \langle \lVert \mathbf{c}^m_j \rVert_2^2 \rangle \biggr] \biggr] \\
    -\text{KL}(Q_{\mathcal{A}}(\mathcal{A}) \Vert P(\mathcal{A})) &= \sum_{m=1}^M \sum_{j=1}^{p} \biggl[ -\bar{a}_{\alpha}^m \log \bar{b}_{\alpha,j}^m + a_{\alpha} \log b_{\alpha} + \log \frac{\Gamma(\bar{a}_{\alpha}^m )}{\Gamma(a_{\alpha})} \notag \\
    &\quad - b_{\alpha} \langle \alpha_j^m \rangle + \bar{a}_{\alpha}^m + (a_{\alpha} - \bar{a}_{\alpha}^m)(\Psi(\bar{a}_{\alpha}^m) - \log \bar{b}_{\alpha,j}^m) \biggr] \\
    -\text{KL}(Q_{\phi}(\boldsymbol{\phi}) \Vert P(\boldsymbol{\phi})) &= \sum_{m=1}^M \sum_{r=1}^{q_m} \biggl[ -\bar{a}_{\phi} \log \bar{b}_{\phi,r}^m + a_{\phi} \log b_{\phi} + \log \frac{\Gamma(\bar{a}_{\phi})}{\Gamma(a_{\phi})} \notag\\
    &\quad - b_{\phi} \langle \phi_r^m \rangle + \bar{a}_{\phi} + (a_{\phi} - \bar{a}_{\phi})(\Psi(\bar{a}_{\phi}) - \log \bar{b}_{\phi,r}^m) \biggr] \\
    -\text{KL}(Q_d(\mathbf{d}) \Vert P(\mathbf{d})) &= \frac{q}{2} + \frac{q}{2} \log \beta + \frac{1}{2} \log|\Sigma_d| - \frac{1}{2} \beta \langle \lVert \mathbf{d} \rVert_2^2 \rangle
\end{align}
Here, $\Gamma(\cdot)$ is the gamma function, and $\Psi(\cdot)$ is the digamma function.
All moments $\langle \cdot \rangle$ can be readily computed from the approximate posterior distributions given in equations \ref{eq:qx_freq_app}--\ref{eq:qalpha_freq_app}.

\clearpage

\section{Leave-Group-Out Prediction}
\label{app:mdlag_pred_lgo}

Each of the three methods we considered here is designed to optimize a slightly different objective function (mDLAG-time: equation~\ref{eq:lb_time}, mDLAG-inducing: equation~\ref{eq:lb_induce}, mDLAG-frequency: equation~\ref{eq:lb_freq_app}).
We therefore developed a common performance metric to facilitate comparison across these methods: leave-group-out prediction \citep{klami_group_2015,gokcen_uncovering_2023}.
Leave-group-out prediction measures an mDLAG model's ability to capture interactions across observed groups. 
In brief, we used a model fit to training data to predict (on held-out test trials) the unobserved activity of held-out units in one group, given the observed activity of units in the remaining groups.
Our three methods each give rise to a distinct way to perform this prediction.

\subsection{Prediction via mDLAG-time}
\label{app:mdlag_pred_lgo_time}

Let us first collect observed variables (for one trial) in a manner that highlights group structure.
We collect observations in group $m$ on trial $n$ in $\check{\mathbf{y}}_n^m = [\mathbf{y}_{n,:,1}^{m\top} \cdots \mathbf{y}_{n,:,T}^{m\top} ]^{\top} \in \mathbb{R}^{q_{m}T}$ by vertically concatenating the observed activity $\mathbf{y}^m_{n,:,t} \in \mathbb{R}^{q_m}$ in group $m$ across all times $t = 1,\ldots,T$.
Then, we collect the observations for the remaining $M-1$ groups in $\check{\mathbf{y}}_n^{-m}\in \mathbb{R}^{\sum_{k\neq m}q_{k} T}$, obtained by vertically concatenating the ordered set of observations $\{ \check{\mathbf{y}}_n^{k} \}_{k\neq m}$.

Our goal is to predict $\check{\mathbf{y}}_n^m$ given $\check{\mathbf{y}}_n^{-m}$.
We do so by inferring the latents given observations $\check{\mathbf{y}}_n^{-m}$, and then predicting the held-out activity $\check{\mathbf{y}}_n^m$ from the inferred latents.
Toward that end, we similarly collect the latents for group $m$ on trial $n$ in $\check{\mathbf{x}}_n^m = [\mathbf{x}_{n,:,1}^{m\top} \cdots \mathbf{x}_{n,:,T}^{m\top} ]^{\top} \in \mathbb{R}^{pT}$ by vertically concatenating the latents $\mathbf{x}_{n,:,t}^{m} \in \mathbb{R}^p$ across all times $t = 1,\ldots,T$.
Then the latents for the remaining $M-1$ groups can be collected in $\check{\mathbf{x}}_n^{-m}\in \mathbb{R}^{(M-1) p T}$, obtained by vertically concatenating the ordered set of latents $\{ \check{\mathbf{x}}_n^{k} \}_{k\neq m}$.
This variable reorganization then allows us to rewrite the mDLAG-time state model as
\begin{equation} \label{eq:xprior_pred_time}
    \begin{bmatrix}
    \check{\mathbf{x}}^m_{n} \\
    \check{\mathbf{x}}^{-m}_{n}
    \end{bmatrix} \sim \mathcal{N}\left(\mathbf{0}, \begin{bmatrix} \check{K}_{m,m} & \check{K}_{m,-m} \\
    \check{K}_{-m,m} & \check{K}_{-m,-m}\end{bmatrix}\right)
\end{equation}
where the elements of the GP covariance matrices $\check{K}_{m,m} \in \mathbb{S}^{pT \times pT}$, $\check{K}_{m,-m} = \check{K}_{-m,m}^{\top} \in \mathbb{R}^{pT \times (M-1)pT}$, and $\check{K}_{-m,-m} \in \mathbb{S}^{(M-1)pT \times (M-1)pT}$ are computed using equations \ref{eq:k} and \ref{eq:delta_k}.

Next, for each group $m$, define $\langle\check{C}^m\rangle \in \mathbb{R}^{q_m T \times pT}$, $\langle\check{\Phi}^m\rangle \in \mathbb{S}^{q_m T \times q_m T}$, and $\langle\check{R}^m\rangle \in \mathbb{R}^{pT \times pT}$ as block diagonal matrices comprising $T$ copies of the matrices $\langle C^m \rangle$, $\langle \Phi^m \rangle$, and $\langle R^m \rangle = \langle C^{m \top} \Phi^m C^m \rangle$, respectively.
Define also $\langle\check{\mathbf{d}}^m\rangle \in \mathbb{R}^{q_m T}$ by vertically concatenating $T$ copies of $\langle \mathbf{d}^m \rangle$.
The parameters corresponding to the remaining $M-1$ populations can then be collected into the block diagonal matrices \\ $\langle \check{C}^{-m} \rangle = \text{diag}(\{ \langle \check{C}^{k} \rangle \}_{k\neq m}) \in \mathbb{R}^{\sum_{k\neq m}q_{k} T \times (M-1) p T}$, $\langle \check{\Phi}^{-m} \rangle = \text{diag}(\{ \langle \check{\Phi}^{k} \rangle \}_{k\neq m}) \in \mathbb{R}^{\sum_{k\neq m}q_{k} T \times \sum_{k\neq m}q_{k} T}$, $\langle \check{R}^{-m} \rangle = \text{diag}(\{ \langle \check{R}^{k} \rangle \}_{k\neq m}) \in \mathbb{R}^{(M-1) p T \times (M-1) p T}$, and the vector $\langle \check{\mathbf{d}}^{-m} \rangle \in \mathbb{R}^{\sum_{k\neq m}q_{k} T}$, obtained by vertically concatenating the elements of the set $\{ \langle \check{\mathbf{d}}^{k} \rangle \}_{k\neq m}$.

Similar to the update equations for the posterior distribution over latents ($Q_x(X)$, equation~\ref{eq:qx_time}), we compute the inferred latents given only the observations $\check{\mathbf{y}}_n^{-m}$ according to
\begin{align}
    \check{\Sigma}^{-m}_{x} &= \bigl(\check{K}_{-m,-m}^{-1} + \langle \check{R}^{-m}\rangle\bigr)^{-1} \\
    \check{\boldsymbol{\mu}}^{-m}_{x_n} &= \check{\Sigma}^{-m}_{x} \langle \check{C}^{-m} \rangle^{\top} \langle \check{\Phi}^{-m} \rangle \bigl( \check{\mathbf{y}}^{-m}_n - \langle \check{\mathbf{d}}^{-m} \rangle \bigr)
\end{align}
We then use equation \ref{eq:xprior_pred_time} to infer the latents for group $m$ according to
\begin{equation}
    \check{\boldsymbol{\mu}}_{x_n}^{m} = \check{K}_{m,-m} (\check{K}_{-m})^{-1} \check{\boldsymbol{\mu}}_{x_n}^{-m} \label{eq:xpred_time}
\end{equation}
and take predictions of the observations in group $m$ to be
\begin{equation}
    \smash[t]{\overset{\star}{\mathbf{y}}}^{m}_n = \langle \check{C}^m \rangle \check{\boldsymbol{\mu}}_{x_n}^{m} + \langle \check{\mathbf{d}}^m \rangle \label{eq:ypred_time}
\end{equation}
We employed this approach for models fit via mDLAG-time and for models fit via mDLAG-frequency in \fig{npx}{b} and \suppfig{npx_supp_S32}{a}.

\subsection{Prediction via mDLAG-inducing}
\label{app:mdlag_pred_lgo_induce}

Predictions via inducing variables take the same form as equation~\ref{eq:ypred_time}, above.
However, the predicted latents in group $m$, $\check{\boldsymbol{\mu}}_{x_n}^{m}$, are computed differently than in equation~\ref{eq:xpred_time}.
Similar to the updates of the posterior distribution over the inducing variables, $Q_w(W)$ (equations \ref{eq:qw_cov_app} and \ref{eq:qw_mean_app}), we compute the inferred inducing variables given observations in all but the held-out $m$\textsuperscript{th} group.

Toward that end, let us first define several variables.
First construct \\ $\bar{\mathbf{w}}_n = [\mathbf{w}^{\top}_{n,1,:} \cdots \mathbf{w}^{\top}_{n,p,:}]^{\top} \in \mathbb{R}^{pT_{\text{ind}}}$ by vertically concatenating the inducing variables $\mathbf{w}_{n,j,:}$ for latents $j = 1,\ldots,p$ on trial $n$.
Collect the inducing variable covariance matrices $K^w_j$ for $j = 1,\ldots,p$ (equation~\ref{eq:wprior}; \fig{method_intro}{d}, top) into the block diagonal matrix $\bar{K}^w = \text{diag}(K^w_1, \ldots, K^w_p) \in \mathbb{S}^{pT_{\text{ind}} \times pT_{\text{ind}}}$.
Then, let $\mathbf{k}^{xw}_{m,j,t} \in \mathbb{R}^{T_{\text{ind}}}$ be the $t$\textsuperscript{th} row of $K^{xw}_{m,j} \in \mathbb{R}^{T \times T_{\text{ind}}}$, the $m$\textsuperscript{th} block (for group $m$) of the covariance matrix between latent $j$ and its inducing variable, $K^{xw}_j$ (equation~\ref{eq:xprior_induce}; \fig{method_intro}{d}, bottom).
For time point $t$, collect each $\mathbf{k}^{xw}_{m,j,t}$ for latents $j = 1,\ldots,p$ into the block diagonal matrix $\bar{K}^{xw}_{m,t} = \text{diag}(\mathbf{k}^{xw\top}_{m,1,t},\ldots,\mathbf{k}^{xw\top}_{m,p,t}) \in \mathbb{R}^{p \times pT_{\text{ind}}}$.

Given observations in all but the $m$\textsuperscript{th} group, we can then evaluate the posterior covariance, $\bar{\Sigma}_w^{-m} \in \mathbb{S}^{pT_{\text{ind}} \times pT_{\text{ind}}}$, and mean, $\bar{\boldsymbol{\mu}}_{w_n}^{-m} \in \mathbb{R}^{pT_{\text{ind}}}$, of $\bar{\mathbf{w}}_n$ for each trial $n$:
\begin{align}
    \bar{\Sigma}_w^{-m} &= \left((\bar{K}^w)^{-1} + (\bar{K}^w)^{-1} \left[\sum_{k\neq m} \sum_{t=1}^T \bar{K}^{wx}_{k,t} \langle (C^k)^{\top} \Phi^k C^k \rangle \bar{K}^{xw}_{k,t} \right] (\bar{K}^w)^{-1} \right)^{-1} \\
    \bar{\boldsymbol{\mu}}_{w_n}^{-m} &= \bar{\Sigma}_{w} (\bar{K}^w)^{-1} \left[ \sum_{k\neq m} \sum_{t=1}^T \bar{K}^{wx}_{k,t} \langle C^k \rangle^{\top} \langle \Phi^k \rangle \bigl( \mathbf{y}^k_{n,t} - \langle \mathbf{d}^k \rangle \bigr) \right]
\end{align}
If we collect the covariance matrices $K^{xw}_{m,j}$ for group $m$ and for latents $j = 1,\ldots,p$ into the block diagonal matrix $\bar{K}^{xw}_m = \text{diag}(K^{xw}_{m,1},\ldots,K^{xw}_{m,p}) \in \mathbb{R}^{pT \times pT_{\text{ind}}}$, then we can compute the predicted latents in group $m$ based on equation~\ref{eq:qx_mean_induce_app}:
\begin{equation}
    \bar{\boldsymbol{\mu}}_{x_n}^{m} = \bar{K}^{xw}_m (\bar{K}^w)^{-1} \bar{\boldsymbol{\mu}}_{w_n}^{-m} \label{eq:xpred_induce}
\end{equation}
The vector of predicted latents in equation~\ref{eq:ypred_time}, $\check{\boldsymbol{\mu}}_{x_n}^{m}$, can be obtained by a reorganization (permutation) of the elements in $\bar{\boldsymbol{\mu}}_{x_n}^{m}$, above.
Then predicted observations can be obtained directly from equation~\ref{eq:ypred_time}.
We employed this approach for models fit via mDLAG-inducing in \fig{npx}{b} and \suppfig{npx_supp_S32}{a}.

\subsection{Prediction via mDLAG-frequency}
\label{app:mdlag_pred_lgo_freq}

As in Section~\ref{sec:mdlag-freq}, we now consider the unitary DFT of the time series of observations for unit $r$ in group $m$ on trial $n$: $\widetilde{\mathbf{y}}^m_{n,r,:} = U_T \mathbf{y}^m_{n,r,:}$, where $U_T \in \mathbb{C}^{T \times T}$ is the unitary DFT matrix.
Then let $\widetilde{\mathbf{y}}^m_{n,l} \in \mathbb{R}^{q_m}$ be the frequency domain observations of group $m$ on trial $n$ and frequency $l$, and further collect the observations for the remaining $M-1$ groups on trial $n$ and frequency $l$ in the set $\widetilde{Y}_{n,l}^{-m} = \{\widetilde{\mathbf{y}}_{n,l}^{k} \}_{k \neq m}$.
Ultimately, our goal is still to predict the time domain observations $\mathbf{y}^m_{n,r,:}$ for all units $r = 1,\ldots,q_m$ in group $m$.
We will do so, however, by using the frequency domain observations $\widetilde{Y}_{n,l}^{-m}$, and by performing inference of the latents in the frequency domain.

Similar to the updates of the posterior distribution over the frequency domain latents, $Q_{\widetilde{x}}(\widetilde{X})$ (equations \ref{eq:qx_cov_freq_app} and \ref{eq:qx_mean_freq_app}), we compute the inferred latents given the frequency domain observations $\widetilde{Y}_{n,l}^{-m}$ according to
\begin{align}
    \widetilde{\Sigma}_{x,l}^{-m} &= \biggl(S_{l}^{-1} + \sum_{k\neq m} (H_{l}^{k})^{\mathsf{H}}\langle (C^k)^{\top} \Phi^k C^k \rangle H_{l}^{k} \biggr)^{-1} \\
    \widetilde{\boldsymbol{\mu}}_{x_{n,l}}^{-m} &= \widetilde{\Sigma}_{x,l}^{-m} \sum_{k \neq m} (H_{l}^{k})^{\mathsf{H}} \langle C^k \rangle^{\top} \langle \Phi^k \rangle \bigl( \widetilde{\mathbf{y}}_{n,l}^k - \langle \widetilde{\mathbf{d}}_{l}^k \rangle \bigr)
\end{align}
Recall the diagonal phase shift matrix $H^m_l \in \mathbb{C}^{p \times p}$, defined in equation~\ref{eq:mdlag_obs1_freq}.
We then take predictions of the frequency domain observations in group $m$ to be
\begin{equation}
    \smash[t]{\overset{\star}{\widetilde{\mathbf{y}}}}^m_{n,l} = \langle C^m \rangle H_{l}^{m} \widetilde{\boldsymbol{\mu}}_{x_{n,l}}^{-m} + \langle \widetilde{\mathbf{d}}_{l}^m \rangle \label{eq:ypred_freq}
\end{equation}
To convert these predictions from the frequency domain to the time domain, we take the inverse unitary DFT of the frequency domain activity for each unit $r = 1,\ldots,q_m$ in group $m$ on each trial $n$: $\smash[t]{\overset{\star}{\mathbf{y}}}^m_{n,r,:} = U^{\mathsf{H}}_T \smash[t]{\overset{\star}{\widetilde{\mathbf{y}}}}^m_{n,r,:}$.
We employed this approach in \fig{scalingT}{a} and \fig{scalingM}{a}, see Section~\ref{app:mdlag_pred_luo} below.

\subsection{A Common Performance Metric: Leave-Group-Out \texorpdfstring{$R^2$}{R2}}
\label{app:R2lgo}

Let $\mathbf{y}^m_{n,t} \in \mathbb{R}^{q_m}$ be the observed activity for group $m$ at time $t$ on trial $n$ of a held-out test set, and let $\smash[t]{\overset{\star}{\mathbf{y}}}^m_{n,t} \in \mathbb{R}^{q_m}$ be its predicted value.
This prediction may come from any of the methods above (Sections \ref{app:mdlag_pred_lgo_time}--\ref{app:mdlag_pred_lgo_freq}).
Furthermore, let $\boldsymbol{\mu}_y^{m} \in \mathbb{R}^{q_m}$ be the sample mean for each unit in group $m$, taken over all time points and trials.

We then define a leave-group-out $R^2$ value as follows:
\begin{equation}
    R^{2}_{\text{lgo}} = 1 - \frac{\sum_{m=1}^M \sum_{n=1}^N \sum_{t=1}^T \lVert \mathbf{y}^m_{n,t} - \smash[t]{\overset{\star}{\mathbf{y}}}^m_{n,t} \rVert^2_F}{\sum_{m=1}^M \sum_{n=1}^N \sum_{t=1}^T \lVert \mathbf{y}^m_{n,t} - \boldsymbol{\mu}_y^{m} \rVert^2_F} \label{eq:r2_lgo}
\end{equation}
where $\lVert \cdot \rVert_F$ is the Frobenius norm.
The value $R^2_{lgo} \in (-\infty, 1]$, where a value of $1$ implies perfect prediction of held-out observations, and a negative value implies that estimates predict these held-out observations less accurately than simply the sample mean.
The $R^2_{\text{lgo}}$ metric is normalized by the total variance of observed activity within each dataset, thereby facilitating comparison across datasets, in which the variance of observed activity could vary widely.

\subsection{Leave-Unit-Out prediction}
\label{app:mdlag_pred_luo}

In the group number scaling experiments (Section~\ref{sec:scaling_M}, \fig{scalingM}), the number of groups, $M$, varied across simulated datasets while the number of units in total, $q$, remained fixed (see also \tbl{simulated-summary}, Scaling, $M$).
A fairer metric for comparison across datasets (as in \fig{scalingM}{a}), then, would involve not leave-group-out prediction, but rather leave-unit-out prediction.
To create a leave-unit-out prediction metric, we leveraged the development of leave-group-out prediction, above.

Suppose we have a fitted mDLAG model, for $M$ groups and $q_m$ units per group, where $m = 1,\ldots,M$.
We can then simply re-index the parameters of this model so that each unit $r = 1,\ldots,q$, with $q = \sum_m q_m$, is treated as its own ``group.''
The exact same equations developed for any of the leave-group-out prediction methods, above (equations \ref{eq:ypred_time}, \ref{eq:xpred_induce}, or \ref{eq:ypred_freq} for generating predictions, and equation \ref{eq:r2_lgo} for measuring performance), can then be used for leave-unit-out prediction: replace the index $m = 1,\ldots,M$ over groups with the index $r = 1,\ldots,q$ over individual units. 

Generating leave-unit-out predictions via mDLAG-time (Section~\ref{app:mdlag_pred_lgo_time}), however, was prohibitively expensive for the group number scaling experiments (\fig{scalingM}{a}).
Computationally, predictions require $\mathcal{O}(q^3)$ operations.
We therefore generated predictions via the more efficient mDLAG-frequency (Section~\ref{app:mdlag_pred_lgo_freq}; $\mathcal{O}(q)$ operations), regardless of which method (mDLAG-time, mDLAG-inducing, mDLAG-frequency) was used to fit the given model parameters.

Yet predictions via mDLAG-frequency exhibit edge effects (Section~\ref{sec:bias}, \fig{circulant}{c}).
To mitigate the impact of these edge effects on our measures of performance, we evaluated equation~\ref{eq:r2_lgo} for only a middle portion of the test trials.
For squared exponential GP timescales of 100 ms, removing the first and final 10 time points (200 ms) from evaluation was sufficient for edge effects to be rendered neglible.
For the group number scaling experiments, where there were $T = 50$ time points per test trial, we therefore used only the middle 30 time points ($t =$ 11 through 40).
We also employed leave-unit-out prediction via mDLAG-frequency in the trial length scaling experiments (Section~\ref{sec:scaling_T}, \fig{scalingT}), for consistency across simulated experiments and for computational benefit.
Test trials comprised $T = 500$ time points, but we similarly used only the middle $480$ time points ($t =$ 11 through 490).

\end{document}